\documentclass[11pt,a4paper]{article}
\usepackage[T1]{fontenc}
\usepackage[utf8]{inputenc}
\usepackage{lmodern,amsmath,amssymb,amsthm,mathtools}
\usepackage[a4paper,margin=25mm]{geometry}
\usepackage{microtype,booktabs,longtable,array,enumitem,needspace,placeins}
\usepackage[colorlinks=true,linkcolor=blue,citecolor=blue,urlcolor=blue]{hyperref}
\usepackage{xurl,pgfplots}
\pgfplotsset{compat=1.18}
\setlist{nosep,leftmargin=1.7em}
\newtheorem{theorem}{Theorem}[section]
\newtheorem{proposition}[theorem]{Proposition}
\newtheorem{lemma}[theorem]{Lemma}
\newtheorem{corollary}[theorem]{Corollary}
\theoremstyle{definition}
\newtheorem{definition}[theorem]{Definition}
\newtheorem{example}[theorem]{Example}
\theoremstyle{remark}

\newcommand{\one}{\mathbf 1}

\newcommand{\bits}{\{0,1\}}
\newcommand{\cl}{\operatorname{cl}}

\title{\textbf{Representational Acquisition Under Bounded Resources}}
\author{Yiling Wu\\[0.35em]\normalsize BridgeM, Inc.}
\date{}
\hypersetup{pdftitle={Representational Acquisition Under Bounded Resources},pdfauthor={Yiling Wu}}

\begin{document}
\maketitle
\begin{abstract}
A learner's obligation to keep acquired content usable imposes a storage--access tradeoff, with exact endpoints that force different physical organizations. For an exact concept-learning task, we determine the storage--read curve within a constant factor. One-read compound queries force answer-specific cells and exponentially costly updates; one-bit, order-independent updates instead force fact-specific cells. At minimum storage capacity, a robust counterpart holds: when update stretch is averaged uniformly over all semantic cube edges, small excess stretch bounds expected distance from a coordinate encoding under the uniform content distribution. It also bounds the information inaccessible to a one-read consumer. Preservation imposes further constraints. On a common invariant ready-state domain, old content preserved by ordinary zero-read overwrites must factor through the unwritten coordinates. Distributed encodings can require reading every cell to choose a preserving update. For conditional Horn reasoning, we obtain a second pair of exact endpoints: maintaining answers makes use local, while minimum storage and one-change named revision force a rule vector whose optimal query-read cost is determined exactly. These results explain how acquired distinctions, subsequent uses, and update constraints jointly determine the costs and organization of a retained resource.
\end{abstract}
\clearpage
\tableofcontents
\clearpage

\section{The cost of making acquired content usable}
\label{focus:introduction}

The central result is a resource tradeoff for a learner that must keep using what its evidence has established. Truthful examples create $N$ independent exclusions. Retaining them takes $N$ bits, but answering every compound exclusion request with at most $r_{\rm qry}$ reads requires storage on the scale $r_{\rm qry}2^{N/r_{\rm qry}}$. We determine this storage cost within a factor $2\mathrm e$, give exact read-local and write-local endpoints, and identify the physical organizations they force. The issue is how acquired information becomes an economically usable resource: an implementation can do work when new evidence arrives, when a query arrives, or through a maintained cache, and these choices have different costs.

An example fixes the meaning of the task. A learner has seen negative examples of a monotone conjunction on $d$ features. It has not yet identified the target criterion, but must answer questions about the remaining possibilities. Among the truthful negative examples there are $N=\binom d{\lfloor d/2\rfloor}$ independent exclusions. A final conjunction has a $d$-bit description; exact interim uncertainty has a much larger information requirement. If a consumer asks whether every member of a supplied collection is excluded, the whole request receives one read budget. Section~\ref{d:width-section} varies the collection's maximum size and locates exactly where the exponential one-read cost enters. The resulting lower bound concerns this declared collective use. A classifier asked only separate membership questions has a different access obligation.

The two endpoints explain why resource constraints can force different organizations of the same content. With one read per compound query, each nonconstant answer needs a designated physical cell. The final missing example can change $2^{N-1}$ of those answers, forcing that many cells to change. Conversely, if each fresh example changes one bit and different orders of the same examples reach the same complete physical state, each exclusion must have its own designated cell. A query combining all exclusions then needs $N$ reads. For canonical endpoints and nonadaptive queries, block caches attain both constant-factor optimal storage and the leading optimal update exponent $N/r_{\rm qry}$ throughout $r_{\rm qry}\le\alpha\log_2N$, for every fixed $0<\alpha<1$. Canonicality is required for that maintenance statement, but not for the storage lower bound or the one-read endpoint.

The distinction between constructing a representation and selecting one already available has a long history in theories of concept learning. Fodor's (1975) learning argument concerns the representational resources presupposed by hypothesis testing. Carey (2009) develops an account of conceptual change through bootstrapping, and Piantadosi (2021) gives a computational account of the emergence of higher-level representations from a more primitive dynamical basis. The present question is narrower and resource-relative: how must an acquired code be organized if each piece of evidence can change only a small part of memory? A finite collection of contents can always be indexed when implementation resources are unrestricted. Calling a retained word an address leaves open how experience can change it. We prove that a one-bit, order-independent acquisition process forces its index to have the coordinate structure of the acquired distinctions. Almost every minimum-length random index instead requires changing about half its bits per evidence item on average. The distinction concerns the geometry of actual updates, not the label attached to the code. Fixed-description bounds answer the separate question of how much family-specific information can have been supplied before experience. At tight storage the conclusion is robust: small excess update stretch, averaged uniformly over all semantic cube edges, bounds the expected distance from a signed coordinate permutation under uniform contents. Theorem~\ref{d:robust-tight} thereby links acquisition geometry to information accessible in one read. This is a constraint on the organization of an acquired code, rather than an account of the origin of its primitive vocabulary. Section~\ref{d:systematicity} relates the same results to the systematicity argument of Fodor and Pylyshyn (1988): exact success forces functional coordinates, whereas coordinates located in fixed cells require storage and update premises.

Keeping earlier content imposes another use obligation on the same storage. We characterize safe installation neighborhoods inside the fibers of an arbitrary distributed old-content decoder, including the stronger covering condition needed for repeated installation. A read-tree characterization shows how update reads must distinguish states requiring incompatible preserving writes. In particular, zero-read ordinary overwrites preserving old content on a common invariant ready-state domain force the restricted old decoder to factor through the cells they leave unwritten. These results separate capacity from the information an updater must obtain before safely changing memory. They also apply during learning: old contents must remain readable while negative examples incrementally establish a new uncertainty resource. Finally, a bounded-access information measure quantifies the difference between retaining a target and making it available to a permitted consumer. Complete target information can coexist with zero information in every permitted read transcript.

\subsection{Contribution and relation to the acquisition analyses}

The article connects three kinds of obligation. Task analysis determines what distinctions must survive future evidence; acquisition analysis requires actual transitions to establish those distinctions; resource analysis determines how storage and access constrain their implementation. \emph{Tasks That Force Representational Acquisition} develops the first question (Wu, 2026c), \emph{Acquiring, Keeping, and Reacquiring Representations} states the second as a capacity specification (Wu, 2026a), and \emph{From Learning Guarantees to Usable Content} connects learning algorithms to their native deployment (Wu, 2026b). The argument does not depend on access to these companion articles: the definitions used here are stated explicitly, and the external mathematical theorems are cited where they enter the proofs.

The contribution is a set of resource conclusions that distinguishes implementations agreeing on every correct answer. Minimum-length random indices and coordinate descriptions retain the same information but have sharply different acquisition geometry. Distributed old-content decoders can have the same read complexity and different safe-installation capacities. The same conditional Horn resource can be stored as rules, models, or answers, with different update obligations; allowing named deletion additionally forces the distinction between equivalent rule bases. At tight storage and one-change named revision, the physical code is forced to be the rule vector, and its optimal conditional-query cost is determined exactly. These comparisons turn a semantic success criterion into explicit, testable constraints on implementation.

Four questions organize the results:
\begin{enumerate}[leftmargin=*,itemsep=3pt]
\item \emph{What does continued use cost?} Theorems~\ref{focus:central} and~\ref{focus:star} give the learning tradeoff and its maintenance constraints; Theorems~\ref{d:horn-one-read} and~\ref{d:horn-query-exact} give the conditional Horn endpoints.
\item \emph{When must an index become a structured description?} Corollary~\ref{loc:coordinate-index} gives the exact local-update criterion, Theorem~\ref{loc:random-index} quantifies random-index costs, and Theorem~\ref{d:robust-tight} gives the tight-capacity stability result.
\item \emph{What permits acquiring new content while preserving old content?} Theorems~\ref{focus:retention-local} and~\ref{focus:retention-recurrent} distinguish one-shot and recurrent capacity; Proposition~\ref{shared:parity-update-reads} and Theorem~\ref{shared:learning-capacity} connect preservation to update reads and sequential evidence.
\item \emph{How much retained information can a bounded consumer use?} Theorem~\ref{d:availability-sharing} separates stored from accessible information, and Proposition~\ref{d:availability-fano} translates the access gap into native error.
\end{enumerate}
Table~\ref{focus:contribution-table} identifies the mathematical tools and the implementation conclusions derived from them. Figure~\ref{fig:resource-curves} displays the principal storage and maintenance bounds for a fixed number of learned distinctions.

Version spaces supply the learning setting (Mitchell, 1982). Exponential growth of a particular boundary representation is a classical concern (Haussler, 1988), while consistency-based alternatives show why such growth alone is not a representation-independent storage lower bound (Hirsh, Mishra, \& Pitt, 2004). Our count separates acquired states through native queries. The probe analysis uses the data-structure distinction between information stored and locations read (Fredman \& Saks, 1989; Buhrman et al., 2002). Maintenance is measured separately by changed cells and by total update accesses. The latter distinction matters when applying incremental-connectivity results or comparing with nonadaptive dynamic data structures (Brody \& Larsen, 2015).

\begin{table}[t]
\centering\small
\begin{tabular}{@{}>{\raggedright\arraybackslash}p{.29\linewidth}>{\raggedright\arraybackslash}p{.30\linewidth}>{\raggedright\arraybackslash}p{.34\linewidth}@{}}
\toprule
Established mathematical basis & Derivation supplied here & Consequence for acquisition \\
\midrule
Version spaces, probe spans, cube rigidity & Learning reduction, maintenance counts, width and endpoint-multiplicity bounds & Exact endpoints and a near-optimal nonadaptive resource curve \\
\addlinespace
Cube geometry, FKN stability, description counting & Exact and robust coordinate organization; random-index costs & Near-local acquisition bounds the one-read information gap \\
\addlinespace
Write-efficient coding, decision trees & Safe fibers, update-read bounds, and zero-read factorization & Capacity and update reads are separate retention constraints \\
\addlinespace
Predictive information, sharing, Fano bounds & Bounded-read availability and its relation to native error & Complete stored information can be unusable within the read budget \\
\addlinespace
Horn closure and view maintenance & Exact answer-local and revision-local endpoints & Semantic refinement and cheap maintenance need not increase together \\
\bottomrule
\end{tabular}
\caption{Mathematical sources and the implementation conclusions proved here. External theorems are attributed where applied; the table distinguishes the derived statements from their tools with the access and update assumptions made explicit.}
\label{focus:contribution-table}
\end{table}

\subsection{Scope of the necessity claims}

The core results use complete finite states, exact answers, and declared evidence and query interfaces. All evidence-dependent information available later is charged, including pointers, caches, and retained randomness. A cell is a physically addressable unit in this model; a probe is an access to that unit. These terms do not identify a neuron with a bit or count a whole neural forward pass as one probe. Applying the bounds to hardware, a neural implementation, or an external-memory agent requires a justified mapping of its states and accesses to this contract. Word and bounded-error results extend the storage analysis, but do not inherit every exact binary carrier conclusion.

Order independence concerns complete physical endpoints and is stated only where used. Section~\ref{d:coalescence-section} replaces it by a bound on the number of endpoints reached through selected acquisition paths. The resulting theorem quantifies how history dependence weakens the maintenance bound. Section~\ref{focus:open-questions} collects the remaining questions about adaptive reads, history dependence, shared retention, and robust organization.

Allowing arbitrarily long evidence streams does not remove a lower bound that is forced at a permitted prefix. A learner required to answer after every update must retain the distinctions needed at that prefix, however much additional evidence may arrive later. Thus more subsequent experience cannot substitute for the storage required by these interim uses. This statement concerns the declared interfaces and their worst-case prefixes; it is compatible with memory--sample tradeoffs for learning tasks with different output obligations.

The main sections develop the learning curve, word and randomized variants, full-task storage, and local indices. Section~\ref{focus:partition-cost} separates update reads from writes in relational learning. Sections~\ref{focus:retention}--\ref{d:horn-resources} treat shared preservation, bounded information access, and Horn acquisition costs; Section~\ref{focus:fault} gives the additional local costs of fault tolerance. The appendices prove the supporting factorization, carrier, geometric, and information lemmas. Appendix~\ref{focus:verification} describes the finite checks and the precisely delimited Lean certificates, which are available from the author on request.

\section{Content, operations, and resource accounting}
\label{d:framework}

\subsection{The content used by an implementation}

A task specifies a content space $M$, a query family $T$, and an answer function $b:M\times T\to Y$. Its legal evidence histories form a prefix-closed set $\mathsf{Hist}\subseteq E^*$, with prescribed content $m(e)$ after evidence word $e$. Evidence need not identify a unique final hypothesis. For example, its content may be the hypotheses still consistent with the observations, and a query may ask whether an instance can belong to the target according to that uncertainty.

An implementation has a complete retained state $q\in Q$, actual evidence updates $\delta_a$ for $a\in E$, and an installed query operation $O(q,t)$. Following the native-content definition in Wu (2026a, Definition~4.1), its content candidates are
\[
 C(q)=\{m\in M:O(q,t)=b(m,t)\text{ for every }t\in T\}.
\]
If the queries separate $M$, this set is empty or a singleton. In the latter case write $c(q)=m$. The target $m(e)$ and the native content $c(q)$ are different notions: a coherent answer profile can express an incorrect content. The definition concerns the system's installed uses, not an additional decoder available only to an analyst. This is a functional criterion; it does not replace a general account of biological function or worldly reference (Piccinini, 2022; Shea, 2023).

For an evidence word $e=a_1\cdots a_\ell$, write $A_e=\delta_{a_\ell}\circ\cdots\circ\delta_{a_1}$. The empty word $\varepsilon$ acts as the identity. The continuation profile is
\[
 \widehat O(q)(e,t)=O(A_eq,t).
\]
Two states are continuation-equivalent when their legal continuation tests and resulting answers agree. This is the continuation-content criterion of Wu (2026a). If admissibility depends on the evidence already received, that dependence is part of the interface. Present answers can coincide although later evidence distinguishes the states, as in dormant Horn rules and restricted classification interfaces (Wu, 2026c). Exact acquisition requires $O(A_eq_0,t)=b(m(e),t)$ at every declared query boundary; continued use and updating must use those same actual operations.

An \emph{intermediate-ready} task requires answers after every legal evidence prefix. A \emph{terminal-ready} task requires them only after a specified completion event. Retaining one completed hypothesis and retaining intermediate uncertainty are consequently different obligations. The main results concern the latter and specify when a query consumes several acquired distinctions in one request.

\subsection{Three levels of organization}

\emph{Content organization} consists of semantic distinctions and their dependencies. \emph{Operational organization} consists of laws of the actual updates and queries. \emph{Physical organization} consists of storage locations and their permitted access. A product of semantic factors does not, without additional premises, identify physical carriers. Likewise, commuting updates on behavioral equivalence classes need not commute as maps on complete physical states.

\begin{proposition}[Recoding invariance]
\label{d:recoding}
Let $\phi:Q\to Q'$ be a bijection and install
\[
 O'(q',t)=O(\phi^{-1}(q'),t),\qquad
 \delta'_a=\phi\circ\delta_a\circ\phi^{-1},\qquad q'_0=\phi(q_0).
\]
The implementations have identical answers on every legal evidence--query history. Their native and continuation profiles correspond under $\phi$. This does not imply identical access or update costs relative to a fixed physical cell decomposition.
\end{proposition}
\begin{proof}
Induction gives $A'_e\phi(q)=\phi(A_eq)$. Substitute this identity into the readout equation. A bijection need not preserve coordinate projections or Hamming adjacency: the map $(x_1,x_2)\mapsto(x_1\oplus x_2,x_2)$ preserves information but changes the cost of reading the first logical bit and of updating the second. Thus behavioral equivalence alone determines no such cost.
\end{proof}

The conjugated programs must actually be installed. The proposition does not attribute native content merely because some hypothetical decoder could recover it. Information-state analysis characterizes which distinctions suffice for a task (Sak\c{c}ak et al., 2024); the present question concerns the costs and physical organization of implementing their acquisition and use.

\subsection{A complete resource boundary}

Unless stated otherwise, a ready state consists of $s$ binary cells. All evidence-dependent information available to a delayed query belongs to these cells: pointers, controller modes, caches, learned programs, and retained examples are included. The query program receives the public query but no free evidence-dependent side information. A fixed program or read-only catalogue can depend on the public task parameters; its description length is charged separately as $b_{\rm fix}$.

A deterministic query with budget $r_{\rm qry}$ reads at most $r_{\rm qry}$ cells. The next address may depend on the query and previously observed values. Computation between probes is unrestricted, as in the bit-probe model (Fredman \& Saks, 1989; Miltersen, 1993). A nonadaptive query fixes all addresses before reading. Queries are read-only unless expressly stated otherwise. A $w$-bit word model instead counts $s$ words and states their width; its retained bit capacity is $sw$. A \emph{probe} is one cell access, designated as a read or a write; query-probe bounds therefore count reads, whereas total update-probe bounds count both.

An update budget $u$ bounds the Hamming distance between initial and final ready states, or the number of actual bit writes when that stronger guarantee is stated. Every changed endpoint bit needs a write; repeated intermediate writes can cost more. Write $r_{\rm upd}$ for update reads and $c_{\rm upd}$ for \emph{all} update probes, counting reads and actual writes. A blind one-cell overwrite has $r_{\rm upd}=0$ and $c_{\rm upd}=1$. Thus neither update budget can be substituted for $u$. An update can read extensively while changing few cells. The distinction is decisive for the comparison with incremental connectivity.

The complete ledger is
\[
 (b_{\rm fix},\ s_{\rm ready},\ s_{\rm peak},\ r_{\rm qry},\ u,\ r_{\rm upd},\ c_{\rm upd}),
\]
where peak storage includes temporary workspace. Statements about ready storage alone permit expensive acquisition or query computation. A construction supplies bounds only for the entries it explicitly analyzes. This convention prevents a storage saving from relocating the same information to an uncounted address, archive, or program. Sample count is not an optimized parameter in this ledger: an implementation must meet the stated resource bound at every required query boundary, including hard prefixes that may later receive arbitrarily much additional evidence.

Throughout, $d_H(q,q')$ denotes Hamming distance: the number of cell coordinates on which two equal-length strings differ. The vector $\mathbf b_i$ has a one in coordinate $i$ and zeros elsewhere; $x\oplus\mathbf b_i$ flips that bit. Euler's number is written $\mathrm e$, distinct from an evidence word $e$.

\subsection{Canonical states and task-preserving reductions}

Let $m_0=m(\varepsilon)$ be the initially warranted content and let $M_0\subseteq M$ be the reachable contents. An encoding $\mathsf{enc}:M_0\to Q$ is \emph{canonical} when it is injective, the actual initial state is $\mathsf{enc}(m_0)$, and actual updates take selected codes to the selected codes prescribed by the task. Choosing one representative per content for a static counting argument does not establish this update condition.

For histories in which each independent evidence item occurs at most once, commutation of the corresponding actual updates implies order independence by adjacent swaps. The resulting state can then be indexed by the evidence set. If repetitions are admitted and all histories with the same evidence set must give the same physical state, actual idempotence is additionally sufficient. Neither physical property follows merely from correctness of the answers. The local-update lower bounds below either assume a canonical encoding explicitly or obtain it on the restricted first-presentation histories.

A resource-preserving reduction maps legal evidence to actual updates and native queries to native queries, preserving both answers and the cost being compared. A sequence of separate one-probe calls is not a one-probe compound request. The main learning reduction therefore declares a compound query and charges its \emph{total} reads. Similarly, a lower bound on total update probes does not become a lower bound on endpoint writes. These distinctions allow the same content to be analyzed under several access contracts without conflating their implementation requirements.

Appendix~\ref{d:factors} derives functional product structure from independent assignment and read operations, and characterizes when Horn dependencies permit natural blocks. Appendix~\ref{d:locality} gives the corresponding binary geometric criterion for physical coordinates, while Appendix~\ref{d:access} treats one-cell carriers over general finite alphabets. These distinguish the semantic, operational, and physical levels before they are combined in the learning results.

\section{The resource curve for an acquired uncertainty resource}
\label{focus:central-section}

The central task concerns information that a learner has already acquired, although learning has not identified a unique hypothesis. The learner receives labeled instances, retains their consequences for the remaining possibilities, and answers queries about those possibilities. We first identify independent distinctions established by ordinary negative evidence. We then ask what it costs to retain, consult, and update that resource. The resulting lower bounds concern the installed consumer, not a particular version-space data structure.

\subsection{Evidence, uncertainty, and the native consumer}

Let $d\geq2$, let $X_d=\{0,1\}^d$, and consider the monotone conjunctions
\[
 \mathcal H_d=\{h_A:A\subseteq[d]\},\qquad
 h_A(x)=\mathbf1\{A\subseteq\{i:x_i=1\}\}.
\]
The empty conjunction is included. For a labeled evidence word $e$, its version space $\operatorname{VS}(e)$ contains exactly the masks consistent with every example in $e$. The query $\operatorname{May}(x)$ asks whether some surviving mask labels $x$ one. This is the standard consistency-based interpretation of version spaces (Mitchell, 1982), with the required use specified explicitly.

Put $k_{\rm mid}=\lfloor d/2\rfloor$, $\mathcal Z=\{Z\subseteq[d]:|Z|=k_{\rm mid}\}$, and $N=|\mathcal Z|=\binom d{k_{\rm mid}}$. Write $x^Z$ for the instance with zeros precisely at $Z$. Consider histories of negative examples $(x^Z,0)$, in arbitrary order and with repetitions permitted. Every such history is truthful for the same target $h_{[d]}$. If $S\subseteq\mathcal Z$ records the distinct examples received, then
\begin{equation}
 \operatorname{VS}_S=\{A\subseteq[d]:A\cap Z\ne\varnothing\text{ for every }Z\in S\}.
 \label{focus:version-space}
\end{equation}
For every $Z_0\in\mathcal Z$,
\begin{equation}
 \operatorname{May}_{\operatorname{VS}_S}(x^{Z_0})=1\quad\Longleftrightarrow\quad Z_0\notin S.
 \label{focus:independent-evidence}
\end{equation}
Indeed, an observed negative example cannot remain possibly positive. If $Z_0\notin S$, the mask $A=[d]\setminus Z_0$ intersects every $Z\in S$, because distinct sets of the same cardinality cannot contain one another. This mask survives and labels $x^{Z_0}$ one.

The consumer may also submit a finite compound request $S_{\rm req}\subseteq\mathcal Z$:
\[
 \operatorname{None}(S_{\rm req})=1
 \quad\Longleftrightarrow\quad
 \text{every }x^Z\ (Z\in S_{\rm req})\text{ is excluded from being positive}.
\]
This asks for a conjunction of consequences of the acquired uncertainty resource. If $v_S\in\{0,1\}^N$ is the indicator of $S$, then
\begin{equation}
 \operatorname{None}_{\operatorname{VS}_S}(S_{\rm req})=\prod_{Z\in S_{\rm req}}v_S(Z).
 \label{focus:compound}
\end{equation}
The empty request returns one. A budget of $r_{\rm qry}$ probes applies to the \emph{entire compound request}; it is not separately available for each constituent query. The compound interface is a substantive consumer requirement. With singleton $\operatorname{May}(x^Z)$ queries alone, restricted to $Z\in\mathcal Z$, the direct indicator supports one read and one changed bit per example, and the compound-query tradeoff below would not follow.

Every persistent quantity available to the consumer is counted in a complete binary ready state of length $s$. Query code is fixed for the task size; its addresses may depend on the supplied request and on previously read bits. Updates may use unrestricted computation and temporary workspace, which is erased before the next ready boundary. We write $u$ for the maximum Hamming distance between the ready states immediately before and after one example. Thus $u$ measures changed retained bits, rather than all processor instructions or all writes to temporary workspace.

For the results requiring a common physical code, a \emph{canonical first-presentation realization} means that histories listing each member of $S$ once, in any order, reach the same ready state $\mathsf{enc}(v_S)$. Fresh evidence then takes $\mathsf{enc}(v_S)$ to $\mathsf{enc}(v_{S\cup\{Z\}})$. The learner may have additional states after repeated examples. Alternatively, canonicality on these histories follows when distinct actual evidence updates commute wherever both are fresh: any two orders are connected by adjacent swaps. Commutation alone does not imply that repeated updates are idempotent, and no such implication is needed here.

\subsection{Storage, reading, and maintenance in one theorem}

\begin{theorem}[The resource curve and its exact endpoints]
\label{focus:central}
For the evidence and compound-query task above, the following assertions hold.
\begin{enumerate}
\item Every exact learner needs at least $N$ retained bits, even without an access bound. With $s$ binary cells and at most $r_{\rm qry}$ adaptive probes per compound query,
\begin{equation}
 \sum_{j=0}^{\min(r_{\rm qry},s)}\binom{s}{j}\geq2^N.
 \label{focus:dimension}
\end{equation}
If $s^*(N,r_{\rm qry})$ is the least possible ready storage and $1\leq r_{\rm qry}\leq N$, then
\begin{equation}
 \frac{r_{\rm qry}}{\mathrm e}2^{N/r_{\rm qry}}\leq s^*(N,r_{\rm qry})
 \leq r_{\rm qry}\bigl(2^{\lceil N/r_{\rm qry}\rceil}-1\bigr).
 \label{focus:static-curve}
\end{equation}
The displayed upper and lower bounds differ by a factor smaller than $2\mathrm e$.
\item If every compound query uses at most one probe, then some single truthful example changes at least $2^{N-1}$ retained bits. This bound is attained. Conversely, for a canonical first-presentation realization, if every fresh example changes at most one retained bit, the query $\operatorname{None}(\mathcal Z)$ requires at least $N$ probes. This bound is also attained, independently of how much surplus binary storage is available.
\item For each partition of the $N$ example indices into nonempty blocks of sizes $d_{{\rm blk},1},\ldots,d_{{\rm blk},r_{\rm qry}}$, there is an exact, canonically updated learner with
\begin{equation}
 s=\sum_{\ell=1}^{r_{\rm qry}}(2^{d_{{\rm blk},\ell}}-1),\qquad
 \text{compound-query probes}\leq r_{\rm qry},\qquad
 u=\max_\ell 2^{d_{{\rm blk},\ell}-1}.
 \label{focus:block-family}
\end{equation}
Balanced blocks therefore give $u=2^{\lceil N/r_{\rm qry}\rceil-1}$ and
\begin{equation}
 N\leq r_{\rm qry}\log_2(2u)<N+r_{\rm qry}.
 \label{focus:achievable-curve}
\end{equation}
They use at most the storage in the upper bound of~\eqref{focus:static-curve}, with an additional $O(N+\log s)$ transient bits during acquisition.
\end{enumerate}
\end{theorem}

\begin{proof}
All $2^N$ sets $S$ are reachable, and~\eqref{focus:independent-evidence} separates their native answer profiles. Two distinct profiles cannot share a complete ready state. This proves the $N$-bit lower bound without assuming canonicality. Retaining $v_S$ attains it when access is unrestricted. For possibility queries over the full instance domain, the installed routine can enumerate masks $A\subseteq[d]$, test~\eqref{focus:version-space}, and check the requested label, using the same retained state. This supplementary full-domain consumer uses unrestricted scans and has no $r_{\rm qry}$-probe guarantee; the costed interface in the theorem is the declared compound consumer.

Equation~\eqref{focus:compound} transfers the complete conjunction interface to this learning task. Choose one reached state for each $v\in\{0,1\}^N$; these representatives need not respect updates. By the common probe-span lemma (Lemma~\ref{d:probe-span}), all answers obtainable in $r_{\rm qry}$ binary probes belong to a vector space of dimension at most $\sum_{j\leq\min(r_{\rm qry},s)}\binom sj$. The $2^N$ conjunction functions are linearly independent by Lemma~\ref{d:query-dimension}, proving~\eqref{focus:dimension}. Since $s\geq N\geq r_{\rm qry}$, put $z=r_{\rm qry}/s\leq1$. Then
\begin{equation}
 z^{r_{\rm qry}}\sum_{j=0}^{r_{\rm qry}}\binom sj\leq(1+z)^s\leq \mathrm e^{r_{\rm qry}},
 \label{focus:binomial-estimate}
\end{equation}
so~\eqref{focus:dimension} gives $s\geq(r_{\rm qry}/\mathrm e)2^{N/r_{\rm qry}}$. The balanced construction below supplies the upper bound in~\eqref{focus:static-curve}. Their ratio is less than $\mathrm e2^{\lceil N/r_{\rm qry}\rceil-N/r_{\rm qry}}<2\mathrm e$.

For the one-probe endpoint, Lemma~\ref{d:oneprobe-functions} assigns a fixed identity-or-complement carrier to each nonconstant Boolean query. Distinct nonempty conjunctions are neither equal nor complementary: they differ somewhere and both equal one at $v=1^N$. Hence they have distinct carriers. Present the last missing example $Z_0$ after all other examples. The $2^{N-1}$ conjunctions containing $Z_0$ change from zero to one, forcing their designated cells to change. This argument permits history-dependent states. Storing every nonempty conjunction attains both the $2^N-1$ one-probe storage minimum and this worst-case update bound.

For the one-change endpoint, exact answers make the canonical map $\mathsf{enc}$ injective, and every logical cube edge changes exactly one physical bit. Theorem~\ref{d:rigidity}, whose proof uses the shared four-cycle lemma, gives distinct physical addresses $j_i$ with
\begin{equation}
 \mathsf{enc}(v)=\mathsf{enc}(0)\mathbin\oplus\bigoplus_{i:v_i=1}\mathbf b_{j_i}.
 \label{focus:one-write-organization}
\end{equation}
All other cells are constant. On $\mathsf{enc}(1^N)$, a correct decision tree for the full conjunction must read every $j_i$: if it omits $j_i$, the neighbor with logical coordinate $i$ zero follows the same transcript but requires the opposite answer. This gives $r_{\rm qry}\geq N$. The direct $N$-bit indicator attains $u=1$ and $r_{\rm qry}=N$.

Finally, in each block store every nonempty conjunction of its logical indicators. Initially all these entries are zero. Upon a new negative example, read the old singleton indicators in its block into temporary workspace, set the indicated logical bit to one, and recompute the affected cache entries. Other blocks are unchanged. Induction on the evidence history establishes every stored conjunction invariant. In a block of size $d_{{\rm blk},\ell}$, precisely the $2^{d_{{\rm blk},\ell}-1}$ conjunctions containing the new index can change. They all change when this is the block's last missing example, giving the exact worst-case $u$ in~\eqref{focus:block-family}. A compound request reads one designated entry from each block that it intersects and conjoins the returned values; an empty intersection contributes one without a read. The cache size, workspace bound, and balanced estimates follow. Temporary computation is not an uncharged persistent resource because its values are erased before the next ready query.
\end{proof}

The endpoints force different physical organizations. A single read forces separate carriers for distinct nonconstant answers; a single changed bit, together with actual order independence, forces separate carriers for the independent acquired facts. Intermediate block caches place more of the consumer's conjunction work into the maintained state. Equation~\eqref{focus:achievable-curve} is an achievable update--read relation. The theorem establishes an almost tight storage--read curve and exact maintenance endpoints. The following bound controls maintenance within the canonical model and, for nonadaptive reads in a growing logarithmic range, matches the construction's leading update exponent.

\subsection{Maintenance bounds and the nonadaptive update exponent}

\begin{theorem}[Maintenance from the last-insertion neighborhood]
\label{focus:star}
Let $N\geq1$, and let a canonical first-presentation realization of $N$ independent acquired bits answer all conjunction queries exactly. Suppose every fresh insertion changes at most $u$ retained binary cells. If queries use at most $r_{\rm qry}\geq1$ adaptive probes, then
\begin{equation}
 2^N\leq(2Nu)^{2^{r_{\rm qry}}-1}.
 \label{focus:adaptive-star}
\end{equation}
If the probed addresses are fixed before any cell is read, the sharper bound is
\begin{equation}
 2^N\leq(Nu)^{r_{\rm qry}}\,2^{2^{r_{\rm qry}}-1}.
 \label{focus:nonadaptive-star}
\end{equation}
These bounds permit arbitrary surplus storage and unrestricted computation or reads during acquisition.
\end{theorem}

\begin{proof}
Let $q_*=\mathsf{enc}(1^N)$ and let $q_i=\mathsf{enc}(1^N-\mathbf b_i)$ for $i\in[N]$, where the unit vector in this expression is logical. Inserting the missing coordinate takes every $q_i$ to the same $q_*$. Let $S_{\rm act}$ be the set of physical cells that differ between $q_*$ and at least one $q_i$, and put $d_{\rm act}=|S_{\rm act}|$. Each insertion changes at most $u$ cells, so $1\leq d_{\rm act}\leq Nu$. All other cells are constant on these $N+1$ states. Complement the coordinates in $S_{\rm act}$ whose baseline value at $q_*$ is one; this is only a change of notation, and makes the baseline vector zero.

For every $S_{\rm req}\subseteq[N]$, the answer to its conjunction is one at $q_*$ and is zero at $q_i$ exactly when $i\in S_{\rm req}$. Thus the required queries realize all $2^N$ distinct labelings of the $N$ neighbors, while fixing the baseline answer to one.

For a nonadaptive query, remove constant-coordinate reads and pad to $r_{\rm qry}$ reads, if necessary, by repeating any active coordinate. There are at most $d_{\rm act}^{r_{\rm qry}}$ ordered choices of these coordinates. Their Boolean decoder has at most $2^{r_{\rm qry}}$ input patterns and is fixed to one at the baseline pattern, giving at most $2^{2^{r_{\rm qry}}-1}$ possible decoders. Different choices may yield the same neighbor labeling, which only makes the upper count looser. Hence $2^N\leq d_{\rm act}^{r_{\rm qry}}2^{2^{r_{\rm qry}}-1}$, proving~\eqref{focus:nonadaptive-star}.

For an adaptive query, eliminate constant-coordinate nodes. The resulting tree has depth at most $r_{\rm qry}$ and can be padded to a full depth-$r_{\rm qry}$ binary tree by repeating active reads below shallow leaves and assigning their descendants the same output. This tree has $2^{r_{\rm qry}}-1$ internal nodes and $2^{r_{\rm qry}}$ leaves. There are at most $d_{\rm act}^{2^{r_{\rm qry}}-1}$ assignments of active coordinates to its nodes. The all-zero baseline follows the all-zero branch at every node, fixing one leaf label to one; the remaining leaves admit $2^{2^{r_{\rm qry}}-1}$ assignments. Thus $2^N\leq(2d_{\rm act})^{2^{r_{\rm qry}}-1}\leq(2Nu)^{2^{r_{\rm qry}}-1}$.
\end{proof}

\begin{corollary}[Two-probe acquisition has exponential maintenance cost]
\label{focus:two-probe}
Under the preceding hypotheses, two adaptive probes per compound request require
\[
 u\geq\frac{2^{(N-3)/3}}N.
\]
Two nonadaptive probes require
\[
 u\geq\frac{2^{(N-3)/2}}N.
\]
\end{corollary}
\begin{proof}
Set $r_{\rm qry}=2$ in~\eqref{focus:adaptive-star} and~\eqref{focus:nonadaptive-star}, respectively, and take the corresponding roots.
\end{proof}

\begin{corollary}[Uniformly optimal nonadaptive update exponent]
\label{focus:nonadaptive-range}
Let $u^*(N,r_{\rm qry})$ be the minimum worst-case number of changed retained bits among canonical exact learners with at most $r_{\rm qry}$ nonadaptive probes per compound query, without an additional bound on storage. For $1\leq r_{\rm qry}\leq N$,
\begin{equation}
 \frac{2^{(N-2^{r_{\rm qry}}+1)/r_{\rm qry}}}{N}\leq u^*(N,r_{\rm qry})
 \leq 2^{\lceil N/r_{\rm qry}\rceil-1}.
 \label{focus:nonadaptive-update-curve}
\end{equation}
For every fixed $0<\alpha<1$, uniformly over integers $1\leq r_{\rm qry}\leq\alpha\log_2N$ as $N\to\infty$,
\[
 \log_2u^*(N,r_{\rm qry})=(1+o(1))\frac{N}{r_{\rm qry}}.
\]
In that range, the same balanced block learner is within $2\mathrm e$ of minimum ready storage and attains this optimal leading update exponent.
\end{corollary}
\begin{proof}
Taking logarithms of~\eqref{focus:nonadaptive-star} gives
\[
 \log_2u\geq\frac{N-2^{r_{\rm qry}}+1}{r_{\rm qry}}-\log_2N.
\]
The balanced learner of Theorem~\ref{focus:central} has nonadaptive reads and $u=2^{\lceil N/r_{\rm qry}\rceil-1}$, proving~\eqref{focus:nonadaptive-update-curve}. In the stated range, $2^{r_{\rm qry}}\leq N^\alpha$ and therefore
\[
 1-N^{\alpha-1}-\frac{\alpha(\log_2N)^2}{N}
 \leq\frac{r_{\rm qry}\log_2u^*(N,r_{\rm qry})}{N}\leq1.
\]
Both error terms tend to zero independently of the permitted $r_{\rm qry}$. The storage claim follows from~\eqref{focus:static-curve} for that same construction.
\end{proof}

This simultaneous approximation concerns the storage and update minima at a given read budget. It does not determine the minimum update cost after a further storage limit is imposed. For two adaptive probes, the lower exponent remains $N/3$, compared with $N/2$ for the block upper bound. The adaptive count gives an exponent proportional to $N/(2^{r_{\rm qry}}-1)$, including when the read depth grows. It does not yield the stronger exponent of order $N/r_{\rm qry}$ attained by the block construction. For nonadaptive reads, Corollary~\ref{focus:nonadaptive-range} supplies that leading rate in its stated range, but does not settle larger $r_{\rm qry}$.

Canonicality is essential to the maintenance argument: without a common final ready state, last insertions can reach different physical representatives and no single neighborhood need have at most $Nu$ active cells. The theorem consequently does not establish the two-probe lower bound for arbitrary history-dependent learners. The balanced power-set family has an established analogue in bounded-union generators (Frein, L\'ev\^eque, \& Seb\H{o}, 2008). Our lower bounds allow arbitrary stored Boolean functions and native decoders, and the maintenance argument additionally concerns their actual evidence updates; it is not an optimality claim restricted to union generators.

These costs belong to the acquired uncertainty resource, not automatically to a final classifier. A selected member $h_A$ can be stored and used for classification in $d$ bits, and $d$ bits are necessary to support all $2^d$ distinct conjunction functions: masks differing at coordinate $i$ are separated by the instance whose only zero is at $i$. A learner required to answer the interim possibility queries must instead retain at least $N=\binom d{\lfloor d/2\rfloor}$ bits. The claim concerns distinguishable native uses, rather than the size of any particular version-space boundary representation; the latter alone does not determine reasoning complexity (Hirsh, Mishra, \& Pitt, 2004). The additional maintenance cost appears when the consumer requires compound consequences to be returned after every evidence update with a small total probe budget.
Nonadaptive access restrictions also support strong dynamic-data-structure
bounds for ordered-set tasks (Natarajan Ramamoorthy \& Rao, 2018). Their
median theorem restricts insertion adaptivity; their predecessor theorem
permits adaptive insertion and restricts query adaptivity. The present
bound instead uses conjunction queries, canonical ready endpoints, and
changed-cell cost, with update reads charged separately. Its leading
exponent follows from the stated local neighborhood count and does not
assert a frontier for those different ordered-set interfaces.

Figure~\ref{fig:resource-curves} compares the proved bounds with the block construction at a fixed problem size. Its storage and maintenance panels distinguish the premises under which each curve holds.
\begin{figure}[htbp]
\centering
\begin{tikzpicture}
\begin{axis}[width=.47\linewidth,height=5.1cm,xmin=1,xmax=16,ymin=0,ymax=65,xtick={1,4,8,12,16},xlabel={Query reads $r_{\rm qry}$},ylabel={$\log_2 s$},title={Retained storage},tick label style={font=\small},label style={font=\small},title style={font=\small},legend style={font=\scriptsize,at={(.98,.98)},anchor=north east,draw=none}]
\addplot[blue,thick] coordinates {(1,64.00000000) (2,33.00000000) (3,22.99999948) (4,17.99997799) (5,15.16972931) (6,13.32108252) (7,11.99753235) (8,10.99435344) (9,10.31174832) (10,9.79116289) (11,9.29691621) (12,8.96578428) (13,8.59618976) (14,8.40087944) (15,8.17492568) (16,7.90689060)};\addlegendentry{Block construction}
\addplot[black,dashed,thick] coordinates {(1,62.55730496) (2,31.55730496) (3,21.47560079) (4,16.55730496) (5,13.67923305) (6,11.80893413) (7,10.50751702) (8,9.55730496) (9,8.83834107) (10,8.27923305) (11,7.83491840) (12,7.47560079) (13,7.18082160) (14,6.93608845) (15,6.73086222) (16,6.55730496)};\addlegendentry{Lower bound}
\end{axis}
\begin{axis}[at={(.51\linewidth,0)},anchor=south west,width=.47\linewidth,height=5.1cm,xmin=1,xmax=16,ymin=0,ymax=65,xtick={1,4,8,12,16},xlabel={Query reads $r_{\rm qry}$},ylabel={$\log_2 u$},title={Changed bits},tick label style={font=\small},label style={font=\small},title style={font=\small}]
\addplot[blue,thick] coordinates {(1,63.00000000) (2,31.00000000) (3,21.00000000) (4,15.00000000) (5,12.00000000) (6,10.00000000) (7,9.00000000) (8,7.00000000) (9,7.00000000) (10,6.00000000) (11,5.00000000) (12,5.00000000) (13,4.00000000) (14,4.00000000) (15,4.00000000) (16,3.00000000)};
\addplot[black,dashed,thick] coordinates {(1,63.00000000) (2,24.50000000) (3,13.00000000) (4,6.25000000) (5,0.60000000) (6,0.00000000) (7,0.00000000) (8,0.00000000) (9,0.00000000) (10,0.00000000) (11,0.00000000) (12,0.00000000) (13,0.00000000) (14,0.00000000) (15,0.00000000) (16,0.00000000)};
\end{axis}
\end{tikzpicture}
\caption{Constructive budgets and proved lower bounds for $N=64$ independent exclusions. Storage bounds allow adaptive queries and history-dependent states. The maintenance lower curve assumes canonical endpoints and nonadaptive queries; it combines the exact one-read endpoint with~\eqref{focus:nonadaptive-update-curve} and the elementary bound $u\ge1$. Lines join integer budgets for readability. The gap at larger budgets is visible; the figure does not assert a complete joint frontier.}
\label{fig:resource-curves}
\end{figure}
\FloatBarrier

\subsection{The width of the requested consequence}
\label{d:width-section}

The cost of consulting an acquired resource depends on how much of it one request combines. A batch budget is appropriate when an installed consumer must decide whether an entire set of prerequisites has been established before acting. Separate membership calls, each with its own budget, impose a different obligation. The following result makes that distinction quantitative.

\begin{theorem}[Query width and exact one-read maintenance]
\label{d:query-width}
Require the same independent-exclusion learner to answer only conjunctions over sets of at most $w_{\rm req}$ indices, where $1\le w_{\rm req}\le N$. Put
\[
 \mathsf{ReqCount}(w_{\rm req})=\sum_{j=0}^{w_{\rm req}}\binom Nj.
\]
Every exact realization with $s$ binary cells and at most $r_{\rm qry}$ adaptive reads per query satisfies
\begin{equation}
 \sum_{j=0}^{\min(r_{\rm qry},s)}\binom sj\ge \mathsf{ReqCount}(w_{\rm req}).
 \label{d:width-dimension}
\end{equation}
At one read, the optimal retained storage and worst-case number of changed bits are, simultaneously,
\begin{equation}
 s=\mathsf{ReqCount}(w_{\rm req})-1,\qquad u=\sum_{j=0}^{w_{\rm req}-1}\binom{N-1}j.
 \label{d:width-endpoint}
\end{equation}
These conclusions allow history-dependent physical states.
\end{theorem}
\begin{proof}
Choose one attained state for each logical indicator $v\in\bits^N$. The monomials $\prod_{i\in S}v_i$ for $|S|\le w_{\rm req}$ are linearly independent, being a subfamily of the full Boolean monomial basis. The probe-span lemma bounds the dimension of all depth-$r_{\rm qry}$ readouts by the left side of~\eqref{d:width-dimension}.

For one read, each nonempty conjunction has its own fixed carrier: two distinct such functions are neither equal nor complementary. Thus $\mathsf{ReqCount}(w_{\rm req})-1$ cells are necessary. Just before the last missing exclusion $i$ arrives, precisely the requested conjunctions containing $i$ change value. There are $\sum_{j=0}^{w_{\rm req}-1}\binom{N-1}j$ of them, and their distinct carriers must change. Conversely, store every requested nonempty conjunction and update its value after each example. A fresh exclusion changes only entries containing its index; all of those entries change at the indicated last insertion. This attains both bounds. Repeated evidence changes none.
\end{proof}

For $w_{\rm req}=1$, the endpoint is the direct indicator with $N$ cells and one changed cell per example. Fixed $w_{\rm req}$ gives polynomial rather than exponential storage and maintenance. The full exponential endpoint occurs at $w_{\rm req}=N$, when one request can combine all $N$ distinctions. This describes an implementation contract for a collective decision, not a general lower bound on every use of learned concepts.

The lower bounds also apply to a learner for arbitrary truthful positive and negative examples if its native portfolio includes these compound requests: restrict that learner to the middle-layer negative histories used above. The restriction preserves every charged operation. This transfers the impossibility results to the full task. The matching block-cache construction serves the restricted history family; it does not establish a matching read--write frontier for arbitrary examples. Section~\ref{focus:content-organization} separately determines the full task's retained-information scale.

\subsection{History dependence and the cost of merging acquisition paths}
\label{d:coalescence-section}

Canonicality can be weakened to a measured property of the actual acquisition paths. For each $i\in[N]$, choose a legal history containing every exclusion except $i$, each once. Let $q_i$ be its reached state, and let $q_i^+$ be the state after inserting $i$. The $q_i^+$ all have the same correct content, but need not be the same physical state. Write $J_{\rm end}$ for the number of distinct endpoints among them. This quantity counts a specified set of converging histories; it is not inferred from agreement of their answers.

\begin{theorem}[Maintenance with bounded endpoint multiplicity]
\label{d:coalescence}
For an exact learner with all conjunction requests, $s$ binary cells, and at most $u$ changed bits per insertion, put $N_{\rm coal}=\lceil N/J_{\rm end}\rceil$ and $d_{\rm cap} =\min\{s,N_{\rm coal}u\}$. If queries use at most $r_{\rm qry}\ge1$ adaptive reads, then
\begin{equation}
 2^{N_{\rm coal}}\le(2d_{\rm cap})^{2^{r_{\rm qry}}-1}.
 \label{d:coalescence-adaptive}
\end{equation}
For nonadaptive queries,
\begin{equation}
 2^{N_{\rm coal}}\le d_{\rm cap}^{r_{\rm qry}}\,2^{2^{r_{\rm qry}}-1}.
 \label{d:coalescence-nonadaptive}
\end{equation}
No order-independence assumption is required.
\end{theorem}
\begin{proof}
Some endpoint $q_*$ is shared by at least $N_{\rm coal}$ of the selected paths. Retain exactly $N_{\rm coal}$ of their distinct missing indices, forming a set $I$. The union $S_{\rm act}$ of the cells on which their predecessor states differ from $q_*$ has size at most $N_{\rm coal}u$ and at most $s$. Every conjunction over a subset of $I$ answers one at $q_*$ and zero at predecessor $q_i$ exactly when it contains $i$. The interface therefore realizes all $2^{N_{\rm coal}}$ labelings of these predecessors while fixing the endpoint label. Restricting the native readers to these states and counting trees on $S_{\rm act}$, exactly as in Theorem~\ref{focus:star}, gives respectively $(2|S_{\rm act}|)^{2^{r_{\rm qry}}-1}$ and $|S_{\rm act}|^{r_{\rm qry}}2^{2^{r_{\rm qry}}-1}$ possible labelings. Substituting $|S_{\rm act}|\le d_{\rm cap}$ proves both inequalities.
\end{proof}

For canonical first presentations, $J_{\rm end}=1$. The theorem then sharpens the last-insertion bounds by retaining the storage cap $\min\{s,Nu\}$. For history-dependent implementations, the bound weakens according to how many distinct physical endpoints preserve the same acquired content. If $J_{\rm end}=N$, it gives no exponential-in-$N$ maintenance claim. Thus functional correctness does not justify silently replacing a learner's many representatives by a canonical code.

For adaptive queries, the same neighborhood count gives the following weaker growing-depth estimate
\[
 \log_2 u\ge \frac{N_{\rm coal}}{2^{r_{\rm qry}}-1}-\log_2(2N_{\rm coal}).
\]
For example, when $N_{\rm coal}\to\infty$ and $r_{\rm qry}\le(1-\eta)\log_2N_{\rm coal}$ for a fixed $0<\eta<1$, this implies $\log_2u\ge N_{\rm coal}^\eta-\log_2(2N_{\rm coal})$. It is weaker than the nonadaptive leading exponent $N_{\rm coal}/r_{\rm qry}$. At a prescribed storage budget, the dimension inequality,~\eqref{d:coalescence-adaptive}, and~\eqref{d:coalescence-nonadaptive} are joint outer bounds, while feasible block partitions supply inner bounds. The exact storage--read--write frontier, unrestricted history-dependent maintenance, and the two-adaptive-read gap remain undetermined by these arguments.

\section{Word access and bounded error}
\label{focus:word-random}

The storage--read obstruction survives two changes that are important when
interpreting a bit-probe model: a cell may contain several bits, and the
learner or its consumer may be randomized. These changes do not preserve
every exact carrier theorem. They do preserve quantitative restrictions on
retrieving all the answers required by the learning task.

\subsection{Cells with a finite alphabet}

Write $x\in\{0,1\}^{N}$ for the vector of exclusions established by the
middle-layer examples, and write
\[
 f_{S_{\rm req}}(x)=\prod_{i\in S_{\rm req}}x_i,\qquad S_{\rm req}\subseteq[N],
\]
for the compound answer, with $f_\varnothing=1$. All $2^N$ vectors are
reachable by truthful evidence. Let $s\geq1$ be the number of retained
cells, let each cell have $q_{\rm alph}=2^w$ possible values, where $w\geq1$, and let
$r_{\rm qry}\geq1$ bound the number of adaptive reads per compound request.
All target-dependent information available to the consumer, including
retained addresses and caches, belongs to these cells. The program and
query-dependent computation may be unrestricted, but cannot consult a
discarded evidence history. Queries are read-only.

\begin{theorem}[The word-alphabet dimension bound]
\label{focus:word-bound}
Every exact deterministic implementation of this interface satisfies
\begin{equation}
 \sum_{j=0}^{\min(r_{\rm qry},s)}\binom{s}{j}(2^w-1)^j\ \geq\ 2^N.
 \label{focus:word-dimension}
\end{equation}
In particular, $sw\geq N$. If $1\leq r_{\rm qry}\leq s$, then
\begin{equation}
 s\ \geq\ \frac{r_{\rm qry}}{\mathrm e(2^w-1)}\,2^{N/r_{\rm qry}}.
 \label{focus:word-explicit}
\end{equation}
The claims allow surplus cells and history-dependent ready states.
\end{theorem}

\begin{proof}
For each $x$, select one permitted evidence history yielding that vector
and denote its actual ready word by $\mathsf{enc}(x)\in\{0,\ldots,q_{\rm alph}-1\}^s$.
No assertion is made that other histories with the same $x$ reach this
word. Exactness makes every $f_{S_{\rm req}}$ available from these representatives.

The finite-alphabet probe-span lemma (Lemma~\ref{d:probe-span}) puts every answer function in a common space of dimension at most
\[
 \sum_{j=0}^{\min(r_{\rm qry},s)}\binom sj(q_{\rm alph}-1)^j.
\]
This statement permits adaptive reads; it uses one indicator per nonbaseline cell value, rather than treating a word as a binary cell. Restrict the span to the selected representatives. It must contain the $2^N$ linearly independent conjunctions of Lemma~\ref{d:query-dimension}, proving~\eqref{focus:word-dimension}. Singleton queries distinguish every $x$, so $q_{\rm alph}^s\geq2^N$ and $sw\geq N$.

Finally, for $1\leq r_{\rm qry}\leq s$,
\[
 \sum_{j=0}^{r_{\rm qry}}\binom{s}{j}(q_{\rm alph}-1)^j
 \leq (q_{\rm alph}-1)^{r_{\rm qry}}\sum_{j=0}^{r_{\rm qry}}\binom{s}{j}
 \leq\left(\frac{\mathrm e\,s(q_{\rm alph}-1)}{r_{\rm qry}}\right)^{r_{\rm qry}}.
\]
The last step uses the binomial estimate~\eqref{focus:binomial-estimate}.
Combining the inequalities proves \eqref{focus:word-explicit}.
\end{proof}

\begin{theorem}[Word constructions within a constant factor]
\label{focus:word-construction}
Let $s_w^*(N,r_{\rm qry})$ be the least number of $w$-bit words for the exact compound-query task with at most $r_{\rm qry}$ word probes, for positive integers $N,w,r_{\rm qry}$. If $r_{\rm qry}w\geq N$, then
\begin{equation}
 s_w^*(N,r_{\rm qry})=\left\lceil\frac Nw\right\rceil.
 \label{focus:word-packing}
\end{equation}
If $r_{\rm qry}w<N$, put $k=\lfloor N/r_{\rm qry}\rfloor$ and $k_{\rm rem}=N-kr_{\rm qry}$. Then
\begin{align}
 \frac{r_{\rm qry}}{\mathrm e(2^w-1)}2^{N/r_{\rm qry}}
 &\leq s_w^*(N,r_{\rm qry})\leq(r_{\rm qry}+k_{\rm rem})2^{k-w+1}-r_{\rm qry}
 \label{focus:word-curve}\\
 &\leq r_{\rm qry}\bigl(2^{\lceil N/r_{\rm qry}\rceil-w+1}-1\bigr).
 \nonumber
\end{align}
The displayed construction uses fewer than $4\mathrm e$ times the minimum number of words. Both regimes admit canonical actual evidence updates and nonadaptive, read-only queries. The claims concern ready storage and query probes, with unrestricted update computation and temporary workspace.
\end{theorem}
\begin{proof}
If $r_{\rm qry}w\geq N$, pack the $N$ raw bits into $\lceil N/w\rceil\leq r_{\rm qry}$ words, padding unused positions with constants. Reading these words supplies every requested conjunction. The capacity bound $sw\geq N$ proves optimality.

Now suppose $r_{\rm qry}w<N$. Every feasible implementation has $s\geq\lceil N/w\rceil>r_{\rm qry}$, so~\eqref{focus:word-explicit} applies. Split the $N$ coordinates into $r_{\rm qry}-k_{\rm rem}$ blocks of size $k$ and $k_{\rm rem}$ blocks of size $k+1$. Each block has size $d_{\rm blk}\geq w$. Select $w-1$ low coordinates and let the other $d_{\rm high}=d_{\rm blk}-w+1\geq1$ coordinates be high. For every nonempty subset $S$ of the high coordinates, store one word containing the $w-1$ raw low bits and the conjunction of the high bits in $S$. There are $2^{d_{\rm high}}-1$ words in this block.

For a request whose high part in that block is nonempty, read its corresponding word. Conjoin the stored high-subset value with the requested low bits in that word. If only low coordinates are requested, read any fixed word of the block and ignore its conjunction bit. If the block contributes no requested coordinate, make no read. All addresses depend only on the query. Thus at most one word per block suffices, and no word for the empty high subset is needed. Summing yields
\[
 (r_{\rm qry}-k_{\rm rem})(2^{k-w+1}-1)+k_{\rm rem}(2^{k-w+2}-1)
 =(r_{\rm qry}+k_{\rm rem})2^{k-w+1}-r_{\rm qry}.
\]
The coarser upper bound in~\eqref{focus:word-curve}, divided by its lower bound, is strictly less than
\[
 \mathrm e(2^w-1)2^{\lceil N/r_{\rm qry}\rceil-w+1-N/r_{\rm qry}}<4\mathrm e.
\]

To implement evidence updates, initialize every exclusion bit to zero. In the packing regime, set its packed bit when its evidence arrives. In the cache regime, read the affected block's high singleton entries and its low bits into temporary workspace, set the indicated logical bit to one, and recompute the words in that block. This uses only actually stored information. Induction establishes the cache invariant after every example. The physical endpoint is a function of the exclusion vector alone; updates therefore commute and repeated evidence is idempotent. Queries preserve the words. Workspace is discarded before the ready boundary.
\end{proof}

The separation into regimes is necessary. If a word already holds the relevant raw coordinates, a formula with a negative cache exponent would incorrectly predict fewer than one word per block. Direct packing supplies the capacity-optimal solution in exactly that case. For $w=1$, the cache construction reduces to the binary block learner of Theorem~\ref{focus:central}.

\subsection{Randomized learning and native retrieval}

Fix $0\leq\upsilon_{\rm qry}<1/2$. The bounded-error requirement here is
\emph{pointwise}: for every permitted history and every $S_{\rm req}$, the native
answer is correct with probability at least $1-\upsilon_{\rm qry}$. The
probability includes acquisition and retrieval randomness. It is not
an average over a selected distribution of queries. Acquisition and
retrieval may share public randomness independent of the target and
query; private information retained from acquisition must be included
in the counted ready memory. Retrieval has a worst-case bound of $r_{\rm qry}$
reads on every run, rather than merely an expected bound.

\begin{theorem}[A bounded-error storage--read restriction]
\label{focus:random-bound}
For each fixed $\upsilon_{\rm qry}<1/2$, there is a constant $c_{\upsilon_{\rm qry}}>0$
such that, for all sufficiently large $N$, any implementation just
described satisfies
\begin{equation}
 r_{\rm qry}\bigl(\lceil\log_2s\rceil+w\bigr)\ \geq\ c_{\upsilon_{\rm qry}} N.
 \label{focus:random-communication}
\end{equation}
Consequently,
\begin{equation}
 s\ \geq\ 2^{c_{\upsilon_{\rm qry}} N/r_{\rm qry}-w-1}.
 \label{focus:random-space}
\end{equation}
The bound permits adaptive queries, randomized acquisition, and public
randomness. It concerns the memory accessed by native retrieval and
does not bound the computation used to prepare that memory.
\end{theorem}

\begin{proof}
We use the standard cell-probe-to-communication simulation (Miltersen et al., 1998, Section~1.1) to reduce two-party set disjointness on $[N]$ to native retrieval.
Alice receives $A\subseteq[N]$ and sets $x_i=1$ exactly when $i\notin A$.
She locally presents, in a fixed order, the truthful negative examples
whose exclusion coordinates have $x_i=1$. She simulates acquisition
and keeps its resulting $s$ words. Bob receives $S_{\rm req}\subseteq[N]$ and
simulates the native consumer for this query. The parties use the
implementation's prescribed public randomness; private acquisition
coins belong to Alice and private retrieval coins to Bob.

When the consumer requests a cell, Bob transmits its address using
$\lceil\log_2s\rceil$ bits, and Alice transmits its $w$-bit contents.
Bob then computes the next address or the answer. All information used
by the simulated consumer is available from its query, its allowed
randomness, and the received contents. Any early termination can be
padded by unused reads to make exactly $r_{\rm qry}$ exchanges. When $s=1$, no
address bits are needed. The total communication is at most the left
side of \eqref{focus:random-communication}.

Bob's output has the original error guarantee, and its correct value is
$f_{S_{\rm req}}(x)=1$ exactly when $A\cap S_{\rm req}=\varnothing$. The classical
bounded-error communication lower bound for set disjointness is
$\Omega(N)$, including public-coin protocols (Kalyanasundaram \&
Schnitger, 1992; Razborov, 1992). The usual constant-error formulation
extends to each fixed $\upsilon_{\rm qry}<1/2$ by a constant number of independent
repetitions and majority vote. It therefore yields
\eqref{focus:random-communication}. Since
$\lceil\log_2s\rceil\leq\log_2s+1$, rearrangement proves
\eqref{focus:random-space}.
\end{proof}

The probabilistic conclusion does not require replacing the linear
$N$ in the exponent by $\sqrt N$. Such a weaker bound is available
from approximate rank, but the communication reduction is stronger
for the classical systems considered here. To locate that alternative
precisely, suppose the counted ready state separates private
acquisition randomness from fresh retrieval randomness, with no
uncounted shared seed. Each query's acceptance function on ready
words lies in the common feature span used above. If $\mu_x$ is the
ready-state distribution, averaging its features gives a factorization
of the acceptance matrix of rank at most the left side of
\eqref{focus:word-dimension}. Pointwise error then implies a lower
bound by the entrywise $\upsilon_{\rm qry}$-approximate real rank of the $0$--$1$ matrix
\[
 \mathsf{Disj}_{x,S_{\rm req}}=\prod_{i\in S_{\rm req}}x_i,
 \qquad x\in\bits^N,\quad S_{\rm req}\subseteq[N].
\]
This is the disjointness matrix after identifying the set in its row with $\{i:x_i=0\}$. For each fixed $\upsilon_{\rm qry}<1/2$, this rank is at least $2^{\Omega_{\upsilon_{\rm qry}}(\sqrt N)}$: Razborov's (2003, Section~5) approximate trace-norm argument implies this bound, and Sherstov (2011, Theorems~1.4 and~8.1, with the embedding in Section~9) supplies a pattern-matrix proof. To pass from $0$--$1$ to sign matrices, replace an approximant $\mathsf{App}$ to $\mathsf{Disj}$ by $2\mathsf{App}-\mathsf{Ones}$, where $\mathsf{Ones}$ is the all-ones matrix. The error becomes at most $2\upsilon_{\rm qry}<1$ and the rank increases by at most one. Thus the cited sign-matrix result gives the stated $0$--$1$ lower bound.
An uncounted shared random seed requires additional care in that
factorization, because it may change both factors. The communication
proof already permits such target-independent public randomness.

These bounds concern storage and access. They neither establish a
bounded-error counterpart of designated carriers nor constrain
maintenance by the exact one-write rigidity theorem. Moreover,
pointwise reliability is essential: a test distribution concentrated
on the empty query, whose answer is always one, imposes no such
storage requirement. If an acquisition guarantee instead assigns
probability at least $1-\rho$ to states whose per-query error is at
most $\tau$, it implies the present pointwise guarantee with
$\upsilon_{\rm qry}=\rho+(1-\rho)\tau$, whenever this value is below $1/2$.

\section{How much content, and which physical organization?}
\label{focus:content-organization}

The preceding bounds concern a restricted family of truthful examples
whose independent effects make the resource tradeoff explicit. Two
questions remain. Does that family overstate the storage needed by the
complete learning task? And do the local-access endpoints determine
physical carriers, rather than merely counting states? This section
answers both questions. It also identifies an obstruction to extending
coordinate alignment from independent observations to class mergers.

\subsection{The complete conjunction task has the same leading storage cost}

Let $\mathcal H_d=\{h_A:A\subseteq[d]\}$, where
$h_A(x)=1$ exactly when $A\subseteq\operatorname{supp}(x)$, and include
the empty conjunction. After any finite, consistent history of positive
and negative examples, let $\mathcal W\subseteq\mathcal H_d$ be the nonempty
version space. The native query remains
$\operatorname{May}_{\mathcal W}(x)=[\exists h\in \mathcal W:h(x)=1]$.
Examples must be truthful for some fixed target throughout the history;
the learner must answer after every prefix. Write $\mathsf{Ded}(d)$ for the
Dedekind number, the number of upward-closed families of subsets of
$[d]$, including the empty family, and put $\mathsf{Ded}(0)=2$.

\begin{theorem}[Asymptotically exact retained storage for full learning]
\label{focus:dedekind}
For $d\ge1$, let $s^*_{\rm ready}(d)$ be the least number of retained
binary bits for the complete task above, without bounds on computation,
temporary workspace, or probes within an operation. Then
\begin{align}
 \left\lceil\log_2(\mathsf{Ded}(d)-1)\right\rceil
 &\le s^*_{\rm ready}(d)
 \le\left\lceil\log_2 \mathsf{VSCount}(d)\right\rceil,
 \label{focus:dedekind-sandwich}\\
 \mathsf{VSCount}(d)&=\sum_{p=0}^{d}\binom dp\bigl(\mathsf{Ded}(p)-1\bigr)
 \le 2^d\mathsf{Ded}(d).
 \nonumber
\end{align}
Consequently, with $N=\binom d{\lfloor d/2\rfloor}$,
\[
 s^*_{\rm ready}(d)=(1+o(1))N\qquad(d\longrightarrow\infty).
\]
The number $\mathsf{VSCount}(d)$ is exactly the number of nonempty reachable version
spaces, not necessarily the number of distinct current May profiles.
\end{theorem}

\begin{proof}
First consider histories containing only negative examples. Every
nonempty upward-closed family $\mathcal U\subseteq2^{[d]}$ is reachable: for
each maximal $S\notin \mathcal U$, give the negative example with support $S$.
A candidate $A$ survives exactly when it is contained in none of these
maximal excluded sets, equivalently when $A\in \mathcal U$. These examples are
all truthful for the full conjunction $h_{[d]}$, because $[d]\in \mathcal U$
and therefore every such $S$ is a proper subset of $[d]$. The full
family requires no examples. At this version space,
\[
 \operatorname{May}_{\mathcal U}(x)=1
 \quad\Longleftrightarrow\quad
 \operatorname{supp}(x)\in \mathcal U:
\]
the forward direction uses upward closure, and the reverse direction
uses the surviving candidate $A=\operatorname{supp}(x)$. Thus the
$\mathsf{Ded}(d)-1$ nonempty families have distinct native profiles and require
at least that many complete boundary states.

For an arbitrary consistent history, let $F_+$ be the intersection of
the supports of its positive examples, taking $F_+=[d]$ if there are
none. Positive evidence requires $A\subseteq F_+$. Each negative
example requires $A\nsubseteq S$ for its support $S$, an upward-closed
condition. Hence the surviving masks form a nonempty upward-closed
family inside $2^{F_+}$. Its unique largest element is $F_+$.
Conversely, every nonempty upward-closed family inside $2^{F_+}$ is
reachable by giving the positive example with support $F_+$, followed
by negative examples at its maximal excluded subsets of $F_+$.
The target $h_{F_+}$ makes this history truthful. Different choices of
$F_+$ cannot describe the same nonempty family, since $F_+$ is its largest
element. Counting gives precisely $\mathsf{VSCount}(d)$.

Enumerate these families and retain the index of the current family
in $\lceil\log_2 \mathsf{VSCount}(d)\rceil$ bits. An update decodes it, intersects
with the consistency condition of the incoming example, and encodes
the result; a query decodes it and evaluates May. The algorithms can
enumerate finite families uniformly from $d$ and need no
target-dependent preinstalled catalogue. Their running time and
temporary workspace are unrestricted here. Finally $\mathsf{Ded}(p)\le \mathsf{Ded}(d)$
for $p\le d$, so $\mathsf{VSCount}(d)\le2^d\mathsf{Ded}(d)$. Kleitman's enumeration theorem
gives $\log_2\mathsf{Ded}(d)=(1+o(1))N$ (Kleitman, 1969), and $d=o(N)$.
Both sides of~\eqref{focus:dedekind-sandwich} therefore have leading
term $N$.
\end{proof}

The complete task thus requires asymptotically the same retained
storage as the independent middle-layer examples. This conclusion
does not transfer the entire probe--update curve to every version
space. Nor does the enumerative upper bound control peak storage:
decoding and re-encoding may use much more workspace than the retained
index. Those are distinct entries in the resource account.

Current May answers also need not identify the complete version
space. On two features, $\{\{1\}\}$ and
$\{\{1\},\{1,2\}\}$ have the same May answers. The theorem avoids
equating these objects: its lower bound counts distinguishable native
profiles, whereas its constructive upper bound retains a sufficient
state for subsequent evidence. It therefore does not assume that an
analyst's particular representation of a version space is already
forced by the current query interface.

\subsection{Local writes force carriers for native uncertainty judgments}

The write-local endpoint has a useful learning interpretation when
each instance initially has unknown membership. Take $d$ instances,
an arbitrary unknown target class, and truthful first observations
$(i,\xi)$ with $\xi\in\{0,1\}$. The acquired content is a partial labeling
$x\in\{\bot,0,1\}^d$. Native uncertainty judgments are
\[
 (\operatorname{Must}_i(x),\operatorname{May}_i(x))=
 \begin{cases}(0,1),&x_i=\bot,\\(0,0),&x_i=0,\\(1,1),&x_i=1.
 \end{cases}
\]
Together these answers distinguish every partial labeling. An
acquisition edge replaces an unknown coordinate by either label.
A canonical physical encoding $\mathsf{enc}$ assigns one boundary state to
each partial labeling, and actual evidence updates must follow these
edges. This condition concerns actual states, not representatives
chosen afterwards from several states with the same answers.

\begin{theorem}[Fixed carriers for Must and May]
\label{focus:carriers}
Suppose a canonical exact learner stores its complete boundary state
in $s$ ternary cells and changes at most one cell per first
observation. Then, for each $i$, there are fixed cells
$j_i^{\rm must}$ and $j_i^{\rm may}$ whose respective values determine
$\operatorname{Must}_i$ and $\operatorname{May}_i$ throughout the reachable
state space. The two cells may coincide; cells assigned to different
instances cannot coincide. Every other cell is constant.
More precisely, the two observation arms of an instance either share
one three-valued cell or use two cells, one for each label. If
$s=d+s_{\rm free}$, at most $s_{\rm free}$ instances use two cells, and this bound is attained
for $0\le s_{\rm free}\le d$.
\end{theorem}

\begin{proof}
Exactness makes $\mathsf{enc}$ injective, and each acquisition edge changes exactly one cell. Label the edge by that cell's address. A square obtained from observations of two different instances has four distinct encoded vertices. Lemma~\ref{d:square-locality} applies because every cell is ternary: opposite edges have equal addresses and adjacent edges have distinct addresses.

Fix one label arm of instance $i$. Varying another partial label
along an acquisition edge carries this arm to the opposite edge of
a square. All contexts of the other instances are connected by such
edges. The arm therefore has a fixed address, independently of those
contexts. The square argument also makes the addresses of arms from
different instances disjoint. Changing any other instance leaves
these cells unchanged, so their values depend only on $x_i$;
cells assigned to no arm remain constant.

If the two arms of $i$ share a cell, its three status values are
distinct by injectivity, and either native judgment can be read from
it. If the addresses are distinct, write the pair's values as
\[
 \bot\mapsto(v^{\rm base}_0,v^{\rm base}_1),\qquad
 0\mapsto(c_0,v^{\rm base}_1),\qquad
 1\mapsto(v^{\rm base}_0,c_1),\qquad c_\xi\ne v^{\rm base}_\xi.
\]
Then the judgments are read at the zero-arm and one-arm addresses, respectively:
\[
 \operatorname{May}_i=[q_{\,{j_i^{\rm may}}}=v^{\rm base}_0],
 \qquad
 \operatorname{Must}_i=[q_{\,{j_i^{\rm must}}}=c_1].
\] Each split instance uses an additional distinct
cell, so there are at most $s-d=s_{\rm free}$ of them. Sharpness follows by
putting $d-s_{\rm free}$ statuses in individual ternary cells and the remaining
$s_{\rm free}$ in two-cell codes $\bot\mapsto00$, $0\mapsto10$, $1\mapsto01$.
\end{proof}

Surplus storage changes whether the two judgments share a cell; it
does not remove their fixed carriers under these premises. For binary
cells, the two arms cannot share an address: from a common initial
bit value both would produce the same endpoint. The proof therefore
also yields $s\ge2d$, attained by the two-bit code above. This is the
uncertainty counterpart of the write-local endpoint: local evidence
updates force accessible organization, whereas a mere capacity bound
would require only $\lceil d\log_2 3\rceil$ bits.

\begin{proposition}[An operational sufficient condition for canonicality]
\label{focus:commutation}
For truthful first observations, if the actual updates for different
unobserved instances commute at every reachable state where both
orders are admissible, then all histories giving the same partial
labeling reach the same state. Exactness and one-cell updates
therefore imply Theorem~\ref{focus:carriers}. If repeated observations
are admitted, the same conclusion follows when these commuting
updates are also physically idempotent on reachable states.
\end{proposition}

\begin{proof}
Two duplicate-free histories of the same partial labeling are
permutations of the same observations. Adjacent swaps transform one
into the other, and actual commutation preserves the endpoint at
each swap. With repetitions, proceed by induction on the history.
For an observation already present, the duplicate-free prefix can be
reordered so that its earlier copy is last. Physical idempotence
therefore makes the repeated observation leave that endpoint fixed.
A first observation extends the duplicate-free history as before.
The resulting canonical encoding is injective because exact native
answers distinguish partial labelings.
\end{proof}

Semantic idempotence alone is insufficient: a one-variable learner
can move through $00\to10\to11\to10\to\cdots$ under repeated
evidence and read both noninitial states as ``observed.'' Each update
changes one bit, but a set of observations has no unique physical
endpoint. Thus the operational premise should specify either first
observations or actual idempotence.

Canonicality is substantive even without repetitions. For two
instances in three ternary cells, use symbol $0$ for unknown and
symbols $1,2$ for the two observed labels. Put the first observation
in the cell with that instance's address; put the second observation
in cell $3$. Reachable states are
\[
 (0,0,0),\ (\xi,0,0),\ (0,\xi^{\prime},0),\ (\xi,0,\xi^{\prime}),\ (0,\xi^{\prime},\xi),
 \qquad \xi,\xi^{\prime}\in\{1,2\}.
\]
Decode $(\xi,0,c)$ as statuses $(\xi,c)$ when $\xi\ne0$, and $(0,\xi^{\prime},c)$
as $(c,\xi^{\prime})$ when $\xi^{\prime}\ne0$. All native judgments are exact and every
observation changes one cell. Yet no single cell determines either
Must or May for either instance: its label can vary while that cell
is held fixed. The two learning orders generally reach different
states, so the updates do not commute. Extra capacity without the
operational premise therefore does not force designated carriers.

\subsection{Coupled class mergers obstruct one-cell canonical learning}

Independent observations generate commuting squares. Same-class
evidence instead generates a lattice of partitions, because equality
is transitive. Its local geometry rules out the analogous encoding
over small alphabets.

\begin{theorem}[A partition-merger obstruction]
\label{focus:partition-obstruction}
For a domain of at least three objects, no canonical exact learner
of the warranted equivalence relation can implement every strict
pair merger by changing at most one cell if every cell has at most
four values. The number of cells may be arbitrarily large. For
exactly three objects, five cell values suffice.
\end{theorem}

\begin{proof}
Restrict to three objects, leaving all others separate. There are
five relevant partitions: the discrete partition $\pi_{\rm disc}$, the three
partitions $p_1,p_2,p_3$ merging just one pair, and the common top
partition $z$ merging all three. Each $p_i$ is one strict merger
from $\pi_{\rm disc}$ and one strict merger from $z$. Native same-class queries
distinguish these five contents, so their canonical codes are
distinct. Each code for $p_i$ must be a Hamming-distance-one neighbor
of both $\mathsf{enc}(\pi_{\rm disc})$ and $\mathsf{enc}(z)$.

Two distinct vertices in a Hamming graph have at most two common
neighbors when each coordinate alphabet has size at most four.
Indeed, distance greater than two gives none; distance two gives
exactly the two ways of changing one differing coordinate; distance
one permits common neighbors only in that coordinate, with at most
two values other than the endpoint values. There cannot therefore
be three distinct codes for the $p_i$. For three objects and a
five-valued cell, encode the five partitions by its five values;
every strict merger changes that one cell.
\end{proof}

This obstruction explains why the independent-fact endpoints do not
automatically imply a cell for each learned class. Small-cell
locality and canonical acquisition are already incompatible with
three-object merging. A learner must relax its write budget, its
order-independent physical coding, or its cell alphabet. These are
resource choices, not failures of the semantic notion of a class.

\section{Local indices and acquired descriptions}
\label{focus:catalogue}

The intermediate learning task has $2^N$ distinguishable acquired contents.
An implementation can name its retained word an index into those contents or
a description of them.  The operational question is how that word changes
with evidence and how the installed consumer reads it.  This section gives
three answers.  Free fixed computation can absorb a finite catalogue; local
acquisition constrains the geometry of its index; and a bound on the fixed
description charges information specific to the content family.
This resource-relative analysis sharpens the behavioral limitation stated
in Wu (2026a, Section~10.3): indexing a content space and constructing a
description cannot be separated by an unrestricted input--output test alone.

\subsection{Fixed catalogues and the resource boundary}

\begin{proposition}[Compiling a fixed catalogue]
\label{loc:compilation}
Consider a deterministic implementation with a finite catalogue $C$ fixed
before experience and a complete mutable state.  If fixed computation and
fixed program description are unrestricted, an implementation with $C$
embedded in its program has the same mutable states and transitions and the
same answers.  Its mutable query probes, update reads, and endpoint Hamming
costs are unchanged.  If accesses to the fixed catalogue were also charged
as probes, their removal cannot increase these probe counts.
\end{proposition}
\begin{proof}
Replace each read of catalogue location $j$ by evaluation of the fixed
function $j\mapsto C_j$.  Its finite truth table can be embedded in the
program.  Simulate all other instructions unchanged.  Induction on execution
steps preserves the controller and mutable memory, hence every mutable
access and endpoint.  The fixed description now contains the catalogue
information and the simulation code.
\end{proof}

The proposition is a nonuniform statement at each finite task size.
A uniform program-size restriction across task sizes remains a substantive
restriction.  In particular, charging fixed description length $b_{\rm fix}$ prevents
catalogue information from disappearing merely by moving it into code.
When computation is free, however, a lower bound obtained by counting
literal stored windows is a bound on that storage format, not on every
implementation that selects among a finite family.  The following results
instead concern the state that experience actually changes.

\subsection{One-bit acquisition determines the index geometry}

Let $x\in\bits^N$ record which of $N$ independent evidence items have been
acquired.  Every subset is reachable, and a fresh item $i$ changes $x_i=0$
to $x_i=1$.  The native queries separate these contents.  A canonical
implementation has a complete ready state $\mathsf{enc}(x)\in\bits^s$, independent
of the order in which the acquired items arrived, and the actual fresh
update takes $\mathsf{enc}(x)$ to $\mathsf{enc}(x+\mathbf b_i)$.  Here $\mathbf b_i$ is the
$i$th unit vector and this notation is used only when $x_i=0$.

\begin{corollary}[Local indices are coordinate descriptions]
\label{loc:coordinate-index}
Suppose every fresh update in this canonical implementation changes at
most one retained bit.  There are an injection $\pi:[N]\to[s]$ and a
fixed $\xi\in\bits^s$ such that
\begin{equation}
 \mathsf{enc}(x)_{\pi(i)}=x_i\mathbin\oplus \xi_{\pi(i)},\qquad
 \mathsf{enc}(x)_j=\xi_j\quad(j\notin\operatorname{im}\pi).
 \label{loc:coordinate-index-form}
\end{equation}
For $s=N$, precisely $2^N N!$ of the $(2^N)!$ bijective encodings have
this one-bit update property.
\end{corollary}
\begin{proof}
Exact separating queries make $\mathsf{enc}$ injective.  Every undirected edge of
the semantic cube is a legal fresh update in one direction, so its two
images are distinct and have Hamming distance exactly one.  Binary
coordinate rigidity, Theorem~\ref{d:rigidity}, gives
\eqref{loc:coordinate-index-form}.  When $s=N$, choose the permutation
$\pi$ and the $N$ independent complements.  Each choice yields a different
bijection, and every displayed bijection changes exactly one bit on each
semantic edge.
\end{proof}

Thus an address satisfying this acquisition contract already contains a
separate physical coordinate for each acquired distinction.  A native
consumer can recover $x_i$ by reading cell $\pi(i)$ and applying its fixed
complement.  A fresh item can be installed by writing the known value
$1\oplus \xi_{\pi(i)}$ to that cell, with no retained-state read
($r_{\rm upd}=0$) and at most one write.  These are available
implementations; the corollary does not assume that an existing consumer
uses its shortest possible reading program.  A compound conjunction may
still require many reads, as in the central tradeoff.

Canonicality is essential to the conclusion.  It requires the same actual
ready state after the same acquired set, rather than merely the same
answers.  On histories in which each item occurs at most once, commuting
actual fresh updates provide this premise; admitting repetitions also
requires their physical idempotence to obtain the same conclusion this way.
At tight storage
$s=N$, exactness itself supplies a unique reachable state for each
content.  With surplus storage and history-dependent states, an arbitrary
choice of one representative from each content fiber need not respect the
actual updates. Section~\ref{d:robust-locality} proves a tight-storage stability theorem and connects nearly local acquisition to bounded-read availability; it also states the additional endpoint control needed for noisy runs.

\subsection{The cost of an unstructured index}

To quantify what coordinate structure contributes, consider the uniform
distribution on bijections $\mathsf{enc}:\bits^N\to\bits^N$.  This distribution
models an unstructured assignment of addresses to contents.  It is not a
distribution on learning histories.  Let $\mathsf{Edges}_N$ be the set of undirected
cube edges, let $n_{\rm addr}=2^N$, and put
\begin{equation}
 Z(\mathsf{enc})=\frac{1}{|\mathsf{Edges}_N|}\sum_{\{x,y\}\in \mathsf{Edges}_N}d_H(\mathsf{enc}(x),\mathsf{enc}(y)),\qquad
 U(\mathsf{enc})=\max_{\{x,y\}\in \mathsf{Edges}_N}d_H(\mathsf{enc}(x),\mathsf{enc}(y)).
 \label{loc:edge-costs}
\end{equation}
These are the average and worst endpoint write requirements across all
fresh-evidence transitions.  A program's actual writes can exceed them.

\begin{theorem}[Random index geometry]
\label{loc:random-index}
For a uniformly random bijection and $N\geq1$,
\begin{equation}
 \mathbb E Z=\frac{N2^{N-1}}{2^N-1},\qquad
 \operatorname{Var}(Z)\leq\frac{4N^2}{2^N}.
 \label{loc:mean-variance}
\end{equation}
Consequently, for every $\eta>0$,
\begin{equation}
 \Pr\bigl(|Z-\mathbb EZ|\geq\eta N\bigr)
 \leq 2\exp\!\left(-\frac{\eta^2 2^N}{8}\right).
 \label{loc:concentration}
\end{equation}
For $N\geq2$,
\begin{equation}
 \Pr\bigl(U<N\bigr)
 \leq\frac{2(2^N-1)}{N2^N}+\frac{2}{2^N-3}.
 \label{loc:antipodal-bound}
\end{equation}
In particular $Z/N$ converges in probability to $1/2$, and $U=N$ with
probability tending to one.
\end{theorem}
\begin{proof}
There are $n_{\rm edge}=|\mathsf{Edges}_N|=N n_{\rm addr}/2$ edges.  The images of the two ends of any fixed
edge are uniform distinct words.  Conditional on its first image, exactly
$n_{\rm addr}/2$ of the remaining $n_{\rm addr}-1$ words differ in each specified coordinate.
Summing over coordinates gives expected distance $N n_{\rm addr}/(2(n_{\rm addr}-1))$ and hence
the asserted mean.

Swapping the images of two vertices changes at most $2N$ terms in the
sum defining $Z$.  Each term changes by at most $N$, so this changes $Z$
by at most $\ell=2N^2/n_{\rm edge}=4N/n_{\rm addr}$.  Reveal the images in a fixed vertex
order and let $\mathsf{Mart}_j$ be the conditional expectation of $Z$ after $j$
images have been revealed.  Given a prefix, completions with either of
two possible next images can be coupled by swapping those two image
values.  Thus the possible conditional values of $\mathsf{Mart}_j$ lie in an
interval of width at most $\ell$.  Their conditional variance is at most
$\ell^2/4$.  Martingale differences are orthogonal in expectation:
if $i<j$, conditioning on the first $j-1$ images makes the $j$th
difference have conditional mean zero.  Therefore
\[
 \operatorname{Var}(Z)
 =\sum_{j=1}^{n_{\rm addr}}\mathbb E(\mathsf{Mart}_j-\mathsf{Mart}_{j-1})^2
 \leq n_{\rm addr}\ell^2/4=4N^2/n_{\rm addr}.
\]
The same conditional-range bound gives an exponential tail. For a
conditionally centered variable $\Delta_{\rm mart}$ with range width at most $\ell$,
Hoeffding's exponential bound is $\mathbb E(\mathrm e^{\mathsf{tilt}\,\Delta_{\rm mart}}\mid\mathcal F)
\le \mathrm e^{\mathsf{tilt}^{2}\ell^2/8}$. Iterating over the exposed images and optimizing the real parameter $\mathsf{tilt}$
gives $\Pr(|Z-\mathbb EZ|\ge v)\le 2\exp(-2v^2/(n_{\rm addr}\ell^2))$.
Taking $v=\eta N$ proves~\eqref{loc:concentration}. This is the
conditional-range form of the martingale concentration argument
(Hoeffding, 1963; Azuma, 1967).

For the worst case, the antipodal pairs of output words form a perfect
matching.  Its pullback by a uniform bijection is a uniform perfect
matching on the $n_{\rm addr}$ input vertices.  Let $X$ count cube edges in this
matching.  A specified edge is matched with probability $1/(n_{\rm addr}-1)$,
so $\mu_{\rm match}:=\mathbb EX=n_{\rm edge}/(n_{\rm addr}-1)$.  Two edges with a common endpoint
cannot both occur, and two disjoint edges both occur with probability
$1/((n_{\rm addr}-1)(n_{\rm addr}-3))$.  For $n_{\rm addr}\geq4$ it follows that
\[
 \mathbb E[X(X-1)]\leq\frac{n_{\rm edge}^2}{(n_{\rm addr}-1)(n_{\rm addr}-3)},\qquad
 \operatorname{Var}(X)\leq\mu_{\rm match}+\frac{2\mu_{\rm match}^2}{n_{\rm addr}-3}.
\]
The event $U<N$ is exactly $X=0$.  Applying Chebyshev's inequality at
deviation $\mu_{\rm match}$ gives
\[
 \Pr(X=0)\leq\frac{1}{\mu_{\rm match}}+\frac{2}{n_{\rm addr}-3}
 =\frac{2(n_{\rm addr}-1)}{N n_{\rm addr}}+\frac{2}{n_{\rm addr}-3},
\]
as required.  For $N=1$ both encodings have $U=1$.
\end{proof}

The theorem distinguishes deliberately local addresses from random
addresses even though both use the information-theoretic minimum of $N$
mutable bits.  It gives a probabilistic comparison between encodings;
it does not make random addressing an unavoidable property of selection.
The coordinate encodings of Corollary~\ref{loc:coordinate-index} are
explicit exceptions.

Write locality and query locality must also be measured separately.
For example, the bijection
\begin{equation}
 \mathsf{enc}(x)=(x_1,\ x_1\oplus x_2,\ldots,x_1\oplus x_N)
 \label{loc:reads-counterexample}
\end{equation}
changes all $N$ cells when the first semantic bit changes.  Yet a query
for $x_1$ reads one cell, and a query for any other $x_i$ reads two:
$x_i=\mathsf{enc}(x)_1\oplus \mathsf{enc}(x)_i$.  Thus a large endpoint write requirement
alone implies no comparable lower bound for a single-coordinate query.

A concrete address scheme illustrates the same distinction.  Concatenating
binary Lyndon words whose lengths divide $N$ in lexicographic order gives
the lexicographically minimal de Bruijn cycle (Fredricksen \& Maiorana,
1978).  Index each $N$-bit content by the ordinary binary encoding of its
window's offset, with offset zero at the initial $0^N$ window.  Exhaustive
edge calculations for $4\leq N\leq14$ give $U=N$ throughout, but their
means vary irregularly and are not the random-bijection means.  These
finite calculations were performed with the verification programs
described in Appendix~\ref{focus:verification}.  They establish costs of these specified encodings, not an
asymptotic law for de Bruijn indices.  Indeed, polynomial-time decoding of
positions in this cycle is available (Kociumaka, Radoszewski, \& Rytter,
2016), independently of the Hamming distance between consecutive acquired
indices.

\subsection{Charging information specific to a content family}

The distinction between a fixed family and its mutable index also admits
a description-length bound.  Fix a deterministic interpreter
$\mathsf{Dec}(\mathsf{prog},z,i)$, where $\mathsf{prog}\in\bits^{b_{\rm fix}}$ is the complete fixed description,
$z\in\bits^s$ is the complete retained state, and $i\in[L]$ is a native
query.  The public parameters $L,s,b_{\rm fix}$ and $n_{\rm fam}$ do not encode the target
family.  A supported content is an $L$-bit answer profile produced by
some $z$ on all $L$ queries.  Nonterminating or nonbinary profiles can
simply be omitted.  A family is supported if all its profiles are
supported; additional profiles are allowed.

\begin{proposition}[Fixed-description cost of a family]
\label{loc:family-count}
Among the $n_{\rm fam}$-element subsets of $\bits^L$, the fraction supported by
some $b_{\rm fix}$-bit fixed description and $s$ retained bits is at most
\begin{equation}
 \min\left\{1,\;
 \frac{2^{b_{\rm fix}}\binom{\min(2^s,2^L)}{n_{\rm fam}}}{\binom{2^L}{n_{\rm fam}}}\right\}.
 \label{loc:family-count-bound}
\end{equation}
Here $1\leq n_{\rm fam}\leq2^L$, and a binomial coefficient is zero when its upper
argument is smaller than $n_{\rm fam}$.  In particular, if $n_{\rm fam}=2^s$, the fraction
is at most $2^{b_{\rm fix}}/\binom{2^L}{n_{\rm fam}}$.
\end{proposition}
\begin{proof}
For each fixed $\mathsf{prog}$, at most $2^s$ retained states produce total answer
profiles, so its set $R_{\mathsf{prog}}$ of supported profiles has cardinality at most
$\min(2^s,2^L)$.  It supports precisely the $n_{\rm fam}$-element families included
in $R_{\mathsf{prog}}$, of which there are at most
$\binom{\min(2^s,2^L)}{n_{\rm fam}}$.  Taking the union over the $2^{b_{\rm fix}}$ descriptions
and dividing by the total number $\binom{2^L}{n_{\rm fam}}$ of families proves the
bound.  Overlap between descriptions only decreases the union.
\end{proof}

At tight retained capacity $n_{\rm fam}=2^s$, if
$b_{\rm fix}\leq\log_2\binom{2^L}{n_{\rm fam}}-k$, at most a $2^{-k}$ fraction of families
can be supported.  This is a finite counting statement about descriptions
relative to a fixed interpreter.  It identifies where family-specific
information must be charged, without turning that information requirement
into a read or update lower bound.

For example, take any two distinct $L$-bit profiles $v^0,v^1$, including
profiles with long minimal descriptions.  Put both in the fixed program
and retain one selector bit $z$.  Query $i$ reads that bit and returns
$v^z_i$.  A command specifying the next selector value writes one bit,
with no update read.  Hence $s=r_{\rm qry}=u=1$ and $r_{\rm upd}=0$ suffice,
while the profiles are paid for in $b_{\rm fix}$.  This example concerns a selection
interface that supplies the next choice; it makes no claim about the
experience needed to identify that choice.  It shows why complexity of a
family, by itself, does not entail nonlocal reading or acquisition.

The substantive distinction is therefore resource-relative.  Fixed
description bounds constrain which families can be supplied in advance.
Update bounds constrain how their experience-dependent indices can be
organized.  Read bounds constrain how those indices support the declared
uses.  For the independent intermediate contents, one-bit canonical
acquisition forces the index itself to be a coordinate description,
whereas almost every minimum-length random index requires changes to
about half its bits on average.  These conclusions compare actual
implementations while leaving their choice of the word ``selection''
immaterial.

\subsection{When computation itself is charged}
The current probe model permits free computation between charged accesses. Charging computation would introduce further constraints on a decoder or acquisition algorithm. Some strong static lower bounds are connected to established complexity problems: Dvir, Golovnev, and Weinstein (2019) derive matrix-rigidity and circuit consequences for linear data structures, while Corrigan-Gibbs and Kogan (2019) prove related barriers for systematic data structures and function inversion with preprocessing. Their conclusions depend on the relevant models and parameter regimes. Transferring either barrier to the present dynamic learning task would require a reduction preserving those restrictions. These results motivate stating the decoder model precisely; they do not imply that unconditional bounds for every computational acquisition model are unavailable. The bounds proved here use the explicit bit-access and endpoint constraints and require no such complexity assumption.

\section{Relational learning and the cost of finding what to update}
\label{focus:partition-cost}

The central task isolates the number of changed cells. Relational learning illustrates why that quantity must remain distinct from the work needed to locate those cells. Let $[n_{\rm obj}]$, $n_{\rm obj}\geq2$, be public object identifiers. An example $(i,j)$ asserts that the two objects belong to the same class. The acquired content $\mathcal P_e$ is the least equivalence relation containing the observed pairs, and the native query $\operatorname{Same}(i,j)$ asks whether $i\mathrel{\mathcal P_e}j$. This is warranted equivalence from the available evidence: a negative answer does not assert that two objects belong to different classes in an incompletely identified target partition. Every history is consistent with at least the one-class target.

\begin{proposition}[Few writes can require substantial update reads]
\label{focus:partition-writes}
The relational task requires at least $\lceil\log_2 B_{n_{\rm obj}}\rceil$ retained bits, where $B_{n_{\rm obj}}$ is the $n_{\rm obj}$th Bell number. With $w=\lceil\log_2(n_{\rm obj}+1)\rceil$, it admits an exact implementation using $2n_{\rm obj}w$ retained bits in which a query reads at most $2(\lfloor\log_2n_{\rm obj}\rfloor+1)$ words and writes none. An example changes at most two words, after at most $2\lfloor\log_2n_{\rm obj}\rfloor+4$ word reads. Transient workspace is $O(\log n_{\rm obj})$ bits.
\end{proposition}
\begin{proof}
Every partition can be acquired by supplying a spanning tree of pairs within each block. Distinct partitions disagree on a pair query, proving the counting lower bound. For the construction, keep a parent and a size word per object, initially a self-parent and size one. Follow parent words without path compression to find a root. For a new pair, find both roots; if they differ, attach the smaller component's root to the larger component's root and add the smaller size to the larger size. Ties are broken by public indices. Only the attached root's parent and the surviving root's size change. Nonroot size fields need not be cleared.

Initially root equality is equality of objects. Each example merges exactly the two indicated equivalence blocks, so induction gives root equality exactly when $i\mathrel{\mathcal P_e}j$. Whenever a node gains a parent edge above it, the size of its containing component at least doubles. Its depth is therefore at most $\lfloor\log_2n_{\rm obj}\rfloor$. Two read-only root searches give the query bound; reading the two root sizes adds two reads to an update. The arrays use $2n_{\rm obj}w$ bits, and a constant number of index and size registers suffices as workspace. Two changed words change at most $2w$ physical bits.
\end{proof}

The evidence interface is exactly incremental connectivity: replace each example by insertion of an undirected edge, and each native query by a connectivity query. This translation preserves operation sequences and their memory accesses. The relevant operation is \emph{linking arbitrary objects}, not \emph{union of supplied representatives}. An update must discover the relevant components unless an additional interface supplies their representatives. The elementary weighted-union implementation above deliberately retains its separate two-write guarantee; replacing it by a faster algorithm does not automatically preserve that guarantee.

For comparison, let $c_{\rm upd}$ and $c_{\rm qry}$ count \emph{all} cell probes in a worst-case arbitrary-link update and Boolean connectivity query, with words of $O(\log n_{\rm obj})$ bits. Blum's (1986) structure gives a balanced $O(\log n_{\rm obj}/\log\log n_{\rm obj})$ worst-case operation bound. For representative-returning $\operatorname{Find}$ queries with $\operatorname{Union}$ on supplied representatives, the optimal worst-case tradeoff is $c_{\rm find}=\Theta(\log n_{\rm obj}/\log(2+c_{\rm union}))$, with the usual lower bound of one operation and parameters restricted to the nontrivial range (Alstrup, Ben-Amram, \& Rauhe, 1999). Two finds implement a Boolean connectivity query, and two finds followed by a union implement an arbitrary link. These reductions give the upper bounds $c_{\rm qry}=O(c_{\rm find})$ and $c_{\rm upd}=O(c_{\rm union}+c_{\rm find})$. Thus the efficient upper tradeoff transfers when union time is at least of the order of find time; the fast-union endpoint does not transfer unchanged. The separate lower bound for Boolean queries under arbitrary links is the following result.

Indeed, P\u{a}tra\c{s}cu and Thorup (2011, Theorem~1) prove that incremental connectivity with worst-case $c_{\rm upd}=o(\log n_{\rm obj}/\log\log n_{\rm obj})$ has worst-case $c_{\rm qry}\geq n_{\rm obj}^{1-o(1)}$ in that cell-probe model. These are costs of the whole update, including reads. They do not give the same lower bound with $c_{\rm upd}$ replaced by the number of writes: Proposition~\ref{focus:partition-writes} has only two word writes but logarithmically many update reads. Nor is this a claim about amortized operation costs, which obey different bounds. The exact deterministic case suffices for the present comparison; no bounded-error extension is inferred here.

Thus relational learning adds a precisely located qualification to the central result. The same-class answers can change for many pairs while only two stored words change, because traversal performs the remaining work during both acquisition and use. Conversely, constraining the \emph{total} effort spent incorporating evidence can sharply increase later use costs. These established data-structure results contribute a second learning interface and a warning against identifying write locality with fast acquisition; they are not a second complete curve for the paper's separate storage, read, and changed-cell budgets.

\section{Keeping old content while installing new content}
\label{focus:retention}

The read--write question changes when an update must preserve an acquired
resource. It is no longer enough to count all nearby physical states:
only states preserving the old native answers are admissible. The cost of
reading those old answers determines the geometry of this admissible region.
The following results make that dependence explicit. They concern installing
a new resource after the evidence has identified its intended content; they
do not replace the evidence requirements of learning. In the installation account of Wu (2026b), a learned description must enter an actual consumer; here the added question is which new descriptions can be installed under a write budget while the earlier consumer remains correct.

Let the old content be $x\in\bits^{k_{\rm old}}$, and let ready states occupy $s=k_{\rm old}+s_{\rm free}$
binary cells. All $2^{k_{\rm old}}$ old contents must be realizable. The fixed native
query for $x_i$ is deterministic and read-only; its program and address
calculation have no uncounted target-dependent inputs. Additional new
contents are labels in a finite set $\mathcal L$, with $n_{\rm new}=|\mathcal L|$, decoded by a fixed
installed program. Distinct labels must require different native answer
profiles. An installation is successful when its endpoint decodes the
requested new label and preserves every old answer. The write budget $u$
bounds the number of retained cells changed between ready endpoints;
at least that many actual writes are necessary. Constructions below attain
their endpoint bounds without writing any cell twice.

\begin{theorem}[Sharp local installation capacity with one-read retention]
\label{focus:retention-local}
Suppose each old-bit query uses at most one bit read on every ready state,
including states reached after installation. Then $k_{\rm old}$ distinct physical
cells carry the old bits, each up to a fixed complementation. From any
specified ready state, installation preserving the old content and changing
at most $u$ cells can support at most
\begin{equation}
 n_{\rm new}\le L(s_{\rm free},u):=\sum_{j=0}^{\min(u,s_{\rm free})}\binom{s_{\rm free}}{j}.
 \label{focus:retention-ball}
\end{equation}
The bound is sharp across all encodings: a fixed decoder and a uniform
installation algorithm attain it from a blank $s_{\rm free}$-bit suffix, for every
old content.
\end{theorem}
\begin{proof}
Corollary~\ref{d:nativebitcarrier} applies to the fixed old-bit readers over all reachable ready states, including the installation endpoints. It yields $k_{\rm old}$ distinct carriers, each with a fixed identity or complement decoder. Preserving the old content therefore fixes these physical cells.

Only the remaining $s_{\rm free}$ cells can change. At most $L(s_{\rm free},u)$ suffixes are
within distance $u$ of the current suffix. Different new labels require
different endpoints because their decoder is fixed. This proves the bound
without requiring canonical states or imposing any restriction on surplus
states within an old-content fiber.

For attainment, store $x$ in the first $k_{\rm old}$ cells and initialize the remaining
cells to zero. Enumerate the suffix words of weight at most $u$ in a fixed
order and assign one new label to each. On receiving label $z_{\rm new}$, compute its
word and set exactly its one positions. The old prefix is untouched and
at most $u$ bits change. The native new-label decoder reads the suffix and
computes its enumeration index. These programs are uniform in $s_{\rm free},u$;
an exponential catalogue is unnecessary. A simple exhaustive implementation
uses $s_{\rm free}$ suffix reads and $O(s_{\rm free}+\log(n_{\rm new}+1))$ temporary bits, with potentially
exponential computation. Temporary storage is charged to peak storage and
cleared before the state becomes ready.
\end{proof}

The theorem strengthens a reserved-suffix construction into an optimality
result over all encodings meeting the old read contract. Its premises do
not say that a system happens to retain one fixed old content: the same
installed readers must support every $x\in\bits^{k_{\rm old}}$. If an old bit were
fixed throughout the entire task, its reader could return a constant and
would not force a carrier. Likewise, a free target-dependent routing
register would violate the declared state boundary.

\begin{example}[Paying for old reads increases installation capacity]
\label{focus:retention-majority}
Take one old bit, three retained bits, and a one-write installation budget.
One-read access to the old bit permits at most $L(2,1)=3$ new labels.
With three-read access, four labels are possible. Decode the old bit as
the majority of the three cells. Decode new label $0$ when all cells agree,
and label $i\in\{1,2,3\}$ when cell $i$ is the unique minority. Use $000$
as the initial old-zero state and $111$ as the initial old-one state.
From either initial state, doing nothing or flipping the designated cell
installs any of the four new labels. All eight states have a unique old
bit and new label. Three reads suffice for either decoder, and one write
suffices for installation. The larger capacity is purchased by a more
expensive old-content reader, not by extra retained storage.
\end{example}

The example is deliberately a one-shot result. From $001$, changing the
minority position to the second cell while keeping majority zero requires
two writes. Supporting one installation from a blank state need not support
arbitrary repeated replacements at the same cost.

\subsection{Repeated installation and perfect-code partitions}

Call a system \emph{recurrently installable} for $\mathcal L$ if, from every ready
state reachable after any finite sequence of requests, every next label
$z_{\rm new}\in \mathcal L$ can be installed while preserving the old content. Repeating the
current label is permitted and may require zero writes. No reset or fresh
blank suffix is supplied between requests. The query and update programs
remain fixed, and any retained phase counter belongs to the $s$ cells.

\begin{theorem}[When the one-write capacity remains attainable forever]
\label{focus:retention-recurrent}
Under the one-read old-content contract of
Theorem~\ref{focus:retention-local}, let $s_{\rm free}\ge1$. Recurrent installation
of $n_{\rm new}=s_{\rm free}+1$ new labels with at most one changed bit per request is possible
if and only if $s_{\rm free}+1$ is a power of two. At this capacity, for each fixed
old content the new-label fibers partition the entire residual cube
$\bits^{s_{\rm free}}$ into binary $1$-perfect codes. A Hamming-syndrome construction
attains the bound using at most $s_{\rm free}+1$ suffix reads and one write per update.
\end{theorem}
\begin{proof}
Fix any reachable ready state and its old content. The preceding theorem
fixes the $k_{\rm old}$ carrier cells. From any reachable suffix $z$, the possible
one-write endpoints lie in the closed ball $B_1(z)$, which has $s_{\rm free}+1$
elements. Successful installation of $s_{\rm free}+1$ distinct labels requires that
every element of this ball be a reachable ready endpoint, with exactly
one point of each label. In particular every neighbor of $z$ is reachable.
By connectivity of the binary cube, iteration reaches every suffix.
This proves the full-cube conclusion from a contract stated only on
reachable states; no additional correctness assumption on initially
unreachable words has been imposed.

Let $H^{\rm new}_{z_{\rm new}}$ be the suffixes carrying label $z_{\rm new}$. Every $B_1(z)$ meets $H^{\rm new}_{z_{\rm new}}$
once. Equivalently, the radius-one balls centered at $H^{\rm new}_{z_{\rm new}}$ partition the
cube: membership is symmetric in Hamming distance, so every cube point
belongs to exactly one such ball. Thus
\[
 |H^{\rm new}_{z_{\rm new}}|(s_{\rm free}+1)=2^{s_{\rm free}}.
\]
Consequently $s_{\rm free}+1$ divides $2^{s_{\rm free}}$, and hence is a power of two. Each $H^{\rm new}_{z_{\rm new}}$
is a $1$-perfect code. The argument does not assert that these codes are
linear.

Conversely, write $s_{\rm free}=2^{r_{\rm syn}}-1$, and let the columns of an $r_{\rm syn}\times s_{\rm free}$ binary
matrix $\mathsf H$ be all distinct nonzero $r_{\rm syn}$-bit vectors. Label a suffix $z$ by
the syndrome $\mathsf Hz$, with arithmetic over $\mathbb F_2$. Given requested
label $z_{\rm new}$, compute $\mathsf{synDiff}=z_{\rm new}+\mathsf Hz$. If $\mathsf{synDiff}=0$, leave the suffix unchanged.
Otherwise flip the unique coordinate whose column of $\mathsf H$ is $\mathsf{synDiff}$.
The resulting syndrome is $\mathsf Hz+\mathsf{synDiff}=z_{\rm new}$. The prefix never changes, and the
procedure works from every suffix, so induction establishes recurrent
installation for every finite request sequence. Computing $\mathsf Hz$ requires
at most $s_{\rm free}$ reads. If a bit must change, read that selected coordinate
once more and overwrite it with its complement. This ordinary overwrite
implementation uses at most $s_{\rm free}+1$ reads; no atomic-flip primitive is
assumed. The columns can be generated from their indices, using
$O(r_{\rm syn}+\log(s_{\rm free}+1))$ temporary bits, and exactly zero or one retained bit is
written. The new-label reader uses $s_{\rm free}$ reads for the syndrome computation;
old-bit readers still use one read.
\end{proof}

This is a coding-theoretic characterization applied to preservation and
new acquisition. Zhang and Li (2005, Definition~1 and Theorems~14--15)
formulate bounded-change embedding with a fixed decoder and relate
saturated embedding partitions to perfect codes. Their Example~1 gives
the syndrome construction used in matrix embedding; Westfeld's (2001)
F5 algorithm is an established application. The contribution here is the
reduction from the old native read contract to protected carrier cells,
and the consequence that saturating a recurrent installation contract on
reachable states forces the entire remaining cube. Neither perfect codes
nor the syndrome algorithm is claimed as new.

For $s_{\rm free}=2$, one-shot installation supports three labels, but recurrent
one-write installation cannot support three because $3$ does not divide
$4$. Two recurrent labels are attainable by using one suffix bit and
leaving the other unused. For $s_{\rm free}=3$, four labels are recurrently attainable.
These are distinctions between capabilities under the same ready-storage
and write budgets. They do not imply that a non-saturating recurrent code
must be perfect. Nor does the theorem give the optimum recurrent capacity
for every $s_{\rm free}$ and every write budget.

\subsection{Capacity and irreversible loss}

The local results supplement the total-capacity constraints in
Theorem~\ref{d:retentionentropy}. Equation~\eqref{d:oldnewentropy} bounds
simultaneous approximate retention of independent old and new bits without
requiring one-probe consumers. Those information bounds do not specify
which new contents are reachable within a particular update budget.
Conversely, a store can have enough total states while its current
old-content fiber contains too few locally reachable new states.

Adding empty storage also cannot recover a distinction that a previous
compression has erased. The following elementary fact separates capacity
for future acquisition from recovery of past evidence.

\begin{proposition}[Blank storage cannot undo a collision]
\label{focus:retention-collision}
Let $X$ be an evidence-history set, $f:X\to Q$ its retained-state map,
and $g:X\to Z$ a desired recovered content, where $Z$ is nonempty.
A recovery function $R:Q\to Z$ satisfying $R\circ f=g$ exists exactly
when $g$ is constant on each fiber of $f$. If $f(x)=f(y)$ but
$g(x)\ne g(y)$, no recovery based on that retained state and fixed blank
ancillary storage succeeds on both histories. Randomized recovery has
worst-case error at least $1/2$ on the pair.
\end{proposition}
\begin{proof}
If $R\circ f=g$, equality of retained states entails equality of outputs.
Conversely, fiber constancy defines $R$ on the image of $f$; choose any
element of $Z$ elsewhere. Appending the same blank ancillary state leaves
the collision intact. A randomized recovery receives the same input on
$x$ and $y$, so it has the same output distribution. The probabilities
assigned to the two distinct correct contents sum to at most one; at
least one success probability is therefore at most $1/2$.
\end{proof}

The local, recurrent, and collision results identify different costs of
preserving a usable resource. One-read access fixes physical carriers and
limits the admissible update neighborhood; recurrence constrains how all
such neighborhoods fit together; irreversible compression removes
distinctions that larger blank memory alone cannot restore. All three
claims concern the declared readers, retained state, and permitted writes.
They do not predict a task-independent amount of catastrophic forgetting.

\subsection{Shared old-content fibers}
\label{shared:fibers}

A distributed old-content decoder need not protect a fixed set of cells.
Let $g:\bits^s\to\bits^{k_{\rm old}}$ be the installed old decoder, assumed surjective,
and let $H_x^{\rm old}=g^{-1}(x)$. Each of its $k_{\rm old}$ native bit queries uses at most
$r_{\rm old}$ reads. All binary words are admissible ready words in this
subsection. With a restricted ready set, the endpoint and recurrence
characterizations below use its intersection with each neighborhood. The
certificate lower bound additionally requires that the indicated
certificate subcube remain admissible; otherwise only its intersection
with the ready set may be counted. Queries are read-only and occur at ready
boundaries. Retention is required between completed updates, rather than
after each internal write.

Put
\[
 N_u^g(q)=H_{g(q)}^{\rm old}\cap B_u(q),\qquad
 B_u(q)=\{z:d_H(q,z)\le u\}.
\]
A fixed new decoder $h:\bits^s\to[n_{\rm new}]$ assigns the installed content label.
Different labels have different native answer profiles. The decoder and the
update program have no free target-dependent inputs. Fix one initial ready
word $q_x\in H_x^{\rm old}$ for every old content $x$. The one-shot optimum below is
for this one-start-per-old-fiber contract. If several starting words share
an old fiber, one fixed new decoder must label all their neighborhoods
simultaneously; the smallest neighborhood size is then an upper bound and
need not be attainable.

\begin{lemma}[Feasibility of installation and recurrence]
\label{shared:fiber-theorem}
The following statements hold for finite $s$ and nonnegative integer $u$.
\begin{enumerate}[label=(\roman*)]
\item For fixed $g,h$, all labels can be installed from $q$ while preserving
old content and changing at most $u$ bits exactly when
\[
 h(N_u^g(q))=[n_{\rm new}].
\]
Allowing the fixed new decoder and installer to be designed, the largest
one-shot label set supported from every $q_x$ is
\begin{equation}
 n_{\rm new,one}^*(g,(q_x),u)=\min_x |N_u^g(q_x)|.
 \label{shared:one-shot}
\end{equation}
\item A fixed $g,h$ supports arbitrary finite sequences of installations
from the $q_x$ exactly when there are sets $R_x\subseteq H_x^{\rm old}$ containing
$q_x$ such that
\begin{equation}
 h\bigl(R_x\cap B_u(q)\bigr)=[n_{\rm new}]
 \qquad(x\in\bits^{k_{\rm old}},\ q\in R_x).
 \label{shared:recurrent-cover}
\end{equation}
Consequently every recurrent realization satisfies
\[
 n_{\rm new}\le\min_x\min_{q\in R_x}|R_x\cap B_u(q)|,
 \qquad n_{\rm new}\le\min_x|H_x^{\rm old}|\le2^{s-k_{\rm old}}.
\]
\end{enumerate}
\end{lemma}
\begin{proof}
A label is installable precisely when a corresponding endpoint belongs to
the indicated neighborhood. Distinct labels need distinct endpoints. For
one-shot attainment choose $n_{\rm new}=\min_x|N_u^g(q_x)|$ endpoints in each old fiber
and give them the $n_{\rm new}$ distinct new labels. Different old fibers are disjoint,
so these assignments define one decoder; assign arbitrary labels elsewhere.
An installer reads its initial word, selects the assigned endpoint, and
overwrites exactly the differing positions. It uses at most $s$ reads and
$u$ writes. This proves (i), without an efficiency claim for the finite
selection computation or its fixed description. Any lookup information used
by this existential construction is charged to the fixed program; it is
not assumed to have a compact description.

For (ii), take $R_x$ to be all states actually reachable under finite
installation sequences. Every next request has an endpoint in $R_x$ within
the permitted distance, proving necessity. Conversely, at each $q\in R_x$
and requested label choose one witnessing endpoint in
\eqref{shared:recurrent-cover}. Reading the word and overwriting its differing
positions implements this choice. Induction proves that every request
remains in $R_x$ and is correctly installed. The neighborhood bound follows
by counting labels. Every $R_x$ contains all $n_{\rm new}$ labels, whence
$|H_x^{\rm old}|\ge n_{\rm new}$; summing the disjoint fiber sizes gives $2^{k_{\rm old}} n_{\rm new}\le2^s$.
\end{proof}

The feasibility lemma states which endpoints an installer must reach.
The next result bounds their number through the old reader's physical
access and the sensitivity of its decoder.

\begin{proposition}[Old-read certificates and installation neighborhoods]
\label{shared:fiber-geometry}
Let all words in $\bits^s$ be admissible, and let $c_g(q)$ be the minimum
number of physical coordinates whose values at $q$ force the entire old
output $g(q)$. For every nonnegative integer $u$,
\begin{equation}
 \sum_{j=0}^{\min(u,s-c_g(q))}\binom{s-c_g(q)}j
 \le |N_u^g(q)|\le\sum_{j=0}^{\min(u,s)}\binom sj,
 \qquad c_g(q)\le\min\{s,k_{\rm old}r_{\rm old}\}.
 \label{shared:certificate-neighborhood}
\end{equation}
For $u=1$, the exact neighborhood size is
\begin{equation}
 |N_1^g(q)|=1+s-\operatorname{sens}_g(q),\qquad
 \operatorname{sens}_g(q)=|\{i:g(q\mathbin\oplus\mathbf b_i)\ne g(q)\}|.
 \label{shared:sensitivity}
\end{equation}
\end{proposition}
\begin{proof}
Fixing a minimum certificate leaves a subcube of dimension
$s-c_g(q)$ inside $H_{g(q)}^{\rm old}$. Its radius-$u$ neighborhood gives the lower
bound. Follow each old-bit reader's actual path from $q$. Fixing the union
of their probed coordinates reproduces all those paths, including their
adaptive choices, and hence all old outputs. This union has at most
$k_{\rm old}r_{\rm old}$ coordinates. The upper bound counts the entire ball. At
radius one, exactly the sensitive neighbors leave the old fiber, giving
\eqref{shared:sensitivity}.
\end{proof}

The recurrence condition is a covering requirement throughout the states
that installations themselves can reach: every new-label class must meet
every permitted neighborhood. Theorem~\ref{focus:retention-local} is the
one-probe case, where the old-reader contract forces fixed carriers and
leaves an ordinary residual cube. At saturated one-write capacity,
Theorem~\ref{focus:retention-recurrent} strengthens covering to perfect-code
partitioning. For general distributed readers, local sensitivity and the
shape of the fibers replace the assumption of a reserved suffix.

These geometric quantities are not determined by the old read bound alone.
For three-bit majority, a unanimous initial word has four old-preserving
radius-one endpoints, as in Example~\ref{focus:retention-majority}. Each old
fiber is a three-leaf star. Its largest recurrent one-write label set has
two labels: any set with at least three states includes a leaf, whose closed
neighborhood has only two states; conversely, labeling the center $0$ and
all leaves $1$ works. Three-bit parity also requires three reads to read
its old bit, but every one-bit change alters it, so its radius-one capacity
is one. Extra old-read cost can enlarge an installation neighborhood, but
it does not guarantee that enlargement or its recurrent availability.

\subsection{Reading enough to preserve the old resource}
\label{shared:update-reads}

Endpoint distance measures how much must change. It does not say how the
installer finds a preserving change. For this subsection, an ordinary write
is $\operatorname{set}(i,\mathsf{val})$, with $\mathsf{val}$ determined by the request, the program,
and values already read. It does not return the overwritten value. An
optional blind-flip instruction replaces a bit by its complement without
reading it; this is a stronger primitive. In neither model can an instruction
copy an unread bit into another cell. All address selection and computation
use only the request and the acquired read transcript.

Fix a request $z_{\rm new}$, an allowed starting set $\Omega$, and a valid-endpoint
relation $\mathsf{Target}_{z_{\rm new}}(q)$. For installation it is
$\{z:g(z)=g(q),\ h(z)=z_{\rm new}\}$, with any required ready-state restriction added.
An overwrite action $\omega=(W,\mathsf{val})$ sets the distinct cells in $W$ to the
specified constants and leaves all other cells unchanged. Write
$\omega(q)$ for its endpoint. Here $u$ bounds actual write instructions,
which also bounds changed cells. Define
\[
 \mathcal A_{z_{\rm new}}(q)=\{\omega:|W|\le u,\ \omega(q)\in\mathsf{Target}_{z_{\rm new}}(q)\}.
\]

\begin{lemma}[Read transcripts select a common valid action]
\label{shared:action-tree}
There is a deterministic ordinary-overwrite updater using at most
$r_{\rm upd}$ reads and $u$ writes from every $q\in\Omega$ exactly when a
binary read decision tree of depth at most $r_{\rm upd}$ has, at every leaf
$L$ reached by some allowed input, a nonempty intersection
\begin{equation}
 \bigcap_{q\in\Omega_L}\mathcal A_{z_{\rm new}}(q)\ne\varnothing,
 \label{shared:action-intersection}
\end{equation}
where $\Omega_L$ is the set of allowed inputs consistent with that leaf's
read transcript. The characterization also covers interleaved reads and
writes when temporary workspace is available and charged: deferring writes
requires at most $O(u\log(s+1))$ extra working bits.
\end{lemma}
\begin{proof}
For a read-then-write updater, one leaf fixes its entire write action.
Correctness on every compatible input is exactly
\eqref{shared:action-intersection}. Conversely, choose one common action
at each nonempty leaf and execute it after the reads.

To handle an interleaved updater, simulate its writes in a temporary map
from written addresses to their most recent values. A simulated read of
such an address uses that known value; every other read probes the original
store. At termination, perform only the last write to each distinct written
address. The simulated control path and final word are unchanged, while
physical reads and writes do not increase. The map uses the stated working
space. This gives the required read decision tree and one fixed overwrite
action at each leaf.
\end{proof}

\begin{corollary}[Zero-read overwrites separate old content from written cells]
\label{shared:zero-read-separation}
Let $g:\bits^s\to\bits^{k_{\rm old}}$ be the installed old decoder. For each request
$z_{\rm new}$ in a finite set, suppose a deterministic zero-read ordinary-overwrite
updater $\omega_{z_{\rm new}}$ maps a common nonempty ready set
$\Omega\subseteq\bits^s$ into itself and preserves $g(q)$ for every
$q\in\Omega$. Let $W_{z_{\rm new}}$ be the
coordinates to which its final overwrite action assigns constants, and
put $W_*=\bigcup_{z_{\rm new}} W_{z_{\rm new}}$. Write
$Z=\{q|_{[s]\setminus W_*}:q\in\Omega\}$.
Then there is a function
\[
 \bar g:Z\longrightarrow\bits^{k_{\rm old}},
 \qquad g(q)=\bar g\bigl(q|_{[s]\setminus W_*}\bigr)\quad(q\in\Omega).
\]
Thus the old content is recoverable without the cells written by any of these
requests on the common forward-invariant ready set $\Omega$. This is a
factorization of $g|_\Omega$, not a claim that $g$ ignores those coordinates
on the whole physical cube. If
$g(\Omega)=\bits^{k_{\rm old}}$, then
\[
 2^{k_{\rm old}}\le |Z|\le2^{s-|W_*|},
 \qquad k_{\rm old}\le s-|W_*|.
\]
Conversely, this factorization makes every ordinary overwrite supported in
$W_*$ that maps $\Omega$ into itself preserve the old content.

The full cube and cylinders
$\Omega=\{q:q|_{[s]\setminus W_*}\in Z\}$ are special cases in which
overwrites supported in $W_*$ automatically preserve the ready set.
\end{corollary}
\begin{proof}
A zero-read updater has one fixed overwrite action for each request.
Compose these actions once each in a fixed order. The composition leaves
every coordinate outside $W_*$ unchanged and sets every coordinate in
$W_*$ to a fixed constant, namely the value supplied by the last action
that writes it. Any two starting words agreeing outside $W_*$ therefore
have the same endpoint under this composition. Forward invariance of
$\Omega$ makes every intermediate state an admissible input to the next
action. Thus $g$ is preserved at every ready endpoint, and the common final
endpoint gives $g(q)=g(q')$. This proves the factorization. Surjectivity
on $\Omega$ and counting its distinct remaining-coordinate strings give
the inequalities. The converse follows because a supported overwrite
leaves those strings unchanged and, by hypothesis, ends in $\Omega$.
\end{proof}

The factorization is relative to the ready domain. For example, let
$\Omega=\{00,10\}$ and $g(q)=q_1\oplus q_2$. Overwriting $q_2$ by zero
preserves both $\Omega$ and $g|_\Omega$, which is simply the first-coordinate
projection. The decoder $g$ on the whole cube nevertheless depends on $q_2$.

The domain condition matters. Let $g$ be three-bit majority, and allow
only $000$ and $111$ as starting words. Each of the three zero-read actions
that sets one coordinate to zero preserves $g$ on this starting set.
Together they write every coordinate, although the two old contents remain
distinct. Their endpoints need not lie in the allowed starting set, so
this one-shot preservation promise cannot justify successive applications
of the argument. Nor does the corollary apply to all updaters with short
read transcripts: those reads can select different actions for different
old states. Ordinary overwrites are also essential; two blind flips
preserve two-bit parity, which depends on both changed cells.

\begin{proposition}[A sharp cost of preserving a distributed old bit]
\label{shared:parity-update-reads}
Let $s\ge2$, let the old decoder be
$g(q)=q_1\oplus\cdots\oplus q_s$, and let the new content be
$h(q)=(q_1,\ldots,q_{s-1})$. All $2^s$ words are allowed starting states.
With write budget $u=s$, installing every requested
$z_{\rm new}\in\bits^{s-1}$ while preserving the old bit has optimal worst-case
update-read cost
\[
 r_{\rm upd}=s\quad\text{with ordinary overwrites},\qquad
 r_{\rm upd}=s-1\quad\text{when blind flips are also allowed}.
\]
The old bit itself requires $r_{\rm old}=s$ reads. In contrast, decoding
the old bit as $q_s$ allows installation of the same new payload with zero
reads and at most $s-1$ writes.
\end{proposition}
\begin{proof}
Fix $z_{\rm new}$. Every payload coordinate $i<s$ must be read. Otherwise two inputs
differing only there follow the same transcript. The requested payload
forces that difference to disappear, while every other endpoint coordinate
is identical for the pair. Their old parities differ, so both cannot be
preserved. This argument also allows blind flips: a fixed flip cannot set
both possible input values to the same prescribed payload bit.

Choose a payload whose parity differs from that of $z_{\rm new}$. Preserving the old
parity requires complementing the last cell. If it is unread, the two inputs
with that same payload and opposite last bits have the same transcript.
An ordinary final action either leaves the last cell unchanged or sets it
to a fixed value; neither complements both inputs. Some such execution must
therefore read all $s$ cells. Reading the whole word, computing its parity,
and writing the unique target endpoint attains this bound.

With blind flips, read the $s-1$ payload cells, overwrite them with $z_{\rm new}$, and
flip the final cell exactly when the payload parity changes. This uses at
most $s$ writes and preserves the old parity. The payload lower bound makes
its read cost sharp. Finally, parity requires all $s$ reads because an
unread bit could change its value. The projection decoder keeps its last
cell untouched while blindly overwriting the payload.
\end{proof}

The extra read is an operational requirement of ordinary overwrite memory.
It would be incorrect to infer it solely from a Hamming-distance argument
or to retain it after granting an unread-bit flip instruction.

\subsection{New evidence while earlier content remains available}
\label{shared:sequential-learning}

The installation results take the intended new label as an input. A learner
instead receives evidence that incrementally determines its resource.
Consider the $N$ independent exclusions supplied by the truthful
negative-example reduction in Theorem~\ref{focus:central}. Let
$v\in\bits^N$ record the exclusions observed so far, starting at $0^N$;
evidence $i$ requires $v\leftarrow v\lor\mathbf b_i$. Evidence may repeat, and no
freshness flag is supplied. Require the old $k_{\rm old}$ bits to remain readable and
the native new consumer to answer
$\operatorname{None}(S_{\rm req})=\prod_{i\in S_{\rm req}}v_i$ after every evidence prefix.
The reachable ready states may form a proper subset of the physical cube.
Since every evidence item may follow every history, however, their common
set is forward invariant under every evidence update. A freshness promise
can instead produce request-dependent starting sets and needs a separate
check of this invariance.

\begin{theorem}[Shared capacity and sequential evidence obligations]
\label{shared:learning-capacity}
Suppose every old content $x\in\bits^{k_{\rm old}}$ has an allowed initial state and
every exclusion word is admissible from that state. Then:
\begin{enumerate}[label=(\roman*)]
\item Every correct learner needs at least $2^{k_{\rm old}+N}$ distinct ready states,
so $s\ge k_{\rm old}+N$. For each fixed old content, its fiber contains at least
$2^N$ attainable states. At every reachable state $q$ with new content $v$,
its actual evidence transition must satisfy
\[
 g(\delta_i(q))=g(q),\qquad v(\delta_i(q))=v(q)\lor\mathbf b_i.
\]
\item If each old-bit query uses one read and each compound new query
uses at most $r_{\rm qry}$ reads, then
\begin{equation}
 \sum_{j=0}^{\min(r_{\rm qry},s-k_{\rm old})}\binom{s-k_{\rm old}}{j}\ge2^N.
 \label{shared:old-new-dimension}
\end{equation}
Thus the storage--read lower bound of Theorem~\ref{focus:central} applies
to the $s-k_{\rm old}$ unprotected cells. Block caches in these cells, together with
$k_{\rm old}$ direct old carriers, attain the corresponding upper bounds.
\end{enumerate}
\end{theorem}
\begin{proof}
Choose one actual evidence history for each $v$ from each old initial
content $x$. Different pairs $(x,v)$ have different native answers: an old
bit separates distinct $x$, and a singleton new query separates distinct
$v$. Their endpoints are therefore distinct. This proves both counts.
Correctness after appending evidence $i$ gives the two transition equations.

For (ii), the one-probe carrier lemma fixes $k_{\rm old}$ distinct old cells, up to
fixed complements, on all ready states. Fix $x$. Its old carriers are
constants, so any new query that probes them can replace those probes by
the known constants without increasing its remaining depth. The resulting
read decision trees access at most $r_{\rm qry}$ of the other $s-k_{\rm old}$ cells. The
bounded-probe span lemma (Lemma~\ref{d:probe-span}) therefore bounds the dimension of their functions
by the left side of \eqref{shared:old-new-dimension}. On the chosen
$2^N$ attained new-content states, the $2^N$ conjunction functions are
linearly independent, exactly as in Theorem~\ref{focus:central}. This gives
the inequality. The block-cache implementation modifies only its allocated
new cells, so every old answer remains correct throughout its completed
updates.
\end{proof}

If all evidence updaters use zero reads and ordinary overwrites,
Corollary~\ref{shared:zero-read-separation} therefore applies to this
reachable ready set. Let $W_*$ be the union of their written coordinates.
The old content must be recoverable from its complement. From any fixed
old initial state, only coordinates in $W_*$ can change, while the $2^N$
new-content states must all be reachable and distinct. Hence
\[
 N\le |W_*|\le s-k_{\rm old}.
\]
This conclusion identifies a set of cells from which the old content is
decodable and which acquisition never overwrites. It imposes no one-cell
carrier for each old bit, and it does not apply to arbitrary positive-read
updaters.

The capacity bound is attained by a direct old prefix and the $N$-bit
exclusion indicator. Evidence $i$ blindly sets its indicator to one, using
no reads and one overwrite, including repeated evidence. A shared encoding
with one old bit instead uses
\[
 q=(v_1,\ldots,v_N,\ x\oplus v_1\oplus\cdots\oplus v_N).
\]
Its old reader takes parity of all $N+1$ cells, while its new readers use the
payload. With repetitions permitted, evidence $i$ first reads $v_i$; if it
is zero, the updater reads and complements the final cell and sets $v_i=1$.
The worst case uses two reads and two writes, and order and repetition do
not alter the resulting code. The two-read bound is necessary for this
fixed code and ordinary writes: the unread $i$th bit would leave repeat and
first-insertion cases indistinguishable, and a first insertion must also
complement an initially unknown final bit. If the protocol promises a
fresh exclusion, $v_i=0$ is already known and only the final-cell read is
needed; allowing a blind flip removes that read as well. Consequently the
cost of a preserving update depends both on the shared encoding and on the
information supplied by the evidence interface.

\section{Stored information and bounded access}
\label{d:availability}

Information capacity and access cost can be compared within the same probability model. The comparison requires a declared target: information about one required answer differs from information about an arbitrary function of the stored state. Fix a distribution of legal acquisition histories, let $\mathbf X$ denote the resulting target, and let $\mathbf Q\in\bits^s$ be the complete ready state, including any retained acquisition randomness. A finite-valued variable $Z=f(\mathbf X)$ specifies the answer whose availability is being assessed. The public query fixes $f$ before the consumer reads memory. It supplies no additional history-dependent information. All entropies and logarithms in this section use base two.

\subsection{Permitted access and actual use}

Let $\mathcal D_{r_{\rm qry}}$ be a declared family of deterministic read procedures. Each procedure is a finite binary decision tree of depth at most $r_{\rm qry}$: an internal node reads a specified physical cell, and a leaf terminates. Addresses may depend on earlier answers. The family can impose additional program or interface restrictions, but contains a no-read procedure and is nested in the budget: $\mathcal D_{r_{\rm qry}}\subseteq\mathcal D_{r_{\rm qry}+1}$. Let $\mathsf{tr}_{\mathsf d}(\mathbf Q)$ be the complete transcript of procedure $\mathsf d$, including addresses, returned bits, and termination. The tree and its public input are fixed independently of the realized history.

\begin{definition}[Information available through permitted reads]
\label{d:available-definition}
The information about $Z$ available with $r_{\rm qry}$ probes is
\[
 \mathsf{Avail}_{r_{\rm qry}}(Z;\mathbf Q)
   =\max_{\mathsf d\in\mathcal D_{r_{\rm qry}}}
       I\bigl(Z;\mathsf{tr}_{\mathsf d}(\mathbf Q)\bigr).
\]
The maximum is over the distinct permitted tree behaviors. There are only finitely many such behaviors for fixed $s,r_{\rm qry}$, even if several programs implement the same tree.
\end{definition}

This is an availability measure for the specified access contract. A system's installed consumer is one particular procedure and one particular answer rule. Its accuracy must be checked for those actual operations. Even a transcript determining $Z$ completely can be followed by an incorrect answer rule. Conversely, a hypothetical full-state decoder does not enter $\mathcal D_{r_{\rm qry}}$ unless it obeys the declared access restrictions. Fixed program storage and workspace retain their separate charges in the resource ledger.

The connection with predictive information is exact when the permitted terminal computation can report any distribution on $Z$ at each leaf. For a fixed tree with transcript $W$, the expected logarithmic loss of a reported distribution $\widehat p_W$ is
\[
 \mathbb E[-\log_2\widehat p_W(Z)]
   = H(Z\mid W)+
     \mathbb E\,\operatorname{KL}_2
       \bigl(\mathbb P(Z\in\cdot\mid W)\,\|\,\widehat p_W\bigr).
\]
Minimizing the loss gives $H(Z\mid W)$; the best no-read prediction has loss $H(Z)$. Hence $\mathsf{Avail}_{r_{\rm qry}}$ is exactly the best reduction in logarithmic loss attainable through the permitted trees with unrestricted terminal predictions. This is a bounded-access instance of predictive $\mathcal V$-information (Xu et al., 2020). If program restrictions exclude those posterior predictions, the corresponding predictive improvement can be smaller than $\mathsf{Avail}_{r_{\rm qry}}$. The transcript definition separates the information exposed by the reads from the subsequent ability to compute or implement a prediction.

\begin{proposition}[Budget, information, and randomization bounds]
\label{d:available-bounds}
For every declared target and nested access family,
\[
 0=\mathsf{Avail}_0(Z;\mathbf Q)
 \le \mathsf{Avail}_{r_{\rm qry}}(Z;\mathbf Q)
 \le \min\{H(Z),\ I(Z;\mathbf Q),\ r_{\rm qry}\}.
\]
The score is nondecreasing in $r_{\rm qry}$ and under enlargement of the permitted family. If reading all $s$ cells is permitted, then
$\mathsf{Avail}_s(Z;\mathbf Q)=I(Z;\mathbf Q)$.
Independent randomization over permitted procedures does not increase the maximum.
\end{proposition}
\begin{proof}
The no-read transcript is constant. Nested families and nonnegativity of mutual information give the monotonicity statements. Each transcript is a deterministic function of $\mathbf Q$, so data processing gives the $I(Z;\mathbf Q)$ bound, and mutual information is at most $H(Z)$. A depth-$r_{\rm qry}$ binary tree has at most $2^{r_{\rm qry}}$ leaves. Its transcript determines a leaf and is determined by that leaf and the fixed tree, so its entropy is at most $r_{\rm qry}$. In particular, adaptive addresses and the stopping decision do not introduce an uncharged information channel. Reading every cell makes the transcript determine $\mathbf Q$, giving equality at full access.

For a randomized consumer, sample its independent coins before execution. At finite depth their only relevant effect is to select one of the finitely many deterministic tree behaviors; denote its index by $\mathbf R$. Then
\[
 I\bigl(Z;\mathbf R,\mathsf{tr}_{\mathsf d_{\mathbf R}}(\mathbf Q)\bigr)
 =\sum_{\iota}\mathbb P(\mathbf R=\iota)
   I\bigl(Z;\mathsf{tr}_{\mathsf d_{\iota}}(\mathbf Q)\bigr)
 \le\mathsf{Avail}_{r_{\rm qry}}(Z;\mathbf Q),
\]
because $\mathbf R$ is independent of $(Z,\mathbf Q)$. This includes fresh random choices made adaptively during a run: their presampled tape selects a total finite tree. A consumer with fewer observable coins has no more information.
\end{proof}

For independent fair target bits $\mathbf X=(X_1,\ldots,X_N)$, permit a separate public coordinate query for each $i$. Its allowed tree family may depend on $i$. Define
\begin{equation}
 \mathsf{A}_{r_{\rm qry}}(\mathbf X;\mathbf Q)
   =\sum_{i=1}^N\mathsf{Avail}_{r_{\rm qry}}(X_i;\mathbf Q),\qquad
 \mathsf{Gap}_{r_{\rm qry}}(\mathbf X;\mathbf Q)
   =I(\mathbf X;\mathbf Q)-\mathsf{A}_{r_{\rm qry}}(\mathbf X;\mathbf Q).
 \label{d:availability-gap}
\end{equation}
The sum evaluates the availability of each separately requested bit. It is not the information obtained by a single procedure with $r_{\rm qry}$ probes, nor is it an estimate of the total information in all possible transcripts. Independence of the targets gives
\[
 \mathsf A_{r_{\rm qry}}(\mathbf X;\mathbf Q)
 \le\sum_i I(X_i;\mathbf Q)
 \le I(\mathbf X;\mathbf Q)\le N.
\]
Indeed, $H(\mathbf X)=\sum_iH(X_i)$ and
$H(\mathbf X\mid\mathbf Q)\le\sum_iH(X_i\mid\mathbf Q)$.
Thus the gap is nonnegative and yields the exact decomposition
\begin{equation}
 N-\mathsf A_{r_{\rm qry}}
   =\bigl(N-I(\mathbf X;\mathbf Q)\bigr)+\mathsf{Gap}_{r_{\rm qry}}.
 \label{d:availability-decomposition}
\end{equation}
The first term measures target information absent from the ready state; the second measures stored information not available for the separately specified coordinate uses at this budget. The independence assumption is essential for this comparison: redundant target coordinates can make a sum of coordinate informations exceed their joint information. Even with independent targets, the gap can include information about relations that does not answer the chosen coordinate queries: storing only $X_1\oplus X_2$ gives one bit of joint information and zero information about either fair coordinate, at every read budget. When $I(\mathbf X;\mathbf Q)=N$, all coordinates are retained, and the gap instead vanishes once their permitted reads can recover them.

\subsection{Complete information with no bounded-probe access}

\begin{theorem}[A maximal access gap under an invertible recoding]
\label{d:availability-sharing}
For every $N\ge1$ and integer $r_{\rm qry}\ge1$, there is a complete ready state of $s=N(r_{\rm qry}+1)$ bits such that
\[
 I(\mathbf X;\mathbf Q)=N,\qquad
 \mathsf A_{r_{\rm qry}}(\mathbf X;\mathbf Q)=0,\qquad
 \mathsf A_{r_{\rm qry}+1}(\mathbf X;\mathbf Q)=N,
\]
where $\mathbf X$ consists of $N$ independent fair bits and all depth-bounded read trees are permitted. Every consumer using at most $r_{\rm qry}$ probes has error $1/2$ on each coordinate target. An invertible recoding of the same $s$ cells instead gives $\mathsf A_1=N$.
\end{theorem}
\begin{proof}
For each coordinate $i$, choose independent fair bits $R_{i1},\ldots,R_{i,r_{\rm qry}}$, jointly independent of $\mathbf X$, and store the block
\[
 Q_{i0}=X_i\oplus R_{i1}\oplus\cdots\oplus R_{i,r_{\rm qry}},\qquad
 Q_{ij}=R_{ij}\quad(1\le j\le r_{\rm qry}).
\]
All shares, including the masks, are counted in the ready state. The parity of block $i$ is $X_i$, so the complete state determines $\mathbf X$ and $I(\mathbf X;\mathbf Q)=N$.

Conditional on any fixed $\mathbf X=x$, a specified proper subset of the $r_{\rm qry}+1$ cells in a block is uniformly distributed. If the parity cell is included, at least one omitted mask remains fair; if it is excluded, the observed cells are independent masks. The blocks are independent conditional on $x$. Consequently any specified set of at most $r_{\rm qry}$ distinct physical cells has a uniform distribution independent of $x$.

Consider a deterministic adaptive tree of depth at most $r_{\rm qry}$. At a reachable leaf, the particular addresses and stopping choice are fixed by the preceding answer string. The event of reaching that leaf is exactly the conjunction of its observed cell values; repeated reads impose no new constraint. If $d_{\rm seen}$ distinct cells were read, this event has conditional probability $2^{-d_{\rm seen}}$ for every $x$. Hence the entire leaf transcript is independent of $\mathbf X$. This argument covers early termination as well as adaptivity. Independent query randomization preserves the conclusion by conditioning on its tree. Every coordinate remains fair given all the information used by such a consumer, so any answer rule has error $1/2$.

With $r_{\rm qry}+1$ reads the consumer reads one complete block and returns its parity, giving exact access to the requested coordinate. Finally, the map from $(\mathbf X,(R_{ij})_{i,j})$ to $\mathbf Q$ is a bijection. In the former encoding each $X_i$ occupies a directly readable cell, so one probe suffices with exactly the same total retained storage. Thus the gap depends on the physical access contract and the encoding, although the recoding preserves total information.
\end{proof}

The construction is an additive secret-sharing device; threshold sharing supplies the general precedent (Shamir, 1979). Its purpose here is to exhibit an exact separation between retention and permitted use. The independent blocks avoid inferring joint invisibility merely from invisibility of individual bits. With shared masks, accessible relations between targets can carry information even when every target separately remains fair. In general, the coordinate score in~\eqref{d:availability-gap} and joint transcript information must therefore be reported separately.

\subsection{Consequences for an installed consumer}

\begin{proposition}[From accessible information to native error]
\label{d:availability-fano}
Let $Z$ take values in an alphabet of size $n_{\rm ans}\ge2$. Suppose an installed consumer uses a permitted depth-$r_{\rm qry}$ tree, independent query randomness, and a terminal answer in the target alphabet. If its error probability is $\upsilon$, then
\begin{equation}
 H(Z)-\mathsf{Avail}_{r_{\rm qry}}(Z;\mathbf Q)
 \le h_2(\upsilon)+\upsilon\log_2(n_{\rm ans}-1),
 \label{d:availability-fano-bound}
\end{equation}
where $h_2$ is binary entropy. In particular, for independent fair coordinate targets and native errors $\upsilon_i$, the mean error satisfies
\begin{equation}
 \frac1N\sum_i \upsilon_i
 \ge h_2^{-1}\!\left(1-\frac{\mathsf A_{r_{\rm qry}}(\mathbf X;\mathbf Q)}N\right),
 \label{d:availability-binary-error}
\end{equation}
where $h_2^{-1}$ takes values in $[0,1/2]$.
\end{proposition}
\begin{proof}
Include the independent coins and the read transcript in $W$ so that the installed answer is a function of $W$. The preceding proposition gives $I(Z;W)\le\mathsf{Avail}_{r_{\rm qry}}(Z;\mathbf Q)$. To obtain the needed form of Fano's inequality, let $B$ indicate that the answer is incorrect. Given $W$ and $B=0$, the target is known; given $W$ and $B=1$, at most $n_{\rm ans}-1$ values remain. Hence
\[
 H(Z\mid W)\le H(B\mid W)+\upsilon\log_2(n_{\rm ans}-1)
 \le h_2(\upsilon)+\upsilon\log_2(n_{\rm ans}-1).
\]
Since $H(Z)-\mathsf{Avail}_{r_{\rm qry}}\le H(Z\mid W)$, this proves~\eqref{d:availability-fano-bound}. For fair binary targets it yields $1-\mathsf{Avail}_{r_{\rm qry}}(X_i;\mathbf Q)\le h_2(\upsilon_i)$. Average over $i$ and use concavity of $h_2$. If the mean error is at most $1/2$, invert its increasing branch; otherwise~\eqref{d:availability-binary-error} holds because its right side is at most $1/2$.
\end{proof}

When the complete state retains all $N$ target bits, the right side of~\eqref{d:availability-binary-error} is $h_2^{-1}(\mathsf{Gap}_{r_{\rm qry}}/N)$. The gap then imposes an error floor on every permitted installed consumer, regardless of its total stored information. At the opposite endpoint, exact native readers for all coordinates give $\mathsf A_{r_{\rm qry}}=N$ and zero gap. In particular, the designated one-probe carriers established by Corollary~\ref{d:nativebitcarrier} have zero gap at $r_{\rm qry}=1$. The converse concerns availability rather than installation: a wrong answer rule can retain full accessible information. Returning the complement of a directly read target bit is an elementary example. Error and accessible information accordingly answer different questions about a native operation.

Process crypticity provides a related precedent for distinguishing stored information from observation-accessible information (Crutchfield, Ellison, \& Mahoney, 2009). Forward crypticity measures predictive causal-state information left unresolved by the entire observable future, $H(S^+\mid\text{future})$ (Mahoney, Ellison, \& Crutchfield, 2009). Its obstruction concerns the observation channel even with unrestricted computation. The present model fixes a physically retained state and varies the charged reads allowed to a consumer. The XOR construction has complete target information in that state; its obstruction disappears when the relevant additional cell can be read.

\subsection{An exact geometric endpoint}
\label{d:availability-geometry}

A separate quantity measures the physical effect of changing an acquired distinction. For an injective deterministic encoding $\mathsf{enc}:\bits^N\to\bits^s$, write $\mathsf{enc}_j$ for its $j$th output bit and let $x^{\oplus i}$ be $x$ with coordinate $i$ flipped. Define the uniform mean edge stretch and Boolean influence by
\[
 \mathsf{Str}(\mathsf{enc})=\frac1N\sum_{i=1}^N
       \mathbb E_{\mathbf X}d_H(\mathsf{enc}(\mathbf X),\mathsf{enc}(\mathbf X^{\oplus i})),
 \qquad
 \operatorname{Inf}_i(\mathsf{enc}_j)=
       \Pr(\mathsf{enc}_j(\mathbf X)\ne \mathsf{enc}_j(\mathbf X^{\oplus i})).
\]
Then
\begin{equation}
 N\mathsf{Str}(\mathsf{enc})=\sum_{j=1}^s\sum_{i=1}^N\operatorname{Inf}_i(\mathsf{enc}_j).
 \label{d:availability-stretch}
\end{equation}
This identity follows by expanding Hamming distance into changed-bit indicators and interchanging finite sums. Injectivity gives $\mathsf{Str}(\mathsf{enc})\ge1$. Equality holds exactly when every logical edge changes one physical bit. The binary cube rigidity theorem then gives distinct designated carriers, with fixed identity-or-complement readings, and constant remaining cells. Consequently equality yields zero one-probe access gap for independent fair coordinates.

The identity measures encoding geometry; interpreting it as an acquisition cost requires actual updates to take $\mathsf{enc}(x)$ to $\mathsf{enc}(x^{\oplus i})$. The next result gives a quantitative relationship under tight storage and uniform semantic sampling. Without those premises, neither this identity nor the information gap identifies a unique storage architecture. Together the measures record two distinct costs under the declared distribution and interface: how widely an update changes memory, and how much target information a bounded consumer can expose.

\subsection{Robust coordinates under a tight storage budget}
\label{d:robust-locality}

At the exact endpoint, one changed cell per semantic edge forces coordinate
carriers.  A quantitative counterpart holds when the store has exactly the
$N$ bits required to distinguish $2^N$ contents.  A small average excess
of changed cells then forces approximate carriers for most coordinates
and bounds the one-probe access gap.  The averaging is over all semantic
cube edges, and the availability statement uses uniform independent target
bits.  These distributional premises are part of the result.

Write $\mathbf X$ for a uniform element of $\bits^N$ and
$\mathbf X^{\oplus i}$ for the vector with bit $i$ flipped.  For a
bijection $\mathsf{enc}:\bits^N\to\bits^N$, retain the definition
\[
 \mathsf{Str}(\mathsf{enc})=
 \frac1N\sum_{i=1}^N\mathbb E\,
 d_H\bigl(\mathsf{enc}(\mathbf X),
          \mathsf{enc}(\mathbf X^{\oplus i})\bigr).
\]
A \emph{signed coordinate permutation} is a map
$g_j(x)=x_{\pi(j)}\oplus \mathsf{sgn}_j$, where $\pi$ is a permutation of $[N]$
and $\mathsf{sgn}_j\in\bits$.  Approximation below means expected Hamming distance,
rather than agreement of the complete words with high probability.

\subsubsection{A quantitative Boolean stability bound}

For a function $f:\bits^N\to\{-1,1\}$, let
$\chi_S(x)=(-1)^{\sum_{i\in S}x_i}$ and
$\widehat f(S)=\mathbb E[f(\mathbf X)\chi_S(\mathbf X)]$.
Its total influence is
\[
 \mathsf I(f)=\sum_{i=1}^N
   \Pr\bigl(f(\mathbf X)\ne f(\mathbf X^{\oplus i})\bigr)
 =\sum_{S\subseteq[N]}|S|\widehat f(S)^2.
\]
The identity follows by expanding the squared discrete derivative in
the orthonormal character basis.

\begin{lemma}[Balanced stability with an explicit constant]
\label{d:balanced-stability}
If $\mathbb Ef=0$, there are $i\in[N]$ and
$\sigma\in\{-1,1\}$ such that
\begin{equation}
 \Pr\bigl(f(\mathbf X)\ne\sigma\chi_i(\mathbf X)\bigr)
 \le\frac32\sum_{|S|\ge2}\widehat f(S)^2
 \le\frac32\bigl(\mathsf I(f)-1\bigr).
 \label{d:balanced-stability-bound}
\end{equation}
\end{lemma}
\begin{proof}
Put $\omega=\sum_{|S|\ge2}\widehat f(S)^2$ and
$v=\sqrt{1-\omega}$.  Parseval's identity and balance give
$\sum_i\widehat f(\{i\})^2=v^2$ and
$\mathsf I(f)-1=\sum_{|S|\ge2}(|S|-1)\widehat f(S)^2\ge\omega$.
If $v=0$, a choice of sign gives disagreement at most $1/2$ and the
claim follows.  Otherwise set
$L=\sum_i\widehat f(\{i\})\chi_i$.
Since $\mathbb E[fL]=v^2$ and $|f|=1$, we have
$\mathbb E|L/v|\ge v$.

The inequality of K\"onig, Sch\"utt, and Tomczak-Jaegermann (1999) used in
the FKN stability theorem states that, for $\sum_i a_i^2=1$,
\[
 \mathbb E\left|\sum_i a_i\chi_i\right|
 \le c_0+(1-c_0)\max_i|a_i|,
 \qquad c_0=\sqrt{2/\pi}.
\]
We use its explicit formulation in Friedgut, Kalai, and Naor
(2021, Theorem~0.2), which also supplies the balanced proof of the
original FKN theorem (Friedgut, Kalai, \& Naor, 2002).
Applying it to $a_i=\widehat f(\{i\})/v$ gives
\[
 a_*:=\max_i|\widehat f(\{i\})|
 \ge\frac{v(v-c_0)}{1-c_0}.
\]
The best signed-coordinate disagreement is $(1-a_*)/2$.  Therefore
\[
 \frac{1-a_*}{2}
 \le\frac{\omega-c_0(1-\sqrt{1-\omega})}{2(1-c_0)}
 \le\frac{1-c_0/2}{2(1-c_0)}\,\omega
 <\frac32\omega,
\]
using $1-\sqrt{1-\omega}\ge\omega/2$ and $c_0<4/5$.
When $\omega=0$, the same computation gives zero disagreement;
the strict final inequality is then replaced by equality.
\end{proof}

\subsubsection{From update geometry to approximate carriers}

\begin{theorem}[Robust coordinate organization at tight storage]
\label{d:robust-tight}
Let $N\ge1$, let $\mathsf{Exc}_{\rm loc}\ge0$, and let
$\mathsf{enc}:\bits^N\to\bits^N$ be a bijection with
$\mathsf{Str}(\mathsf{enc})\le1+\mathsf{Exc}_{\rm loc}$.
There is a signed coordinate permutation $g$ such that
\begin{equation}
 \mathbb E\,d_H\bigl(\mathsf{enc}(\mathbf X),g(\mathbf X)\bigr)
 \le N\min\{1/2,\,3\mathsf{Exc}_{\rm loc}\}.
 \label{d:robust-tight-distance}
\end{equation}
Moreover, for every $0<\eta<1/4$, there is a set $G\subseteq[N]$
of semantic coordinates, with distinct physical cells $j(i)$ and bits
$\mathsf{sgn}_i$, such that
\begin{equation}
 |G|\ge N\left(1-\frac{3\mathsf{Exc}_{\rm loc}}{2\eta}\right),
 \qquad
 \Pr\bigl(X_i\ne\mathsf{enc}(\mathbf X)_{j(i)}\oplus \mathsf{sgn}_i\bigr)
 \le\eta\quad(i\in G).
 \label{d:robust-most-carriers}
\end{equation}
If the permitted one-probe family contains every fixed single-cell read,
then, for $\mathbf Q=\mathsf{enc}(\mathbf X)$,
\begin{align}
 \mathsf A_1(\mathbf X;\mathbf Q)
 &\ge N\bigl[1-h_2(\min\{1/2,3\mathsf{Exc}_{\rm loc}\})\bigr],
 \label{d:robust-availability}\\
 \mathsf{Gap}_1(\mathbf X;\mathbf Q)
 &\le N h_2(\min\{1/2,3\mathsf{Exc}_{\rm loc}\}).
 \label{d:robust-gap}
\end{align}
Each $i\in G$ separately satisfies
$\mathsf{Avail}_1(X_i;\mathbf Q)\ge1-h_2(\eta)$.
\end{theorem}
\begin{proof}
For each output coordinate put $f_j(x)=(-1)^{\mathsf{enc}(x)_j}$.
Bijectivity makes these functions balanced and their values jointly
independent under uniform $\mathbf X$.  Parseval's identity gives
$d_j:=\mathsf I(f_j)-1\ge0$.  The influence--stretch identity gives
\begin{equation}
 \sum_{j=1}^N d_j
 =N\bigl(\mathsf{Str}(\mathsf{enc})-1\bigr)
 \le N\mathsf{Exc}_{\rm loc}.
 \label{d:robust-deficit-sum}
\end{equation}
Choose a best signed-coordinate approximation to each $f_j$, and let
$\upsilon_j$ be its disagreement probability.  Lemma~\ref{d:balanced-stability}
gives $\upsilon_j\le(3/2)d_j$.

Call an output coordinate good when $\upsilon_j<1/4$.
Two good outputs cannot choose the same input coordinate.  Indeed,
after adjusting their signs, both would approximate the same character,
so their mutual disagreement would be at most
$\upsilon_j+\upsilon_k<1/2$.  Distinct output bits of a uniform bijection
are independent and fair, so that disagreement is exactly $1/2$,
including after either fixed sign change.  The chosen input coordinates
of the good outputs therefore form an injection.

Extend this injection to a permutation by assigning the remaining input
coordinates bijectively to the bad outputs.  For each new pair choose
the sign having disagreement at most $1/2$; the two sign choices have
errors summing to one.  If $B$ is the set of bad outputs, then
$|B|\le4\sum_{j\in B}\upsilon_j$.  Thus the completed signed
permutation has total expected coordinate error at most
\[
 \sum_{j\notin B}\upsilon_j+\frac{|B|}{2}
 \le\sum_{j\notin B}\upsilon_j+2\sum_{j\in B}\upsilon_j
 \le2\sum_j\upsilon_j
 \le3N\mathsf{Exc}_{\rm loc}.
\]
Every completed coordinate error is also at most $1/2$, proving
\eqref{d:robust-tight-distance}.  At most
$\sum_j\upsilon_j/\eta\le3N\mathsf{Exc}_{\rm loc}/(2\eta)$ outputs
have error greater than $\eta$.  The rest are good, have distinct input
coordinates, and supply~\eqref{d:robust-most-carriers}.

Use the permutation to assign one reader to each semantic coordinate,
and denote its error by $\upsilon_i^{\rm qry}\le1/2$.  The read cell and its fixed sign
predict the fair bit $X_i$, so the binary conditional-entropy bound yields
\[
 \mathsf{Avail}_1(X_i;\mathbf Q)
 \ge I(X_i;Q_{j(i)})\ge1-h_2(\upsilon_i^{\rm qry}).
\]
Summing, applying concavity of $h_2$, and using
$\frac1N\sum_i \upsilon_i^{\rm qry}\le\min\{1/2,3\mathsf{Exc}_{\rm loc}\}$ proves
\eqref{d:robust-availability}.  The complete bijective state has
$I(\mathbf X;\mathbf Q)=N$, which gives~\eqref{d:robust-gap}.
The same conditional-entropy argument gives the individual bound for
the readers in~\eqref{d:robust-most-carriers}.
\end{proof}

The bound becomes informative for non-coordinate encodings in a small
explicit example. Exchange the images of $0^N$ and $\mathbf b_1$ in the
identity encoding and leave every other image fixed. Its total edge
distortion is $2(N-1)$, so
\[
 \mathsf{Str}-1=\frac{4(N-1)}{N2^N}.
\]
For $N=5$ this excess is $1/10$, and the bound in
\eqref{d:robust-tight-distance} is $3N/10<N/2$. Thus the finite checks
include an explicit non-coordinate case in which the theorem improves on
the trivial signed-permutation bound, in addition to larger-dimensional
checks described in Appendix~\ref{focus:verification}.

The theorem supplies low-error permitted readers.  Attributing the same
accuracy to an installed consumer additionally requires that it use those
cells and the corresponding answer rule.  This preserves the distinction
between accessible information and actual native use.

The averaging cannot be strengthened to a guarantee for every coordinate
or to high-probability agreement of complete words.  For $N\ge2$, consider
\begin{equation}
 \mathsf{enc}(x)=(x_1\oplus x_2,x_2,x_3,\ldots,x_N).
 \label{d:triangular-xor}
\end{equation}
Only changes to the second input coordinate rewrite two cells; every
other direction rewrites one.  Hence
$\mathsf{Str}=1+1/N$.  Nevertheless every single output cell is
independent of $X_1$, so $\mathsf{Avail}_1(X_1;\mathbf Q)=0$.
The first output is also independent of every individual input bit.
For every signed coordinate permutation $g$ it therefore disagrees with
$g_1(\mathbf X)$ with probability $1/2$, and
$\Pr(\mathsf{enc}(\mathbf X)\ne g(\mathbf X))\ge1/2$.
Normalized average Hamming error is the conclusion supported here.

\subsubsection{Actual histories and a quantified canonical defect}

At tight storage, exact native coordinate answers already remove history
dependence at ready boundaries.  If all $2^N$ contents are attainable in
$N$ binary cells, the sets of ready states supporting different exact
answer profiles are disjoint and nonempty.  There are only $2^N$ physical
states, so each profile has exactly one ready state.  This counting
argument applies to supports of randomized acquisition as well.
Consequently an exact actual update to a new content must reach that
content's unique code.  The preceding theorem applies when the measured
mean update cost covers the full uniform family of semantic edges.

For noisy or history-dependent ensembles in the same $N$-bit store,
proximity to a common code must instead be established.  The following transfer statement specifies what such
an assessment must measure.  Select each undirected edge of $\bits^N$
with equal probability and orient it by any fixed rule, giving endpoints
$X^-,X^+$.  Let $R^-,R^+\in\bits^N$ be jointly distributed actual
ready states before and after the corresponding operation.  They may
depend on the histories used to realize that edge and on acquisition
randomness.  Fix a reference bijection $\mathsf{enc}$ and define the
\emph{endpoint canonical defect}, in changed-bit units, by
\[
 \mathsf{Def}_{\rm can}=\mathbb E\bigl[
 d_H(R^-,\mathsf{enc}(X^-))+
 d_H(R^+,\mathsf{enc}(X^+))\bigr].
\]

\begin{proposition}[Transfer through a measured canonical defect]
\label{d:robust-canonical-defect}
If $\mathbb E d_H(R^-,R^+)\le1+\mathsf{Exc}_{\rm run}$, with
$\mathsf{Exc}_{\rm run}\ge0$, then
\begin{equation}
 \mathsf{Str}(\mathsf{enc})
 \le1+\mathsf{Exc}_{\rm run}+\mathsf{Def}_{\rm can}.
 \label{d:robust-defect-transfer}
\end{equation}
For a possibly different evaluation ensemble with uniform target
$\mathbf X\in\bits^N$ and complete ready state $\mathbf Q\in\bits^N$,
suppose
$\mathbb E d_H(\mathbf Q,\mathsf{enc}(\mathbf X))\le N\upsilon_{\rm can}$.
If all fixed single-cell reads are permitted, put
$\upsilon_* =\min\{1/2,3(\mathsf{Exc}_{\rm run}+\mathsf{Def}_{\rm can})+\upsilon_{\rm can}\}$.
Then
\begin{equation}
 \mathsf A_1(\mathbf X;\mathbf Q)\ge N[1-h_2(\upsilon_*)].
 \label{d:robust-defect-availability}
\end{equation}
\end{proposition}
\begin{proof}
For each sampled edge the triangle inequality bounds the distance
between its reference codes by its actual endpoint distance plus the
two endpoint defects.  Averaging proves~\eqref{d:robust-defect-transfer},
because Hamming distance is symmetric and the edge is sampled uniformly.
Theorem~\ref{d:robust-tight} supplies a signed coordinate permutation
whose mean total error on the reference encoding is at most
$3N(\mathsf{Exc}_{\rm run}+\mathsf{Def}_{\rm can})$.
Apply its readers to $\mathbf Q$.  Since each physical cell is assigned
once, the sum of their error probabilities increases by at most
$\mathbb E d_H(\mathbf Q,\mathsf{enc}(\mathbf X))\le N\upsilon_{\rm can}$.
Complement any reader with error greater than $1/2$; this only reduces
its error.  The same conditional-entropy argument and concavity used
above now give~\eqref{d:robust-defect-availability}.
\end{proof}

The defect is an additional metric premise.  Counting histories per
content gives no bound on it.  If only the probabilities of endpoint
disagreement with the reference code are bounded by $\rho_-$ and $\rho_+$,
the general estimate is
$\mathsf{Def}_{\rm can}\le N(\rho_-+\rho_+)$, since a disagreeing pair can differ
in all $N$ cells.  An immediate discrepancy between two update orders
can also be amplified by subsequent updates unless their propagation
is controlled.  Thus local commutation tests need either a bound on
the resulting endpoint defect or a stability assumption on subsequent
transitions before they support this transfer.

A small exact example separates the selected-history statistic $J_{\rm end}$
from global coordinate organization.  Start three-fact acquisition at
$0000$, normally storing content $x$ as $(x,0)$ and inserting a fresh fact
by changing its raw bit.  Make only these two exceptions:
\[
 0100\xrightarrow{\,1\,}0101
       \xrightarrow{\,3\,}0111.
\]
The exceptional states represent $110$ and $111$, respectively;
repeated insertions do nothing.  Every fresh update changes one bit,
and the finite native decoder answers all conjunctions exactly.
The selected orders $231$, $132$, and $123$ all finish at $1110$, so
$J_{\rm end}=1$.  Yet no single cell reads the first logical bit exactly:
$0100$ and $0101$ agree in their first cell and have different first
logical bits, while $0000$ and $1000$ witness the same failure for each
other cell.  Thus coalescence of the selected final histories does not
supply the global metric premise used above.

For an injection into $s>N$ cells, the outputs need not be balanced or
independent.  The nonnegative per-output deficits and the distinctness
argument in Theorem~\ref{d:robust-tight} then no longer follow.
The tight-storage theorem makes no unrestricted surplus-storage claim.
Nor does a bound under one selected distribution of experienced edges
by itself imply its premise for uniform cube edges.

\section{The resource cost of conditional rules and input provenance}
\label{d:horn-resources}

The distinctions established by rule evidence depend on what the acquired resource must subsequently support. Present atomic consequences, consequences under additional facts, and revision of named inputs are different obligations. The semantic hierarchy in Wu (2026c, Theorem~5.10) identifies their sufficient contents. The results below give information bounds and actual read/write implementations for those contents. A finer semantic resource need not have a larger update cost: storing explicit inputs can make revision cheaper than maintaining their consequences for one-probe use.

\subsection{Evidence interfaces and three information requirements}

Fix $n_{\rm at}\geq3$ atoms $\mathsf{At}$ and the public canonical vocabulary
\[
 \mathcal R_{\rm H}=\{(B\to \zeta_0):B\subseteq\mathsf{At},\ \zeta_0\in\mathsf{At}\setminus B\},
 \qquad R_{\rm H}=|\mathcal R_{\rm H}|=n_{\rm at}2^{n_{\rm at}-1}.
\]
The empty antecedent represents a fact. Evidence consists of actual commands $\operatorname{Add}(\mathsf{rule})$, initially applied to the empty base; where stated, $\operatorname{Del}(\mathsf{rule})$ is also permitted. Bases are sets: repetitions and deletion of an absent rule have no additional semantic effect. There is no undeletable background theory and no uncharged copy of the evidence. Let $\cl_K(F)$ be the least set containing $F$ and closed under the acquired rules $K$.

The conditional native portfolio is
\[
 T_{\rm Horn}=\{t=(F,\zeta):F\subseteq\mathsf{At},\ \zeta\in\mathsf{At}\},
 \qquad b(\cl_K,t)=\one[\zeta\in \cl_K(F)].
\]
Here $F$ is part of a read-only hypothetical query. Its processing and the final atomic answer share the declared read budget $r_{\rm qry}$; an implementation cannot preprocess the retained rule information for free before answering. These tests have the same answers as adding the facts in $F$ and then querying $\zeta$, although the hypothetical query preserves the acquired base. Actual fact additions remain available as evidence. Queries with $\zeta\in F$ have the constant answer one.

Let $c_{n_{\rm at}}$ be the number of closure operators on the labeled set $\mathsf{At}$. Ready storage below counts all evidence-dependent bits available to delayed native uses. Unlimited computation does not mean unlimited uncharged storage: additional working memory belongs to peak storage, and a precomputed decoding table belongs to the fixed-program ledger.

\begin{theorem}[Information required by the Horn interfaces]
\label{d:horn-information}
The following bounds concern systems that acquire their states from the declared rule-evidence histories.
\begin{enumerate}
\item If an announced terminal boundary permits only present atomic queries and no further evidence, there are exactly $2^{n_{\rm at}}$ answer profiles. The minimum ready information capacity is $n_{\rm at}$ bits.
\item If further evidence permits arbitrary rule additions but no deletions, or if no further evidence is permitted and every conditional query in $T_{\rm Horn}$ is required, there are exactly $c_{n_{\rm at}}$ distinguishable conditional resources. Every exact implementation requires at least $\lceil\log_2c_{n_{\rm at}}\rceil$ ready bits. Moreover,
\begin{equation}
 \binom{n_{\rm at}}{\lfloor n_{\rm at}/2\rfloor}
 \leq\log_2c_{n_{\rm at}}\leq 2^{n_{\rm at}}-1.
 \label{d:horn-closure-count}
\end{equation}
The information lower bound is attainable when computation, peak storage, and the cost of a fixed decoder are unrestricted and accounted for separately.
\item If every named rule may subsequently be deleted, there are exactly $2^{R_{\rm H}}$ distinguishable explicit bases. Every exact implementation requires at least $R_{\rm H}$ ready bits, even if its only final query is a present atomic query. An explicit rule-presence vector attains this capacity.
\end{enumerate}
\end{theorem}
\begin{proof}
Every subset $A\subseteq\mathsf{At}$ is the present consequence set of the facts in $A$. Different subsets require different atomic answers, and retaining that set in $n_{\rm at}$ bits suffices once further evidence is prohibited. It can be obtained from the actual rule stream by maintaining a sufficient rule resource before the terminal boundary, computing its consequences there, and discarding the additional information. This terminal capacity assertion places no $n_{\rm at}$-bit bound on the preceding acquisition process.

For the second claim, unequal closures differ at some $(F,\zeta)$ and hence are separated by that query, or by the continuation adding $F$ and querying $\zeta$. Equal closures remain equivalent after a common rule addition: for $\mathsf{rule}=(B\to \zeta_0)$,
\begin{equation}
 \cl_{K\cup\{\mathsf{rule}\}}(F)=
 \begin{cases}
 \cl_K(F),&B\not\subseteq \cl_K(F),\\
 \cl_K(F\cup\{\zeta_0\}),&B\subseteq \cl_K(F).
 \end{cases}
 \label{d:horn-single-update}
\end{equation}
If the antecedent is absent from $\cl_K(F)$, that old closed set already satisfies the new rule. Otherwise the new rule derives $\zeta_0$, after which the old closure of $F\cup\{\zeta_0\}$ satisfies both the old theory and the new rule. These observations prove the formula and, by induction, sufficiency through every sequence of additions.

Every closure operator $C$ on a finite set is realized by a rule base: include $F\to \zeta$ for all $\zeta\in C(F)\setminus F$. Its least closure is $C$, since the included rules derive $C(F)$ from $F$, while monotonicity and idempotence make $C(F)$ a model of all those rules. Each base is reachable by inserting its rules. Thus all $c_{n_{\rm at}}$ resources are reached through legal evidence, and different resources require different complete ready states. Enumerating these finitely many closures, using their enumeration indices as ready codes, and decoding, updating, and re-encoding at each operation attains the information capacity. Such enumeration is a computable construction with potentially enormous time, workspace, or fixed-table cost; it supplies no efficiency guarantee.

A closure is determined by its family of closed sets, which contains $\mathsf{At}$ and is closed under intersection. Its membership vector on the other $2^{n_{\rm at}}-1$ subsets gives the upper bound in~\eqref{d:horn-closure-count}. For the lower bound put $k=\lfloor n_{\rm at}/2\rfloor$. For any collection $\mathcal A$ of $k$-element sets, the family consisting of $\mathsf{At}$, all sets of size below $k$, and the members of $\mathcal A$ is intersection-closed. Distinct choices give distinct closures. There are $2^{\binom{n_{\rm at}}k}$ such choices.

For the third claim, choose $\mathsf{rule}=(B\to \zeta_0)$ on which two bases differ. Delete every rule other than $\mathsf{rule}$, and then insert the facts in $B$. The terminal answer to $\zeta_0$ is precisely whether $\mathsf{rule}$ was present. If $B=\varnothing$, no facts need be inserted. Since $\zeta_0\notin B$, the facts alone cannot make the answer true. This separates all $2^{R_{\rm H}}$ bases reached from the empty base. One presence bit per canonical rule, set or cleared by its named command, attains the capacity and preserves the exact explicit base.
\end{proof}

The first item cannot be used as a space bound for a learner that must keep accepting arbitrary rules. A dormant rule may affect none of its present answers yet change an answer after later facts. Conversely, the last item does not require storing every derivation or every occurrence of an input. It requires the canonical set of independently supplied rules because that is exactly what the deletion interface can distinguish. Restricted deletions, occurrence-sensitive identities, or an undeletable background define different tasks.

\subsection{An exact one-read endpoint for conditional use}

The next result turns the semantic resource into a physical maintenance requirement. It applies to arbitrary history-dependent encodings, not just canonical endpoints.

\begin{theorem}[One-probe conditional answers and exact maintenance]
\label{d:horn-one-read}
Suppose an exact binary implementation accepts arbitrary Horn-rule additions and answers every $t\in T_{\rm Horn}$ with at most one retained-cell read. Then it requires at least $R_{\rm H}=n_{\rm at}2^{n_{\rm at}-1}$ ready cells. Some single rule insertion changes at least
\begin{equation}
 U_{\rm H}=(n_{\rm at}+1)2^{n_{\rm at}-2}
 \label{d:horn-maintenance-cost}
\end{equation}
retained cells. Both bounds are simultaneously attained by storing the $R_{\rm H}$ nonconstant conditional answers. In this implementation each insertion uses at most $R_{\rm H}+U_{\rm H}$ total retained-cell accesses and peak storage at most $2R_{\rm H}+O(n_{\rm at})$ bits, in addition to fixed code and the public input.
\end{theorem}
\begin{proof}
For each $F\subseteq\mathsf{At}$ and $\zeta\notin F$, define the Boolean function $f_{F,\zeta}(K)=\one[\zeta\in \cl_K(F)]$. All these functions are zero at the empty base and nonconstant. They are pairwise distinct. If the heads differ, the base containing only $F\to \zeta$ separates $f_{F,\zeta}$ from any $f_{G,\zeta^{\prime}}$ with $\zeta^{\prime}\ne \zeta$. For equal heads and different antecedents, use the single rule whose antecedent is not contained in the other; such a choice exists because the antecedents differ. The common-carrier criterion of Lemma~\ref{d:oneprobe-functions} therefore assigns different physical cells to all $R_{\rm H}$ functions. On every ready state each designated bit equals its answer or its fixed complement. This conclusion allows multiple physical representatives of the same base.

Fix an atom $\zeta_0$ and first acquire
\[
 K_{\zeta_0}=\{\{\zeta_0\}\to \zeta:\zeta\in\mathsf{At}\setminus\{\zeta_0\}\}.
\]
For $\zeta_0\notin F$, its old closure is $F$; after inserting the fact $\zeta_0$, its new closure is $\mathsf{At}$. Exactly $n_{\rm at}-|F|$ nonconstant answers change at that $F$. Summing over the $2^{n_{\rm at}-1}$ subsets excluding $\zeta_0$ gives
\[
 \sum_{F\subseteq\mathsf{At}\setminus\{\zeta_0\}}(n_{\rm at}-|F|)
 =n_{\rm at}2^{n_{\rm at}-1}-(n_{\rm at}-1)2^{n_{\rm at}-2}=U_{\rm H}.
\]
Their distinct designated cells must all change during this actual insertion.

To attain the bounds, keep the table $\mathsf{Ans}_K(F,\zeta)=f_{F,\zeta}(K)$ for $\zeta\notin F$; answers with $\zeta\in F$ are constants. Initialize the table to zero, and update it by~\eqref{d:horn-single-update}. That formula proves the table invariant by induction over the evidence stream. A nontrivial query reads its indexed bit. For any inserted $B\to \zeta_0$, a row can change only if $\zeta_0\notin \cl_K(F)$, since otherwise the old row already satisfies the rule. In particular $\zeta_0\notin F$. At most $n_{\rm at}-|F|$ entries can change in that row, so the preceding sum bounds all changed entries by $U_{\rm H}$. The displayed history attains the bound.

An actual updater reads the old table once into charged workspace, computes the new entries from the old snapshot, and writes just those entries whose values change. The snapshot also supplies the old value for each comparison. This takes $R_{\rm H}$ reads and at most $U_{\rm H}$ writes; the ready table and its temporary snapshot occupy $2R_{\rm H}+O(n_{\rm at})$ bits. Computation is unrestricted but uses no past evidence outside the declared memory. Erasing the snapshot restores the ready boundary. Queries are read-only.
\end{proof}

Here one-read use forces physical carriers for \emph{conditional answers}. It does not force a separate cell for each explicitly supplied rule. The two families happen to have the same cardinality $R_{\rm H}$, but their contents and update costs differ: inserting one rule can change many consequences while changing only one provenance bit.

\subsection{A sharp one-change endpoint for named revision}

The opposite access choice stores the rules actually supplied and computes their conditional consequences when asked. Even here, reading the entire rule vector is unnecessary. Before the target $\zeta$ is first derived, a rule whose antecedent contains $\zeta$ cannot contribute to its derivation. There are only $(n_{\rm at}+1)2^{n_{\rm at}-2}=U_{\rm H}$ rules with antecedents omitting $\zeta$. An adaptive reader can improve this bound further by ceasing to examine a head as soon as that atom is derived. The resulting bound is exact.

\begin{theorem}[Exact conditional-query complexity of the explicit base]
\label{d:horn-query-exact}
Represent an acquired base by its $R_{\rm H}$ rule-presence bits. For a fixed query $(F,\zeta)$, the minimum deterministic worst-case number of retained-bit reads is
\begin{equation}
 \mathsf{DT}(F,\zeta)=
 \begin{cases}
 0,&\zeta\in F,\\
 2^{n_{\rm at}}-2^{|F|},&\zeta\notin F.
 \end{cases}
 \label{d:horn-exact-query}
\end{equation}
A uniform reader attains this bound, leaves the retained base unchanged, and uses at most $R_{\rm H}+O(n_{\rm at})$ working bits. Its maximum over conditional queries is $2^{n_{\rm at}}-1$.
\end{theorem}
\begin{proof}
If $\zeta\in F$, the answer is the constant one. Otherwise put $f=|F|$ and $\ell=n_{\rm at}-f\ge1$.

For the upper bound, initialize a reached set $S=F$ and a working marker for every rule, initially unread. Repeatedly select an unread rule $B\to\zeta_0$ with $B\subseteq S$ and $\zeta_0\notin S$, read its presence bit, and mark it read. If the bit is one, add $\zeta_0$ to $S$ immediately. Return one as soon as $\zeta$ is added. If no eligible unread rule remains, return zero. A fixed public order resolves the selection of eligible rules. No rule bit is read twice. Each addition is justified by an actually present rule. At a zero termination, every present rule with antecedent in $S$ has its head in $S$: an unread exception would still be eligible, and a previously read present exception would already have added its head. Thus $S=\cl_K(F)$ at that termination. This proves correctness without a copy of the evidence history.

Suppose $k$ nontarget atoms have been derived when the reader terminates, so $0\le k\le\ell-1$. Order them by derivation. Before the $i$th such head was added, every rule read with that head had its antecedent among the $f+i-1$ atoms already reached, giving at most $2^{f+i-1}$ probes into that head. Each of the other $\ell-k$ heads, including the target, receives at most $2^{f+k}$ probes: the reader stops immediately if the target is derived. This count includes probes to heads that never become derivable. Hence the total is at most
\[
 \sum_{i=1}^k2^{f+i-1}+(\ell-k)2^{f+k}
 =2^f\bigl((\ell-k+1)2^k-1\bigr)
 \le2^f(2^\ell-1),
\]
where the last inequality uses $h+1\le2^h$ for $h=\ell-k\ge1$. The markers, reached set, and public-order counters use $R_{\rm H}+O(n_{\rm at})$ working bits. The retained rule vector is never modified.

For the lower bound, order the atoms outside $F$ as $v_1,\ldots,v_\ell$, with $v_\ell=\zeta$. Retain as free rule bits only the disjoint groups
\[
 \mathcal R_i=\{B\to v_i:B\subseteq F\cup\{v_1,\ldots,v_{i-1}\}\},
 \qquad |\mathcal R_i|=2^{f+i-1},
\]
and fix every other rule bit to zero. Every assignment to these free bits is a legal acquired base. Let $g_i$ be the Boolean function indicating that $v_i$ is derived in this acyclic restriction. There are
\[
 n_{\rm free}=\sum_{i=1}^{\ell}|\mathcal R_i|=2^{n_{\rm at}}-2^f
\]
free bits. We show that the unique multilinear polynomial for $g_\ell$ has a nonzero coefficient on the monomial containing all $n_{\rm free}$ of them.

Fix the bits in $\mathcal R_1,\ldots,\mathcal R_{\ell-1}$. The target function is the disjunction of those $\mathcal R_\ell$ bits whose rule antecedents are satisfied. The coefficient of the product of \emph{all} $\mathcal R_\ell$ variables is therefore zero unless all those antecedents are satisfied, and otherwise is $(-1)^{|\mathcal R_\ell|+1}$. All those antecedents are satisfied exactly when all preceding atoms $v_1,\ldots,v_{\ell-1}$ have been derived. As a Boolean function of the earlier groups, this event is
\[
 \prod_{i<\ell}g_i
 =\prod_{i<\ell}\operatorname{OR}\{x_{\mathsf{rule}}:\mathsf{rule}\in \mathcal R_i\}.
\]
Indeed, deriving every preceding head requires at least one present rule in each group; conversely, one present rule per group derives every head in order. The groups are disjoint. Each disjunction has full-monomial coefficient $(-1)^{|\mathcal R_i|+1}$, so the coefficient on all $n_{\rm free}$ variables in $g_\ell$ is $\prod_{i=1}^{\ell}(-1)^{|\mathcal R_i|+1}$, which is nonzero. Empty products cover $\ell=1$.

A depth-$d$ deterministic bit-query tree represents a multilinear polynomial of degree at most $d$: at a query node the polynomial is $(1-x)P_0+xP_1$, and repeated queries may be removed. This standard decision-tree bound (Buhrman \& de Wolf, 2002), applied to the restriction above, gives $d\ge n_{\rm free}$. Together with the actual reader, it proves~\eqref{d:horn-exact-query}.
\end{proof}

\begin{corollary}[The tight-storage, one-change Horn endpoint]
\label{d:horn-one-change}
Suppose an exact binary learner accepts every named insertion and deletion from the empty base, answers every conditional query, and has exactly $s=R_{\rm H}$ complete ready bits. If each evidence command changes at most one ready bit, then its ready encoding is the explicit rule-presence vector up to a fixed permutation and fixed bit complements. Its minimum worst-case query cost is consequently $2^{n_{\rm at}}-1$, with the per-query costs in~\eqref{d:horn-exact-query}. These costs are attained with one overwrite per named evidence command and peak storage $2R_{\rm H}+O(n_{\rm at})$. No separate assumption of order-independent endpoints is required here.
\end{corollary}
\begin{proof}
The named-deletion argument in Theorem~\ref{d:horn-information} separates all $2^{R_{\rm H}}$ bases. Each is reached by actual insertions, and no complete ready word can belong to two different bases. There are exactly $2^{R_{\rm H}}$ physical words. Thus every base has exactly one ready word and every word is used: tight capacity itself makes the endpoint independent of its evidence history. Neighboring bases differ by one named rule. Their endpoints are distinct and at Hamming distance at most one, hence exactly one. Binary coordinate rigidity (Theorem~\ref{d:rigidity}) makes this bijection a fixed coordinate permutation with fixed complements. Each physical probe therefore reveals exactly one rule-presence bit, and Theorem~\ref{d:horn-query-exact} gives the lower bound. The direct presence vector, its indexed overwrite commands, and the reader just constructed attain it.
\end{proof}

The two endpoints impose different physical organizations. A one-read consumer requires carriers for conditional answers and makes some rule acquisitions change $U_{\rm H}$ cells. Tight storage with one-change named revision instead requires carriers for the supplied rules and makes some conditional uses read $2^{n_{\rm at}}-1$ cells. The latter conclusion uses the deletion interface and tight complete-state capacity; it does not impose a provenance-vector organization on every additions-only learner or on implementations with spare ready cells.

\subsection{Two alternatives to maintaining every conditional answer}

The one-probe lower bound depends on the native access budget. The same acquired closure admits a smaller model encoding with more expensive reads. Allowing named revision suggests retaining the explicit base instead. Table~\ref{tab:horn-implementations} states concrete attainable budgets, rather than a universal ordering of all implementations. The symbol $r_{\rm upd}$ counts retained-cell reads during an update, $c_{\rm upd}$ counts reads and writes together, and $u$ bounds changed bits.

\begin{proposition}[Operational alternatives for the acquired rule resource]
\label{d:horn-implementations}
There are uniform, exact native implementations with the following costs.
\begin{enumerate}
\item An additions-only model-membership vector uses $s=2^{n_{\rm at}}-1$ ready bits and at most $r_{\rm qry}=2^{n_{\rm at}-1}$ reads per conditional query. Inserting $B\to \zeta_0$ uses at most $2^{n_{\rm at}-|B|-1}$ retained-cell writes and no retained-cell reads; that many bits can actually change. Its worst-case $u=c_{\rm upd}=2^{n_{\rm at}-1}$ is attained. Query and update workspace is $O(n_{\rm at})$ bits.
\item The explicit rule-presence vector uses $s=R_{\rm H}$ ready bits and supports each named insertion or deletion with $u\leq1$ and $c_{\rm upd}=1$. Its conditional reader has the exact worst-case cost $r_{\rm qry}=2^{n_{\rm at}}-1$ and peak storage at most $2R_{\rm H}+O(n_{\rm at})$. Alternatively, repeated full scans give $r_{\rm qry}\leq(n_{\rm at}+1)R_{\rm H}$ with only $O(n_{\rm at})$ workspace. No persistent answer cache is used.
\end{enumerate}
\end{proposition}
\begin{proof}
For the first construction, index one retained bit $v(W)$ by each proper subset $W\subsetneq\mathsf{At}$. Its value is one exactly when $W$ satisfies every acquired rule. The omitted set $\mathsf{At}$ is always a model. Initialize all bits to one. On receiving $B\to \zeta_0$, overwrite $v(W)$ by zero at every index with $B\subseteq W$ and $\zeta_0\notin W$. There are $2^{n_{\rm at}-|B|-1}$ such indices, identified from the public rule alone, so the updater need not read their previous contents. From the empty base they all change. Intersection of the model families proves the invariant after every insertion, including repeated insertions.

For a nontrivial query $(F,\zeta)$, the answer is zero precisely when some surviving model contains $F$ but excludes $\zeta$. The installed reader enumerates those $2^{n_{\rm at}-|F|-1}$ indices and reads their bits, stopping if it finds one. If it finds none, it returns one. This follows from the model-intersection characterization of Horn closure, and the omitted full set cannot be a countermodel. The enumeration needs only $O(n_{\rm at})$ counter and control bits. If $\zeta\in F$, return one without reading. The worst-case query bound is $2^{n_{\rm at}-1}$, and a fact insertion attains the corresponding update bound.

For the second construction, use one presence bit per canonical rule; a named insertion or deletion overwrites its indexed bit. The reader and its exact probe cost are given by Theorem~\ref{d:horn-query-exact}. For the smaller-workspace alternative, store only a working atom set and scan counters, reading all $R_{\rm H}$ retained bits on each forward-chaining pass. There are at most $n_{\rm at}$ successful passes and one final pass. Both readers compute from the rules actually acquired and leave the retained vector unchanged.
\end{proof}

\begin{table}[t]
\centering\small
\begin{tabular}{@{}p{.26\linewidth}rrrr@{}}
\toprule
Installed resource & Ready bits $s$ & Query reads $r_{\rm qry}$ & Changed bits $u$ & Update accesses $c_{\rm upd}$\\
\midrule
All conditional answers & $R_{\rm H}$ & $1$ & $U_{\rm H}$ & $\leq R_{\rm H}+U_{\rm H}$\\
Model membership & $2^{n_{\rm at}}-1$ & $\leq2^{n_{\rm at}-1}$ & $\leq2^{n_{\rm at}-1}$ & $\leq2^{n_{\rm at}-1}$\\
Explicit rule presence & $R_{\rm H}$ & $2^{n_{\rm at}}-1$ & $\leq1$ & $1$\\
\bottomrule
\end{tabular}
\caption{Actual implementations after arbitrary rule evidence. The first two rows support additions; the third also supports named deletions. All rows answer the same conditional queries. Query bounds refer to one request, and the workspace of each implementation has the separately stated peak-storage cost. For the first row, storage and worst-case changed bits are jointly optimal under one-read use. For the last row, the query cost is optimal at tight storage and one-change named revision (Corollary~\ref{d:horn-one-change}); its costs at individual queries are given by~\eqref{d:horn-exact-query}. The table makes no claim to a complete intermediate tradeoff frontier.}
\label{tab:horn-implementations}
\end{table}

The membership vector deliberately forgets which redundant rules were independently supplied. It therefore cannot in general process named deletion, even with unlimited computation: two bases with the same models can require different models after the same deletion. The explicit-base lower bound in Theorem~\ref{d:horn-information} makes the missing distinction precise. Repairing that interface requires additional retained provenance, not a faster decoder of the same closure.

Materialized-view maintenance already studies the propagation of base changes to derived information. Gupta, Mumick, and Subrahmanian (1993) develop derivation counting for nonrecursive views and deletion-and-rederivation for recursive views, including changes to rule definitions. The present finite calculation isolates a different resource statement: across all exact binary implementations of the declared one-read conditional interface, the changed-answer count forces a matching number of changed physical cells. Under less restrictive access, explicit inputs or a model vector can support the same consequences with different costs. These comparisons connect the semantic hierarchy to physical acquisition without identifying semantic refinement with a universal cost hierarchy.

\section{Fault tolerance raises the local read and write costs}
\label{focus:fault}

Robustness is a further resource requirement on the same installation and
use operations. Its tension with inexpensive updating is established in
coding theory. Mazumdar, Chandar, and Wornell (2014) study update efficiency
under probabilistic channel error and rate constraints. Chandran,
Kanukurthi, and Ostrovsky (2014) obtain local updating and decoding for
the \emph{prefix Hamming} metric, which constrains corruption according to
recency of writes. The following simpler result instead requires exact
answers after any $f$ adversarial bit flips. These distinct fault contracts
must not be conflated.

Let each content $m$ have a nonempty fiber $H_m\subseteq\bits^s$ of clean
ready states; native queries separate distinct contents. Corruption occurs
between completed operations and flips at most $f$ bits, where $f$ is a
nonnegative integer.
Programs and temporary computation are reliable during an operation.

\begin{proposition}[Exact robustness imposes two local lower bounds]
\label{focus:fault-lower}
If every native query is correct on every permitted corrupted state,
distinct content fibers have Hamming distance at least $2f+1$.
Consequently every clean-endpoint update changing content requires at
least $2f+1$ bit writes. A deterministic read-only implementation of any
nonconstant native query requires at least $2f+1$ reads in the worst case,
even with adaptive addresses.
\end{proposition}
\begin{proof}
If clean words of different contents were at distance at most $2f$, split
their differing positions into two sets of size at most $f$. A word
agreeing with the first on one part and the second on the other would
be an allowed corruption of both. A separating query could not be correct
on that word. The distance bound follows; each differing endpoint bit
requires a write.

For the read bound choose clean words $q,q'$ requiring different answers.
Suppose a program always stops within $2f$ reads. On a newly probed
position where the words agree, return their common bit. On the first
$f$ newly probed positions where they differ, return the $q$-bit;
on subsequent such positions return the $q'$-bit. Repeat earlier answers
on repeat reads. The resulting transcript disagrees with each clean
word in at most $f$ positions. Completing unread cells once as in $q$
and once as in $q'$ produces two legal corrupted inputs with the same
adaptive transcript and stopping decision but different required answers,
a contradiction.
\end{proof}

The write bound compares clean completed endpoints; it is not a bound on
every transition starting from an already damaged state. Similarly,
trusted preprocessing can restore a store before cheap reads, but its
reads and writes are part of the charged maintenance procedure.

\begin{corollary}[A partial-label learner attaining the local bounds]
\label{focus:fault-repetition}
For $n_{\rm inst}\ge1$ instances with acquired labels in $\{\bot,0,1\}$, put $r_{\rm rep}=2f+1$.
There is a learner using $2n_{\rm inst} r_{\rm rep}$ retained bits that, after any $f$ bit flips,
reads an indicator $[v_i=y]$, for $y\in\{0,1\}$, in $r_{\rm rep}$ reads. From a clean state it uses
two reads and $r_{\rm rep}$ writes for a first label acquisition, $2r_{\rm rep}$ writes
for replacement by the opposite label, and no writes for a repeated label.
Thus first acquisition and single-indicator retrieval attain the preceding
lower bounds. Full restoration uses $2n_{\rm inst} r_{\rm rep}$ reads and at most $f$ writes.
\end{corollary}
\begin{proof}
Encode $\bot,0,1$ as $00,10,01$ and repeat each indicator bit $r_{\rm rep}$ times.
Initialize all blocks to zero. Every block contains at most $f$ errors,
so its majority is its correct indicator. Reading a complete block gives
the stated query cost. At a clean boundary one read per uniform block
identifies the old label. First acquisition changes one entire block;
replacement changes two, and repeated evidence changes none. These actual
evidence updates therefore install the declared partial labels. For
restoration, scan each block, compute its majority, and rewrite only its
minority positions. At most $f$ positions change in total. A one-block
buffer and counters require $O(r_{\rm rep}+\log(n_{\rm inst}r_{\rm rep}))$ temporary bits and avoid a
second scan to locate minority positions.
\end{proof}

The construction is not storage-optimal. It isolates a sharp local cost
of exact fault tolerance while keeping the learner, consumer, and writes
explicit. Faults during execution and randomized decoders with a nonzero
error budget require different specifications.

\section{What resource constraints add to an acquisition theory}
\label{d:positioning}

A content specification identifies distinctions that a system must preserve
through its native operations. The acquisition account of Wu (2026a) treats
entry into, continued use of, and return to the appropriate content states;
the task analysis of Wu (2026c) derives the distinctions required by
particular evidence and query interfaces. Learning guarantees become
guarantees for such operations when the learned result is installed in the
system's consumer (Wu, 2026b). The present analysis supplies the resource
conditions for that installation and use. It starts from actual evidence
histories, counts the states they must distinguish, and then asks which
physical organizations permit the required reads and updates.

This connection matters during learning as well as after identification.
A short description of the eventual target can coexist with many distinct
intermediate resources. Explicit version-space boundaries can grow
exponentially (Haussler, 1988), while consistency-based implementations
support version-space operations without storing those boundaries
(Hirsh, Mishra, \& Pitt, 1997). The lower bounds quantify over all encodings satisfying each theorem's stated premises. Some require canonical endpoints or tight storage; these restrictions remain part of the claim when the result is applied to an implementation. The Horn comparison makes
the role of that interface concrete. Conditional answers, model membership,
and explicit input provenance support different combinations of access and
revision. Maintaining every conditional answer for one-probe use forces
many physical changes after a single acquired rule; retaining that rule's
presence makes its own insertion cheap while shifting work to the reader.
Semantic sufficiency therefore leaves a substantive implementation problem.

Bounded-memory learning supplies a complementary comparison. Raz (2016)
proves a memory--sample tradeoff for streaming parity learning. Gonen,
Lovett, and Moshkovitz (2020) study realizable strong learning under a known
distribution and characterize a regime through statistical-query dimension.
Their learners already have explicit state transitions; the charged resources
are retained memory and samples, with a hypothesis produced at the end.
Here the learner must answer declared questions about intermediate
uncertainty, and its individual reads and changes are charged. The hard
prefixes can occur within evidence streams of arbitrary length: later
samples cannot remove an obligation to answer correctly at an earlier
prefix. Final-identification tasks can instead allow a learner to trade
samples for memory without meeting this interim interface. This contrast
between output obligations explains why the present bounds and
memory--sample bounds answer different questions; it does not rule out
memory--sample tradeoffs for other learning protocols.

\subsection{Correspondence that a consumer can exploit}

Ramsey's (2007) job-description challenge asks what a structure does that
makes its representational characterization explanatory. Structural
accounts locate part of that role in a correspondence used by a cognitive
mechanism. Morgan (2014) disputes the proposed distinction between
structural and receptor representations; G{\l}adziejewski and Mi{\l}kowski
(2017) defend it through causally exploited similarity and endogenous
control. This debate already makes exploitation central. The bounded-access
model specifies an operational constraint on it: the consumer must expose
the required distinctions through its permitted physical reads.

The access-gap construction shows why that constraint adds information.
An invertible recoding can retain every target bit while making every
permitted short read transcript uninformative. Its additional storage is
fully charged, and the obstruction disappears at the proved read threshold.
Predictive $\mathcal V$-information similarly makes usefulness relative to
an allowed predictor family (Xu et al., 2020); the transcript formulation
also separates the information exposed by access from the installed answer
rule. An exact consumer can be tested against both obligations. These
results connect resource accessibility to a functional representational
role while leaving the broader metaphysics of mental content open.

\subsection{Acquiring within a memory that must remain usable}

Repeated installation has a direct coding-theoretic predecessor in
write-efficient memory (Ahlswede \& Zhang, 1989): every current codeword
must permit a bounded-cost transition into every next message class.
Preserving independently queryable old content confines those transitions
to the old decoder's fibers. Their neighborhood sizes, recurrent covering
properties, and the information needed to choose a valid change determine
different costs. The contrast between majority and parity fibers, and the
sharp parity-update read bound, exhibit these differences within explicit
finite systems. Floating codes likewise demonstrate benefits of sharing
cells while preserving other variables, under monotone multilevel writes
(Jiang, Bohossian, \& Bruck, 2007). The transition primitives and native
read obligations must be included when comparing implementations.

Other coding models isolate different restrictions. Write-once memory
permits only monotone cell changes and a finite rewrite horizon (Rivest
\& Shamir, 1982). Defect-masking codes can grant the encoder knowledge of
defective cells unavailable to the decoder (Kuznetsov \& Tsybakov, 1974).
Those primitives and information assumptions change the set of admissible
updates. The reversible Hamming neighborhoods and ordinary-overwrite reads
used here are specified separately, so none of those extra capabilities
enters a preservation proof implicitly.

Continual-learning theory addresses related retention demands.
Knoblauch, Husain, and Diethe (2020) analyze the parameters compatible with
previous tasks. Chen, Papadimitriou, and Peng (2022) prove a worst-case
memory lower bound for one-pass learning of jointly realizable task
distributions, including improper learners, and show how additional passes
can reduce memory. Complementary learning systems explain how rapid
acquisition and interleaved integration can limit interference in
connectionist memory (McClelland, McNaughton, \& O'Reilly, 1995).
Corollary~\ref{shared:zero-read-separation} isolates one architectural
constraint: if a family of ordinary overwrites reads no memory, preserves
old content, and maps a common ready-state domain into itself, then the old
decoder restricted to that domain factors through the coordinates the
family never writes. This is a statement about the allowed states, not
about the decoder on the entire physical cube. A decoder may depend
outside the ready domain on coordinates that are safely overwritten
inside it. Within that domain, sharing information essential to old
content with the written coordinates requires a more informative update,
a different legal-sequence contract, or a different write primitive. This conditional result supplies a precise
partition principle without identifying a biological mechanism.
Our results isolate obligations within the retained store: old queries
must remain correct, new evidence must change the appropriate resource,
and the updater must find a preserving operation using its available
reads. They make the interaction between retention and acquisition
explicit without requiring a particular replay mechanism or neural
architecture.

\subsection{Systematicity and fixed-slot codes}
\label{d:systematicity}

Fodor and Pylyshyn (1988) argued that systematicity, the fact that a capacity to represent some contents comes with a capacity to represent their recombinations, is explained by classical constituent structure: tokens of constituents occur in tokens of complex representations and are processed by structure-sensitive operations. Smolensky (1988, 1990) replied that distributed tensor-product representations have constituents of a nonclassical kind and support systematic processing. Van Gelder (1990) called the relevant property functional compositionality: general and reliable processes produce a representation from its constituents and recover them again, without requiring that constituent tokens be present in the compound. Fodor and McLaughlin (1990) answered that constituents of this kind do not explain why systematicity holds.

The independent-assignment task of Appendix~\ref{d:factors} makes one part of this dispute exact. Its coordinates are fixed by the external interface: each can be set to any value while the others are preserved, and each can be read after any finite sequence of assignments. By Theorem~\ref{d:nativefactors}, every exact implementation reaches every combination of coordinate values, its native reads and actual assignments obey the selective-update law on reachable states, and its behavioral quotient is the product of the coordinate sets. Systematic capacity relative to the declared coordinates is therefore a consequence of exact success, whatever the architecture; it is functional compositionality in van Gelder's sense. It does not place coordinates in storage. A bijective recoding of the complete states preserves every answer (Proposition~\ref{d:recoding}), and the two-cell code $(x_1\oplus x_2,x_2)$ of Appendix~\ref{d:factors} implements two independent bits although $x_1$ is carried by no single cell. No behavioral test of this interface excludes distributed encodings of this kind.

Resource premises supply the further step. If the task is implemented in exactly as many cells as coordinates, each cell taking as many values as a coordinate, and every assignment changes at most one cell, then Corollary~\ref{d:minimalalignment} and Proposition~\ref{d:multivalued} place each coordinate in its own cell under a fixed value code. At tight binary storage, Theorem~\ref{d:robust-tight} gives an approximate counterpart when the update stretch, averaged uniformly over semantic edges, is close to one. These premises force a \emph{fixed-slot code}. It is weaker than classical constituency: the value code may differ between slots, no constituent token is reused across roles, and no structure-sensitive processing beyond the declared reads and assignments follows. Within this model, the step from systematic behavior to slot-located coordinates requires a storage and update premise that the original argument did not state, and the step to classical constituency requires more. The reply that functional compositionality is all that systematic behavior requires is correct about behavior; it must still accept slot-located coordinates wherever these budgets hold. Fodor and McLaughlin's reason, that tensor-product constituents are not tokened when the complex representation is tokened and so cannot be causally efficacious in processing, also has a bounded-access analogue: by Theorem~\ref{d:availability-sharing}, an invertible recoding can retain every coordinate while every read procedure within the permitted depth obtains no information about any of them. Whether such storage and update budgets hold for a particular cognitive or artificial system is an empirical question about its states and accesses, not a consequence of its systematic behavior.

The same separation limits what behavioral evaluation can establish. Benchmarks of systematic generalization test whether a system answers correctly on recombinations of familiar parts (Lake \& Baroni, 2018; Hupkes et al., 2020). Exact success on all assignment-and-read continuations, with the content carried in retained state between an assignment and a later read, establishes the native factors of Theorem~\ref{d:nativefactors} for that interface. A benchmark that presents the complete input at test time does not test retained content, and success on a finite test set does not establish exact success on an unbounded family of continuations. Even complete behavioral success leaves the physical encoding undetermined. A claim about physical constituents therefore requires evidence about internal states, such as interventions of the kind formalized in causal-abstraction analysis (Geiger et al., 2021), or measured storage and update budgets of the kind assumed above. Behavioral benchmarks can at most establish functional compositionality for a declared interface; they cannot by themselves establish physical constituent structure.

\section{What the resource analysis establishes}
\label{focus:discussion}

Acquisition changes an implementation in ways that its subsequent consumer must be able to exploit. The learning curve shows how a restrictive read contract makes maintaining those changes expensive. The locality theorem shows when a putative selector must itself become a description with independently updated coordinates. Neither result requires interpreting a code as historically novel. Their subject is the operational organization that experience establishes in a complete retained state.

Preservation adds a further constraint on the same process. Shared memory is governed by the fibers of its actual old-content decoder. One-shot availability is a neighborhood question; repeated availability is a covering question over the states that future installations can reach. Determining a safe write can require reading the old state even when the write distance is small. The sequential-evidence construction connects these distinctions to acquired uncertainty, so the analysis is not confined to installing a fully specified new label.

The access measure makes the functional interpretation precise at another level. Total mutual information is invariant under an invertible recoding, whereas information exposed by a bounded physical reader can change from complete to zero. This availability score bounds what any permitted consumer can achieve; the actual installed consumer still needs the correct answer rule. The content criterion and the resource criterion therefore remain separate obligations. The robust coordinate theorem connects two of the resource measures: under tight capacity and uniform semantic edges, small average update stretch guarantees approximate coordinate carriers and a small one-read information gap. A worst-case statement about every coordinate does not follow: the triangular encoding $(x_1\oplus x_2,x_2,\ldots,x_N)$ in Section~\ref{d:robust-locality} has near-unit average stretch while its first semantic bit remains inaccessible to every one-cell reader.

Horn acquisition supplies a second exact implementation comparison. A complete conditional-answer table provides one-read use at a sharply determined maintenance cost. At tight capacity, one-change named insertion and deletion force an explicit rule vector; its optimal conditional-query cost is $2^{n_{\rm at}}-2^{|F|}$ for an atom outside $F$. This is an exact counterpart to the answer-table endpoint. Storing models discards provenance but supports additive acquisition with a different read--write balance. The semantic hierarchy determines which resources are sufficient; it does not order all their implementation costs monotonically.

\subsection{Open questions and the limits of the proved frontiers}
\label{focus:open-questions}

The following questions distinguish the established bounds from stronger frontiers that remain open.
\begin{enumerate}[leftmargin=*,itemsep=4pt]
\item \emph{Adaptive and larger-depth maintenance.} At two adaptive reads, the canonical maintenance exponent lies between $N/3$ and the block upper value $N/2$. For growing adaptive depth, the available count has an exponent proportional to $N/(2^{r_{\rm qry}}-1)$; it does not yield the stronger order $N/r_{\rm qry}$. The leading nonadaptive exponent is determined for $r_{\rm qry}\le\alpha\log_2N$ with fixed $0<\alpha<1$, but not for larger read budgets.
\item \emph{History dependence and prescribed storage.} The endpoint-multiplicity bound becomes weak when selected histories end in many different physical states. The arguments do not determine unrestricted history-dependent maintenance or the optimal update cost at every prescribed storage and read budget.
\item \emph{Bounded-error organization.} The randomized storage restriction does not supply bounded-error analogues of all exact carrier and maintenance results. These would require controlling approximate native answers together with the updates that sustain them.
\item \emph{Recurrent installation below saturation.} Perfect codes characterize when the saturated one-write capacity can be maintained indefinitely. The optimal nonsaturated recurrent capacity remains to be determined for general free-storage sizes and write budgets.
\item \emph{Robustness with surplus storage.} Tight complete storage removes endpoint multiplicity by counting. With surplus storage, multiplicity along selected acquisition paths does not establish the global geometry needed for Theorem~\ref{d:robust-tight}. An appropriate robust carrier theorem requires additional assumptions or a stronger measured control on the reachable encoding.
\item \emph{Computing and using installation labels.} Fiber geometry gives an exact finite feasibility criterion for a declared distributed decoder. Computing its optimal labeling, describing that labeling compactly, and implementing a preserving updater within charged read and write budgets are separate algorithmic problems.
\end{enumerate}
These questions concern specified resource models. A stability--plasticity law for gradient-based learning would additionally require a justified correspondence between its actual state, updates, and consumers and the resources charged here.

Within these boundaries, the conclusion is concrete: already acquired information does not determine its own accessibility, safe update cost, or physical organization. Those properties depend on the consumer and update contract. Exact local constraints can force coordinate carriers; weaker constraints permit other encodings with quantitatively different costs. This supplies a resource-relative account of how experience establishes and maintains content that a system can actually use.

\section*{Materials and reproducibility}
\addcontentsline{toc}{section}{Materials and reproducibility}

This is a theoretical study. The finite verification programs and the
selected Lean certificates described in Appendix~\ref{focus:verification}
are available from the author on request.
The finite results are generated by these programs; no participant
dataset is used. The mathematical arguments are stated in the article.

\appendix

\section{Functional factors forced by the task interface}\label{d:factors}

\subsection{Independent assignments and native coordinate reads}

Let $M_1,\ldots,M_N$ be finite sets with at least two elements, and let
$M=\prod_{i=1}^{N}M_i$. The task starts at a specified tuple $x^0\in M$.
An assignment command $(i,a)$ replaces coordinate $i$ by $a\in M_i$;
write $\mathsf{Upd}_{i,a}:M\to M$ for this operation. A read command $i$ asks for
the current value of coordinate $i$. Every finite assignment sequence and
every subsequent coordinate read are admissible. Independence is therefore
a property of the externally specified intervention interface. It is not an
assumption about the implementation's storage.

An implementation has an initial state $q_0$, actual assignment programs
$\delta_{i,a}$, and native read programs $O_i:Q\to M_i$. Let $Q_{\mathrm r}$ be
the states reachable from $q_0$ by these programs, and define
\begin{equation}
 c(q)=(O_1(q),\ldots,O_N(q)).\label{d:nativetuple}
\end{equation}
Here $c$ is determined by the system's actual reads, rather than by a
separately trained decoder. Relative to this interface, it is the native
content map with $b(x,i)=x_i$.

\begin{theorem}[Native factors and selective updates]\label{d:nativefactors}
The implementation answers every admissible task correctly if and only if
the following conditions hold:
\begin{align}
 c(q_0)&=x^0,\label{d:initialtuple}\\
 O_i(\delta_{i,a}q)&=a
       &&(q\in Q_{\mathrm r},\ a\in M_i),\label{d:ownread}\\
 O_j(\delta_{i,a}q)&=O_j(q)
       &&(q\in Q_{\mathrm r},\ j\ne i,\ a\in M_i).
       \label{d:frameread}
\end{align}
When these conditions hold, $c$ is onto $M$ and
\begin{equation}
             c\circ \delta_{i,a}=\mathsf{Upd}_{i,a}\circ c
             \quad\hbox{on }Q_{\mathrm r}.\label{d:coordinateupdate}
\end{equation}
Two reachable states have the same answers to all allowed future
assignment-and-read continuations if and only if they have the same
$c$-value. Consequently the behavioral quotient of $Q_{\mathrm r}$ is
isomorphic, with its designated reads and assignments, to $M$.
\end{theorem}

\begin{proof}
Suppose first that the implementation is correct. Empty-history queries
give~\eqref{d:initialtuple}. If $q$ is reached after evidence word $e$, let $x$ be
the task tuple after $e$. Correctness gives $c(q)=x$. Appending $(i,a)$ and
then asking $i$ gives~\eqref{d:ownread}; asking any $j\ne i$ instead gives
\eqref{d:frameread}. These are properties of the actual transition at $q$,
since the state is complete and the same programs handle the continuation.

Conversely, the two read equations state exactly
\eqref{d:coordinateupdate}. Starting from~\eqref{d:initialtuple}, induction
on the length of an assignment history shows that $c$ of its actual
endpoint equals the task tuple. Every terminal coordinate answer is
therefore correct. To reach any $x\in M$, assign coordinates
$1,\ldots,N$ their values in $x$. The endpoint has $c$-value $x$, so $c$
is onto. If two states have equal $c$-values, iterating
\eqref{d:coordinateupdate} preserves that equality through every common
assignment continuation, and hence through every terminal read. If their
$c$-values differ, an immediate coordinate read distinguishes them. Thus
the fibers of $c$ are exactly the behavioral equivalence classes. Its
induced bijection preserves all designated operations, proving the
quotient claim.
\end{proof}

The result identifies functional components and their update law. The
equations resemble the read--write and noninterference laws of lenses and
the algebra of state (Foster et al., 2007; Foster, Zeyda, \& Woodcock,
2016; Plotkin \& Power, 2002). The distinction needed here is that the
theorem derives the laws on native observations from exact task success.
It does not derive all lens laws on complete raw states.

For example, let a system have states $(x_1,x_2,h)\in\bits^3$, read only
$x_1,x_2$, and execute
\[
 \delta_{1,a}(x_1,x_2,h)=(a,x_2,0),\qquad
 \delta_{2,\xi}(x_1,x_2,h)=(x_1,\xi,1-h).
\]
From $000$ it is an exact implementation of two independent binary
coordinates. Nevertheless, $\delta_{1,a}\delta_{2,\xi}$ ends with hidden bit $0$,
whereas $\delta_{2,\xi}\delta_{1,a}$ ends with hidden bit $1$. Moreover
$\delta_{2,\xi}^2\ne \delta_{2,\xi}$. Commutation and overwrite hold on its behavioral
quotient, while both fail on raw states. Conversely, minimal storage does
not itself identify physical coordinates. The bijective code
\[
                   \mathsf{enc}(x_1,x_2)=(x_1\mathbin\oplus x_2,x_2)
\]
with conjugated updates and reads implements the task in exactly two
binary cells, although reading $x_1$ requires both cells. Physical
conclusions require premises about physical access or change.

\subsection{Natural blocks of Horn content}\label{d:logicalblocks}

Logical dependency can prevent an atomwise product. Fix a finite atom set
$\mathsf{At}$ and a definite Horn theory $K$. A rule is $B\to \zeta$, with $B\subseteq
\mathsf{At}$ and $\zeta\in \mathsf{At}$; the empty body is allowed. Let $\cl_K(F)$ be the
least $K$-closed superset of $F$. The closed contents form
$\mathcal M=\{Z\subseteq \mathsf{At}:\cl_K(Z)=Z\}$. An atom addition acts on content by
$\mathsf{Upd}_{\zeta}(Z)=\cl_K(Z\cup\{\zeta\})$.

Fix a partition $\mathsf{At}=\mathsf{At}_1\sqcup\cdots\sqcup \mathsf{At}_\ell$ into nonempty blocks.
For $F\subseteq \mathsf{At}_i$, put
\[
 \cl_i(F)=\cl_K(F)\cap \mathsf{At}_i,\qquad
 \mathcal M_i=\{Z_i\subseteq \mathsf{At}_i:\cl_i(Z_i)=Z_i\}.
\]
These are closure operators: extensivity and monotonicity are inherited,
and $\cl_K(\cl_K(F)\cap \mathsf{At}_i)\cap \mathsf{At}_i\subseteq \cl_K(\cl_K(F))\cap \mathsf{At}_i=\cl_i(F)$ gives
idempotence together with extensivity. Every globally closed set restricts
to a locally closed set.

\begin{theorem}[Exact criterion for natural logical blocks]\label{d:hornblocks}
The following conditions are equivalent.
\begin{enumerate}
\item The restriction map
\[
 R:\mathcal M\longrightarrow\prod_{i=1}^{\ell}\mathcal M_i,
           \qquad R(Z)=(Z\cap \mathsf{At}_i)_{i=1}^{\ell},
\]
is onto, and hence an order isomorphism.
\item For every $F\subseteq \mathsf{At}$ and every $i$,
\begin{equation}
              \cl_K(F)\cap \mathsf{At}_i=\cl_K(F\cap \mathsf{At}_i)\cap \mathsf{At}_i.
              \label{d:blocklaw}
\end{equation}
\item Each rule $B\to \zeta$ of the given theory, with $\zeta\in \mathsf{At}_i$, satisfies
$\zeta\in \cl_K(B\cap \mathsf{At}_i)$.
\item $K$ has an equivalent definite Horn presentation in which the body
and head of each rule lie within a single block.
\end{enumerate}
Under these conditions, adding $\zeta\in \mathsf{At}_i$ changes only component $i$ of
$R(Z)$, replacing it by $\cl_i((Z\cap \mathsf{At}_i)\cup\{\zeta\})$.
\end{theorem}

\begin{proof}
The restriction map is always injective and order-preserving and
order-reflecting. Suppose it is onto. For any $F\subseteq \mathsf{At}$, let
$Z_i^{\rm loc}=\cl_i(F\cap \mathsf{At}_i)$. Surjectivity implies that
$Z^{\rm loc}=\bigcup_iZ_i^{\rm loc}$ is globally closed. Since $F\subseteq Z^{\rm loc}$, we have
$\cl_K(F)\subseteq Z^{\rm loc}$. Conversely $Z_i^{\rm loc}\subseteq \cl_K(F)$ for every $i$ by
monotonicity, so equality holds and proves~\eqref{d:blocklaw}. If the
block law holds instead, the union of any tuple of locally closed sets
is globally closed, because its closure restricts in each block to that
same local set. Thus the restriction map is onto.

Applying~\eqref{d:blocklaw} to a rule body proves condition 3. Conversely,
assume condition 3 and again put $Z_i^{\rm loc}=\cl_i(F\cap \mathsf{At}_i)$ and
$Z^{\rm loc}=\bigcup_iZ_i^{\rm loc}$. If $B\to \zeta\in K$, $B\subseteq Z^{\rm loc}$, and $\zeta\in \mathsf{At}_i$,
then
\[
 \zeta\in \cl_K(B\cap \mathsf{At}_i)\cap \mathsf{At}_i
     \subseteq \cl_K(Z_i^{\rm loc})\cap \mathsf{At}_i=\cl_i(Z_i^{\rm loc})=Z_i^{\rm loc}.
\]
Hence $Z^{\rm loc}$ is $K$-closed. The preceding two inclusions give $\cl_K(F)=Z^{\rm loc}$ and
the block law. For condition 3 implies condition 4, replace each rule
$B\to \zeta$ with $(B\cap \mathsf{At}_i)\to \zeta$, where $\zeta\in \mathsf{At}_i$. Every replacement
is entailed by $K$, and it entails its original rule, so the theories are
equivalent. Finally, a theory consisting of within-block rules computes
each block's closure independently, proving condition 2 from condition
4. Applying the block law to $Z\cup\{\zeta\}$ proves the update statement.
\end{proof}

The criterion is semantic rather than merely a test for cross-block
syntax. For blocks $\{\zeta_1\}$ and $\{\zeta_2,\zeta_3\}$, the theory
$\{\zeta_2\to\zeta_3,\ \{\zeta_1,\zeta_2\}\to\zeta_3\}$ has a cross-block rule but satisfies the
criterion because that rule is redundant. If the block law fails, choose
$F$ witnessing the failure and an atom
\[
 \zeta\in \cl_K(F)\setminus\bigcup_i \cl_i(F\cap \mathsf{At}_i).
\]
An implementation that replaces global closure by independent local
closures then gives the wrong native answer to the query for $\zeta$.

\begin{corollary}[Finest natural partition from minimal supports]
\label{d:supportgraph}
For each atom $\zeta$, let $\operatorname{Supp}_K(\zeta)$ consist of the inclusion-minimal
sets $B\subseteq \mathsf{At}$ such that $\zeta\in \cl_K(B)$. Form an undirected graph on
$\mathsf{At}$ with an edge between distinct $\zeta,\zeta'$ whenever $\zeta'\in B$ for some
$B\in\operatorname{Supp}_K(\zeta)$. A partition satisfies
Theorem~\ref{d:hornblocks} if and only if every graph component is
contained in one of its blocks. Thus the connected components give the
unique finest natural atom partition satisfying that theorem.
\end{corollary}

\begin{proof}
If the block law holds and $\zeta\in \mathsf{At}_i$, a minimal support $B$ of $\zeta$
satisfies $\zeta\in \cl_K(B\cap \mathsf{At}_i)$. Minimality forces $B=B\cap \mathsf{At}_i$.
Consequently no graph edge crosses blocks. Conversely, if no edge
crosses blocks and $\zeta\in \cl_K(F)\cap \mathsf{At}_i$, finiteness supplies a minimal
support $B\subseteq F$ of $\zeta$. All its elements lie in $\mathsf{At}_i$, so
$\zeta\in \cl_K(F\cap \mathsf{At}_i)$. The reverse inclusion in the block law follows
from monotonicity. The graph description of the finest partition follows
immediately. Empty minimal supports for unconditional facts cause no
exception: they add no edges.
\end{proof}

Minimal supports and direct implicational bases are established
closure-theoretic objects (Bertet \& Monjardet, 2010). The corollary uses
them to make the specified functional factorization checkable; it does
not introduce a new implicational-basis theory. Nor does it characterize
every abstract factorization of a closure lattice. For
\[
 K=\{\zeta_3\to\zeta_1,\ \zeta_3\to\zeta_2,\ \{\zeta_1,\zeta_2\}\to\zeta_3\},
\]
the closed sets are $\varnothing,\{\zeta_1\},\{\zeta_2\},\{\zeta_1,\zeta_2,\zeta_3\}$, an abstract
Boolean square. The support graph is nevertheless connected, so there is
no nontrivial natural atom partition. Here the atom $\zeta_3$ records a
conjunction of two coordinates in an abstract product. The task's chosen
observables determine which claim of independence is being made.

\section{When local acquisition forces physical coordinates}\label{d:locality}

\subsection{The physical model and a canonical encoding}
\label{d:physicalmodel}

Consider the independent binary task of Appendix~\ref{d:factors}. Its
complete persistent state at each task boundary consists of $s$ binary
cells, so reachable raw states belong to $\bits^s$. The actual read
programs determine $c:Q_{\mathrm r}\to\bits^N$, and exact acquisition
requires
\begin{equation}
                  c(\delta_{i,a}q)=\mathsf{Upd}_{i,a}(c(q)).
                  \label{d:physicalcorrect}
\end{equation}
The physical locality condition is
\begin{equation}
                  d_H(q,\delta_{i,a}q)\leq1,
                  \label{d:onebit}
\end{equation}
where $d_H$ is Hamming distance. This constrains the difference between
complete boundary states. A program that writes at most one cell per
operation satisfies it; the converse need not hold, since intermediate
writes may cancel.

An injective map $\mathsf{enc}:\bits^N\to Q_{\mathrm r}$ is a
\emph{canonical encoding closed under the actual updates} when
\begin{equation}
 c(\mathsf{enc}(x))=x,\qquad
 \delta_{i,a}\mathsf{enc}(x)=\mathsf{enc}(\mathsf{Upd}_{i,a}x)
 \quad\hbox{for all }x,i,a.
 \label{d:canonicalclosed}
\end{equation}
Choosing one representative from every content fiber supplies the first
equation, but generally not the second. The second equation excludes
history-dependent choices of representatives along the designated
updates. The theorem below is conditional on both equations, or on
premises that entail them.

\subsection{Rigidity of local canonical encodings}
\label{d:rigiditysection}

\begin{lemma}[Four-cycle locality]\label{d:square-locality}
Let $q^0,q^1,q^2,q^3$ be four distinct words in
$\prod_{j=1}^s A_j$, where $|A_j|\leq3$ for every cell. Suppose
successive words, with indices taken modulo four, differ in exactly one
cell. The four changed addresses alternate between two distinct cells.
In particular, opposite edges have the same address.
\end{lemma}

\begin{proof}
Every cell changed along a closed walk must be changed at least twice:
a single change cannot restore its initial value. With four edges, the
only possibilities are one cell changed four times or two cells changed
twice each. The first possibility confines all four distinct words to
one cell, which has at most three values, and is therefore impossible.
In the second possibility, two successive changes to the same cell
would have to restore that cell before the other cell is changed.
The walk would then repeat a vertex. The two addresses must consequently
alternate.
\end{proof}

\begin{theorem}[Binary coordinate rigidity]\label{d:rigidity}
Let $N\geq1$ and let $\mathsf{enc}:\bits^N\to\bits^s$ be injective. Suppose that
the images of every pair of tuples differing in exactly one coordinate
also differ in exactly one physical coordinate. Then there are an
injection $\pi:\{1,\ldots,N\}\to\{1,\ldots,s\}$ and a fixed
$q^\circ\in\bits^s$ such that
\begin{equation}
 \mathsf{enc}(x)_{\pi(i)}=x_i\mathbin\oplus q^\circ_{\pi(i)},\qquad
 \mathsf{enc}(x)_j=q^\circ_j\quad(j\notin\operatorname{im}\pi).
 \label{d:coordinateembedding}
\end{equation}
In particular, an exact implementation satisfying
\eqref{d:onebit} and~\eqref{d:canonicalclosed} has a canonical image
that is a coordinate subcube. Each semantic coordinate has its own
physical cell, up to a fixed complement.
\end{theorem}

\begin{proof}
For each edge of the semantic cube, label it by the unique physical
coordinate changed by its image. The four distinct images of every
semantic square satisfy Lemma~\ref{d:square-locality}. Opposite edges
therefore have the same label, and adjacent edges have different labels.

Any two semantic edges parallel to coordinate $i$ can be connected by a
sequence of such opposite-edge moves: change, one at a time, the other
coordinates of their starting vertices. Thus all edges parallel to $i$
have a fixed physical label $\pi(i)$. At any one vertex, distinct
incident edges have distinct labels, since otherwise their two endpoints
would have the same image, contrary to injectivity. Hence $\pi$ is an
injection. The argument for $N=1$ requires no squares: there is one edge
and one label.

Put $q^\circ=\mathsf{enc}(0)$. Reach any $x$ from $0$ by toggling once each coordinate for
which $x_i=1$. Its image toggles exactly the cells $\pi(i)$ for those
coordinates and leaves all other cells unchanged. This proves
\eqref{d:coordinateembedding}. In the implementation application,
\eqref{d:canonicalclosed} and~\eqref{d:onebit} imply that each semantic
edge has endpoints at physical distance at most one. They are distinct
by injectivity, so the distance is exactly one and the first part applies.
\end{proof}

For $s=N$, the conclusion is the usual signed-coordinate form of a
hypercube automorphism (Harary, 2000). Hypercube embeddings also have an
established geometric theory (Djokovi\'{c}, 1973; Winkler, 1984). The
elementary proof is included to expose the operational premise: it is
the full collection of semantic squares, together with canonicality and
binary locality, that prevents the physical direction of an update
from depending on the current content.

\begin{corollary}[Tight binary storage supplies canonicality]
\label{d:minimalalignment}
Suppose the independent binary task is implemented exactly in $N$
complete binary cells and every designated assignment changes at most
one cell. Then every reachable state is uniquely determined by its
content, and the actual encoding has the form
\eqref{d:coordinateembedding} with $s=N$ and $\pi$ a permutation.
\end{corollary}

\begin{proof}
Theorem~\ref{d:nativefactors} makes $c$ onto $\bits^N$. There are
$2^N$ contents and at most $2^N$ raw states. Every content therefore has
exactly one reachable state and all physical states are reached. Define
$\mathsf{enc}=c^{-1}$. Exactness gives~\eqref{d:canonicalclosed}; locality and
Theorem~\ref{d:rigidity} give the conclusion.
\end{proof}

Tight storage is one sufficient way to obtain canonicality, rather than
a necessary condition in Theorem~\ref{d:rigidity}. Its larger-cube version
permits unused physical cells but forces them to remain constant on the
canonical image.

\begin{proposition}[Tight storage with multivalued cells]
\label{d:multivalued}
Let every coordinate set $M_i$ of the task in Appendix~\ref{d:factors} and every cell alphabet $A_j$ have the same cardinality $q_{\rm alph}\geq2$. Suppose the task is implemented exactly in $N$ complete cells and every designated assignment changes at most one cell. Then every reachable state is uniquely determined by its content, every physical state is reachable, and there are a permutation $\pi$ of $[N]$ and bijections $f_i:M_i\to A_{\pi(i)}$ such that
\begin{equation}
 \mathsf{enc}(x)_{\pi(i)}=f_i(x_i)\qquad(x\in M,\ 1\leq i\leq N),
 \label{d:multivaluedform}
\end{equation}
where $\mathsf{enc}=c^{-1}$. Consequently $O_i(q)=f_i^{-1}(q_{\pi(i)})$ on every reachable state, and $\delta_{i,a}$ writes $f_i(a)$ into cell $\pi(i)$ and leaves every other cell unchanged. Conversely, these reads and writes, started at $\mathsf{enc}(x^0)$, give an exact implementation that changes at most one cell per assignment.
\end{proposition}

\begin{proof}
Theorem~\ref{d:nativefactors} makes $c$ onto $M$. There are $q_{\rm alph}^N$ contents and $q_{\rm alph}^N$ physical words, so $c$ is a bijection from $Q_{\mathrm r}$ onto $M$ and every word is reachable. Put $\mathsf{enc}=c^{-1}$; exactness gives $\delta_{i,a}\mathsf{enc}(x)=\mathsf{enc}(\mathsf{Upd}_{i,a}x)$. For $a\neq x_i$ these two words are distinct, and~\eqref{d:onebit} places them at Hamming distance one. Thus $\mathsf{enc}$ maps every edge of the Hamming graph on $M$, whose edges join tuples differing in one coordinate, to an edge of the Hamming graph on $\prod_{j=1}^NA_j$. Both graphs have $q_{\rm alph}^N$ vertices and $Nq_{\rm alph}^N(q_{\rm alph}-1)/2$ edges. Since $\mathsf{enc}$ is injective, it maps edges bijectively onto edges and is a graph isomorphism.

Lemma~\ref{d:square-locality} is unavailable when $q_{\rm alph}\geq4$, since four distinct values of one cell form a closed walk of four one-cell steps. Cliques replace squares. The $i$-line through $x$ is $\{\mathsf{Upd}_{i,a}x:a\in M_i\}$, and lines in the cell graph are defined in the same way. A clique containing two vertices that differ in coordinate $i$ lies in their common $i$-line: a third member differing from one of them in another coordinate would differ from the other in two coordinates. Every line is a clique, so the maximal cliques of both graphs are exactly their lines, and the isomorphism maps lines onto lines. The $N$ lines through $x$ are distinct, so their images are all $N$ lines through $\mathsf{enc}(x)$; write $\pi_x(i)$ for the cell direction of the image of the $i$-line.

Let $y=\mathsf{Upd}_{i,a}x$ with $a\neq x_i$. The two vertices share their $i$-line, so $\pi_x(i)=\pi_y(i)=:\ell$. Suppose that $j=\pi_x(i')$ and $j'=\pi_y(i')$ differ for some $i'\neq i$. Choose $b\neq x_{i'}$, and put $z=\mathsf{Upd}_{i',b}x$ and $z'=\mathsf{Upd}_{i',b}y$. These tuples differ only in coordinate $i$, whereas $\mathsf{enc}(z)$ differs from $\mathsf{enc}(x)$ only in cell $j$, $\mathsf{enc}(y)$ differs from $\mathsf{enc}(x)$ only in cell $\ell$, and $\mathsf{enc}(z')$ differs from $\mathsf{enc}(y)$ only in cell $j'$. The cells $j$, $j'$, and $\ell$ are distinct, so $\mathsf{enc}(z)$ and $\mathsf{enc}(z')$ differ in all three, contrary to adjacency. Hence $\pi_x=\pi_y$ for adjacent vertices, and connectivity gives one permutation $\pi$.

If $x$ and $x'$ agree in coordinate $i$, a path changing only coordinates other than $i$ joins them, and each step changes only a cell other than $\pi(i)$. Thus $\mathsf{enc}(x)_{\pi(i)}$ depends only on $x_i$; call it $f_i(x_i)$. If $f_i(a)=f_i(a')$ with $a\neq a'$, then a tuple $x$ with $x_i=a$ and the adjacent tuple $\mathsf{Upd}_{i,a'}x$ would have equal images, contrary to injectivity. Each $f_i$ is therefore a bijection, which proves~\eqref{d:multivaluedform}. Substituting $\mathsf{enc}$ into Theorem~\ref{d:nativefactors} gives the stated reads and writes. The converse is checked coordinatewise.
\end{proof}

For $q_{\rm alph}=2$ the proposition is Corollary~\ref{d:minimalalignment}, with each $f_i$ the identity or the complement. Its conclusion is the operational form of the description of the automorphism group of a Hamming graph as the wreath product $S_{q_{\rm alph}}\wr S_N$ (Brouwer, Cohen, \& Neumaier, 1989, Theorem~9.2.1); the proof is included because the premises are operational. Both premises are needed. The two-cell code of Appendix~\ref{d:factors} is tight but violates locality. The cardinality premise counts values per cell as well as cells: $k$ binary coordinates can share one cell with $2^k$ values, a tight store in which every assignment changes one cell, yet no coordinate has its own cell.

\section{What restricted access requires}
\label{d:access}

The preceding locality results constrain the transitions that maintain a representation. A different restriction concerns what the system's own consumer can read. A bit string may encode many independently variable contents, but that observation alone does not show that a given query can recover its content under a restricted access budget. The results in this section separate information capacity, access cost, and the interpretation of a stored descriptor. The bit-probe formulation follows the standard practice of charging accesses while allowing unrestricted computation between accesses (Miltersen, 1993; Buhrman et al., 2002).

\subsection{A common carrier criterion}

\begin{lemma}[One-probe carriers for Boolean answers]
\label{d:oneprobe-functions}
Let $X$ be a semantic domain and let $R(x)\subseteq\bits^s$ be a
nonempty set of ready states for each $x\in X$. A fixed deterministic
consumer for a Boolean query $f:X\to\bits$ must return $f(x)$ on every
$q\in R(x)$ using at most one bit probe and no other
state-dependent input. If $f$ is nonconstant, there are a fixed address
$j(f)$ and a fixed bit $\xi_f$ such that
\[
 q_{j(f)}=f(x)\mathbin\oplus \xi_f
 \qquad(x\in X,\ q\in R(x)).
\]
If two nonconstant queries share an address, their functions are equal
or complementary on $X$. Consequently, any family of pairwise distinct
nonconstant queries that all take the same value at one common semantic
point requires different designated cells for its members.
\end{lemma}

\begin{proof}
Before its first probe the consumer has no state-dependent information.
A nonconstant query therefore cannot terminate without a probe, and its
only address depends just on the public query and fixed program. Both
answer values occur, so the decoder of the probed bit must be identity
or complementation. This gives the displayed equation on all permitted
ready states. Two queries using the same cell are therefore equal or
complementary as functions of $x$. Pairwise distinct functions exclude
equality, and a common answer value at one point excludes
complementation. Their addresses must be distinct.
\end{proof}

This criterion allows correlated answers and history-dependent states.
The following finite-alphabet result treats the separate requirement
that several independently variable coordinates be recoverable from
one cell.

\subsection{One-cell access and native carriers}
\label{d:singlecell}

\begin{definition}[One-cell coordinate interface]
\label{d:onecellmodel}
Let the semantic space be a full finite product $M=\prod_{i=1}^{N}M_i$, with $|M_i|\geq2$. A ready state has storage cells $q=(q_1,\ldots,q_s)$, where $q_j\in A_j$. Every $m\in M$ is represented by at least one ready state. Query $i$ must return $m_i$ on every ready state representing $m$. Its consumer selects an address $j(i)$ from the public query alone, reads that one cell, and applies a fixed decoder $d_i:A_{j(i)}\to M_i$. There is no other target-dependent input to this consumer.
\end{definition}

The address may depend on $i$, the task parameters, and fixed program constants. It may not depend on an uncounted representation-specific routing table. Reading such a table is an additional state access. Multiple ready states for one semantic tuple are permitted.

\begin{theorem}[Capacity of a shared carrier]
\label{d:cellcapacity}
For each physical cell $j$, put $I_j=\{i:j(i)=j\}$. Every exact implementation of Definition~\ref{d:onecellmodel} satisfies
\begin{equation}
 |A_j|\ \geq\ \prod_{i\in I_j}|M_i|.
 \label{d:cellbound}
\end{equation}
\end{theorem}

\begin{proof}
Consider the map $\mathsf{Dec}_j:A_j\to\prod_{i\in I_j}M_i$ given by $\mathsf{Dec}_j(a)=(d_i(a))_{i\in I_j}$. Every tuple in its codomain extends to a member of the full product $M$. By the realization assumption, that member has a ready state $q$. Correctness gives $\mathsf{Dec}_j(q_j)=(m_i)_{i\in I_j}$. Thus $\mathsf{Dec}_j$ is surjective, and the finite cardinality inequality follows. This argument does not choose or require a unique encoding of $m$.
\end{proof}

\begin{corollary}[Binary native carriers]
\label{d:nativebitcarrier}
If each $M_i=\bits$ and each $A_j=\bits$, then the route $i\mapsto j(i)$ is injective. There are fixed bits $\xi_i$ such that, on every ready state representing $m$,
\[
 q_{j(i)}=m_i\mathbin{\oplus}\xi_i.
\]
In particular, $s\geq N$. The assertion remains true with surplus cells and with more than one ready state for the same $m$.
\end{corollary}

\begin{proof}
If $|I_j|\geq2$, Theorem~\ref{d:cellcapacity} would require $2\geq4$, so the route is injective. Each coordinate query is nonconstant on the full product. Lemma~\ref{d:oneprobe-functions} therefore supplies the displayed identity-or-complement form.
\end{proof}

This is a conclusion about the system's actual consumer. It is stronger than the existence of an external decoder, but explicitly weaker than a claim that all representations must be local. A cell with $2^k$ possible values may carry $k$ independent bits and return any of them through a query-dependent decoder. The carrier conclusion therefore depends on both the one-cell access restriction and the cell alphabet. Related one-bit representation bounds are standard in the data-structure literature (Buhrman et al., 2002); the role of the theorem here is to make the native-use premise and its limits explicit.

\section{Common dimension lemmas and a numerical illustration}
\label{d:cache}

The storage bounds in Sections~\ref{focus:central-section}
and~\ref{focus:word-random} use the same finite-alphabet span argument.
We state it independently of the learning task so that neither the bit
case nor the word case requires a separate proof. The method is the
standard polynomial representation of bounded-depth decision trees
(Buhrman \& de Wolf, 2002; Miltersen, 1993).

\subsection{The bounded-probe span}
\label{d:cache-dimension}

\begin{lemma}[Span of bounded-probe answers]
\label{d:probe-span}
Let $X$ be a finite semantic domain and let
$\mathsf{enc}:X\to\prod_{j=1}^s A_j$ be any encoding in nonempty finite cell
alphabets. The real vector space spanned by all functions on $X$
computed exactly from $\mathsf{enc}(x)$ by deterministic consumers using at most
$r_{\rm qry}\geq0$ cell probes has dimension at most
\begin{equation}
 \sum_{\substack{J\subseteq[s]\\|J|\leq r_{\rm qry}}}
       \prod_{j\in J}(|A_j|-1).
 \label{d:finite-alphabet-span}
\end{equation}
The empty product is one. If all cells have $q_{\rm alph}$ possible values, this
bound is $\sum_{j=0}^{\min(r_{\rm qry},s)}\binom{s}{j}(q_{\rm alph}-1)^j$.
The consumers may be adaptive, and the encoding need not fill the
physical state space.
\end{lemma}

\begin{proof}
Choose a reference value $0_j\in A_j$ for each cell, and write
$\phi_{j,\xi}(q)=\mathbf1\{q_j=\xi\}$ for each $\xi\ne0_j$.
Consider the common family consisting of the constant function and all
products
\[
 \prod_{j\in J}\phi_{j,\xi_j}(q),
 \qquad |J|\leq r_{\rm qry},\quad \xi_j\in A_j\setminus\{0_j\}.
\]
Its cardinality is~\eqref{d:finite-alphabet-span}.

Unfold any permitted consumer into a decision tree. A repeated read
of a cell already observed on a branch can be eliminated by using its
known value. A branch inconsistent with every encoded input may
terminate arbitrarily. Every remaining branch then tests at most $r_{\rm qry}$
distinct cells. Its indicator is a product of their outcome indicators.
An outcome different from $0_j$ contributes one $\phi_{j,\xi}$, whereas
the reference outcome contributes
$1-\sum_{\xi\ne0_j}\phi_{j,\xi}$. Expanding a leaf indicator consequently
expresses it in the displayed family. Taking the sum of these indicators
weighted by their leaf outputs expresses the consumer's answer in the
same span. Restriction to $\mathsf{enc}(X)$ cannot increase dimension. The count
therefore bounds the span of all permitted answer functions, even when
different queries use different decision trees.
\end{proof}

\begin{lemma}[Independent conjunction answers]
\label{d:query-dimension}
For $N\geq1$, the $2^N$ functions
\[
 f_S:\bits^N\to\mathbb R,
 \qquad f_S(x)=\prod_{i\in S}x_i,
 \qquad S\subseteq[N],
\]
are linearly independent, with $f_{\varnothing}=1$.
\end{lemma}

\begin{proof}
Suppose $\sum_S a_S f_S(x)=0$ for every $x$. Evaluate at the indicator
vector of a subset $B$. Since $f_S(\one_{B})=1$ exactly when $S\subseteq B$,
we obtain $\sum_{S\subseteq B}a_S=0$. For $B=\varnothing$ this gives
$a_{\varnothing}=0$. Induction on $|B|$ then gives $a_{B}=0$, because all
proper-subset coefficients have already vanished. Thus every coefficient
is zero.
\end{proof}

\subsection{Numerical scale and interface limits}
\label{d:cache-boundaries}

Table~\ref{d:cache-numerics} illustrates the complete conjunction
interface underlying Theorem~\ref{focus:central}, with 24 independent
acquired indicators. The dimension and block-cache arguments apply to
any number of independent indicators, although their particular
middle-layer realization fixes $N=\binom d{\lfloor d/2\rfloor}$. The
necessary-storage column is the smallest integer $s$ satisfying
\eqref{focus:dimension}, rather than an assertion of the exact optimum.
The other columns are achieved by the balanced block learner. Their
write counts concern changed retained bits; acquisition reads and
transient workspace remain separate resources.

\begin{table}[htbp]
\centering
\small
\begin{tabular}{rrrr}
\toprule
Read budget $r_{\rm qry}$ & Necessary bits & Constructed bits & Writes per insertion \\
\midrule
1  & 16,777,215 & 16,777,215 & 8,388,608 \\
2  & 5,793      & 8,190      & 2,048 \\
3  & 466        & 765        & 128 \\
4  & 143        & 252        & 32 \\
6  & 50         & 90         & 8 \\
8  & 33         & 56         & 4 \\
12 & 25         & 36         & 2 \\
24 & 24         & 24         & 1 \\
\bottomrule
\end{tabular}
\caption{Bounds for compound queries on 24 acquired indicators. Entries
are calculated from the proved formulas. The write column is the
construction's exact worst-case number of changed retained bits.}
\label{d:cache-numerics}
\end{table}

At $r_{\rm qry}=1$, both the storage count and the worst-case change count are
optimal. At $r_{\rm qry}=N$, direct storage uses $N$ bits, at most $N$ reads, and
one change per new indicator. For intermediate $r_{\rm qry}$, the storage--read
bounds differ by less than $2\mathrm e$. The additional maintenance bound in
Theorem~\ref{focus:star} requires canonical first-presentation endpoints;
its nonadaptive form approaches the block learner's leading update
exponent in the range established in Section~\ref{focus:central-section}.
These statements do not determine a general three-resource optimum for
adaptive or history-dependent implementations.

Exactness matters even at the smallest nontrivial dimension. For two
facts, store them directly in two cells and keep singleton queries
exact. For their conjunction, choose with equal probability among
three procedures: return zero without reading, read and return $x_1$,
or read and return $x_2$. This uses at most one probe. It is exact on
$(0,0)$ and has error $1/3$ on each other input. Thus two cells suffice
with bounded error, whereas exact deterministic one-probe use requires
three. Lemma~\ref{d:probe-span} does not by itself extend that exact
carrier conclusion to randomized consumers; the distinct bounded-error
lower bounds are stated in Section~\ref{focus:word-random}.

The one-probe carrier and dimension statements both depend on the
counted interface. A word with $2^N$ values can disclose $N$ bits in one
word read. An external copy of the evidence history or a target-dependent
routing program can also supply information, but belongs in the resource
account if the consumer can consult it. Neither a change of units nor
an omitted information channel is a compression theorem.

\section{Finite memory and cumulative retention}
\label{d:retention}

Repeated acquisition of the current target and cumulative retention of earlier
targets impose different demands on a finite store.  The distinction is not a
claim that a particular learning rule necessarily forgets.  It is an information
bound on any system whose own later consumers must still recover the acquired
distinctions.  The relevant classical random-access encoding bound is due to
Ambainis et al.\ (1999, Theorem~2.1).  We give the proof with unequal error
budgets and an explicit complete-state boundary, so that its application to
acquisition does not depend on an uncounted decoder or memory.

Let $X=(X_1,\ldots,X_N)$ consist of independent, uniformly distributed bits.
After an acquisition history, let $\mathbf Q$ be the system's complete retained state,
with at most $2^s$ possible values.  Acquisition may be randomized.  For each
index $i$, the system has its actual query consumer, returning a bit $Y_i$.
The assumption is the Markov condition
\[
             X\longrightarrow \mathbf Q\longrightarrow Y_i.
\]
Thus the consumer may compute without restriction and may use fresh independent
randomness, but it cannot consult the discarded experience, a target-dependent
external store, or a shared random seed omitted from $\mathbf Q$.  A seed retained for
later use belongs to the complete state.  Each $Y_i$ denotes a query at the
checkpoint under consideration; the theorem alone does not assert that
destructive queries can all be made successively.  Put
$\upsilon_i=\Pr(Y_i\ne X_i)$ and let
\[
 h_2(p)=-p\log_2p-(1-p)\log_2(1-p),\qquad 0\log_2 0=0.
\]

\begin{theorem}[Retention bound for actual consumers]
\label{d:retentionentropy}\label{d:retentionbound}
Under these assumptions,
\begin{equation}
 s\ \geq\ H(\mathbf Q)\ \geq\ I(X;\mathbf Q)
       \ \geq\ \sum_{i=1}^{N}\bigl[1-h_2(\upsilon_i)\bigr].
\label{d:entropyretention}
\end{equation}
This statement holds for every $\upsilon_i\in[0,1]$.
\end{theorem}
\begin{proof}
The cardinality bound gives $H(\mathbf Q)\leq s$, and $I(X;\mathbf Q)\leq H(\mathbf Q)$.
For each coordinate, data processing yields
$H(X_i\mid \mathbf Q)\leq H(X_i\mid Y_i)$.  If
$\mathsf{Err}_i=X_i\mathbin{\oplus}Y_i$, then $X_i$ is determined by $(Y_i,\mathsf{Err}_i)$,
and conversely $\mathsf{Err}_i$ is determined by $(Y_i,X_i)$.  Consequently
\[
 H(X_i\mid Y_i)=H(\mathsf{Err}_i\mid Y_i)\leq H(\mathsf{Err}_i)=h_2(\upsilon_i).
\]
The conditional chain rule and conditioning inequality give
\[
 H(X\mid \mathbf Q)=\sum_iH(X_i\mid X_1,\ldots,X_{i-1},\mathbf Q)
             \leq\sum_iH(X_i\mid \mathbf Q)\leq\sum_i h_2(\upsilon_i).
\]
Since $H(X)=N$, subtracting from $H(X)$ proves the result.  No step requires
$\upsilon_i\leq 1/2$: a reliably inverted bit is still informative.
\end{proof}

\begin{corollary}[Exact retention]
\label{d:exactretention}
If every coordinate is recoverable with zero error, the state space has at
least $2^N$ members and $s\geq N$.  These requirements are attained by storing
$X$ directly and reading it without alteration.
\end{corollary}
\begin{proof}
Equation~\eqref{d:entropyretention} gives $s\geq N$.  More directly, two
different target strings cannot give positive probability to the same state:
a coordinate on which they differ would require that state's consumer to
output both values with probability one.  Every string has a nonempty
state support, so there are at least $2^N$ states.  Direct storage attains
the bound and preserves its contents under read-only queries.
\end{proof}

There is also a sharp heterogeneous example.  For integer $0\leq s\leq N$,
retain a fixed set of $s$ coordinates exactly and answer every other
coordinate by a fresh fair guess.  Then $\upsilon_i=0$ on the retained coordinates
and $\upsilon_i=1/2$ elsewhere, so the right side of
\eqref{d:entropyretention} is exactly $s$.  This example does not claim
attainment for every specified error vector.  The classical asymptotic
coding result of Ambainis et al.\ (1999, Theorem~2.2) concerns a stronger,
uniform decoding guarantee and has an additional logarithmic storage term.

\subsection{Joint obligations to old and new contents}

Partition the $N=k_{\rm old}+k_{\rm new}$ coordinates into $k_{\rm old}$ old and $k_{\rm new}$ newly acquired
independent bits, with $k_{\rm old},k_{\rm new}\geq1$.  The errors below are measured in the
\emph{same final state distribution}, after the new acquisition.  Let
$\bar \upsilon_{\mathrm{old}}$ and $\bar \upsilon_{\mathrm{new}}$ be the respective mean
coordinate errors.

\begin{corollary}[Old--new capacity constraint]
\label{d:oldnewtradeoff}\label{d:oldnewbound}
Every such system satisfies
\begin{equation}
 s\geq k_{\rm old}\bigl[1-h_2(\bar \upsilon_{\mathrm{old}})\bigr]
       +k_{\rm new}\bigl[1-h_2(\bar \upsilon_{\mathrm{new}})\bigr].
\label{d:oldnewentropy}
\end{equation}
In particular, if the respective mean error budgets are
$\upsilon_{\rm old},\upsilon_{\rm new}\in[0,1/2]$, then
\begin{equation}
 s\geq k_{\rm old}[1-h_2(\upsilon_{\rm old})]+k_{\rm new}[1-h_2(\upsilon_{\rm new})].
\label{d:oldnewbudgets}
\end{equation}
\end{corollary}
\begin{proof}
Concavity of $h_2$ bounds each group's sum of entropies by its size times
the entropy of its mean error.  Substitute into
\eqref{d:entropyretention}.  For the second claim, use that $h_2$ is
increasing on $[0,1/2]$ and the actual mean errors do not exceed their budgets.
\end{proof}

The cost of a prescribed new-content budget can be expressed as a lower
bound on old-content error.  Write $s_{\rm new,min}=k_{\rm new}[1-h_2(\upsilon_{\rm new})]$.  If $s<s_{\rm new,min}$, that
new-content guarantee is itself impossible.  Otherwise, with
$h_2^{-1}$ denoting the inverse of $h_2$ restricted to $[0,1/2]$,
\begin{equation}
 \bar \upsilon_{\mathrm{old}}\geq
 h_2^{-1}\!\left(\max\left\{0,\,1-\frac{s-s_{\rm new,min}}{k_{\rm old}}\right\}\right).
\label{d:olderrorlower}
\end{equation}
Indeed, \eqref{d:oldnewentropy} implies that the old mean error's entropy
is at least the argument on the right.  The lower endpoint of the set of
errors having that entropy is the displayed inverse value.  The bound
therefore remains valid even if the actual old mean error exceeds $1/2$.
It is a necessary capacity condition, not a prediction of which memories
a particular update algorithm will discard.

\subsection{Reacquisition, cumulative acquisition, and correlation}

\begin{corollary}[Cumulative retention versus replacement]
\label{d:cumulativeretention}
If all $N$ independent acquired bits remain recoverable with error at most
$\upsilon_{\rm ret}<1/2$ per coordinate, then
\[
                  N\leq\frac{s}{1-h_2(\upsilon_{\rm ret})}.
\]
The same bound holds under a mean coordinate error budget
$\upsilon_{\rm ret}<1/2$.  In contrast, one retained bit suffices for arbitrarily
many exact replacement acquisitions when only the current target is queried.
\end{corollary}
\begin{proof}
The per-coordinate version follows from
\eqref{d:entropyretention}; the mean version follows by concavity.
For replacement, a round supplies the new target bit $x$ as evidence,
sets the retained bit to $x$, and subsequently answers current-target queries
by reading that bit.  A new round overwrites it.  Every round is acquired
and used exactly, although past targets are unavailable.
\end{proof}

Independence matters.  For an arbitrary jointly distributed binary target
vector, the same proof yields only
\[
 s\geq I(X;\mathbf Q)\geq H(X)-\sum_i h_2(\upsilon_i).
\]
Many correlated answers may be generated from a short retained seed.
Likewise, the numerous conjunction answers in the preceding sections have
only $N$ independent underlying bits; their extra space cost follows from
restricted access, not from their count alone.

These results separate two possible obstructions to continued learning.
Insufficient total capacity prevents simultaneous retention of the specified
independent contents; a poorly functioning update rule can instead lose
content despite adequate capacity.  The latter is a question about the
mechanisms of interference or plasticity, studied, for example, by French
(1999) and Dohare et al.\ (2024).  The information inequality neither identifies
those mechanisms nor makes forgetting inevitable in every continually
learning neural system.  Its contribution here is to state precisely which
cumulative preservation obligation cannot be met within a fixed complete
store.

\section{Finite tables and literal windows}
\label{d:literal-appendix}
For a fixed finite horizon and finite input and output alphabets, a deterministic input--output behavior has a finite table implementation: index every admissible history by a state, store its output and successor addresses, and follow those addresses on successive inputs. Induction on history length proves trace equality. The table and its history pointer are separately charged. This elementary observation is the relevant finite version of the pretabulation concern discussed by Block (1981); it asserts no useful bound on uniform program size or acquisition time.

For the stricter \emph{literal-window} architecture, an acquired $N$-bit descriptor must be copied from a contiguous length-$N$ window of one fixed linear store. Supporting all $2^N$ descriptors requires at least $2^N+N-1$ fixed bits: a store of length $B$ has only $B-N+1$ window positions. The bound is attained by a binary de Bruijn cycle of length $2^N$ with its first $N-1$ bits appended (de Bruijn, 1946). To obtain such a cycle, take the directed graph on $(N-1)$-bit words whose edges are $N$-bit words, linking each prefix to its suffix. Every vertex has equal indegree and outdegree, and the graph is strongly connected. An Euler circuit visits every window once. For $N=1$ the graph has one vertex and two loops. Selecting among the descriptors additionally needs $2^N$ distinct mutable address states, or at least $N$ address bits. These counts characterize the stated literal storage format. Section~\ref{focus:catalogue} instead analyzes the geometry and cost of its experience-dependent index, including implementations with a computational decoder.

\section{Notation}
\label{app:shared-notation}

Table~\ref{tab:shared-notation} records the conventions for content,
acquisition, and learning used in the task and capacity analyses.
Table~\ref{tab:resource-notation} gives this article's resource ledger and quantitative measures. The coverage coefficient $\kappa$ used in the task and learning analyses is reserved for that role; geometric excess and endpoint defect use named quantities here.
Local indices and auxiliary finite sets are defined where they occur.

\begingroup\small
\setlength{\tabcolsep}{5pt}
\renewcommand{\arraystretch}{1.12}
\begin{longtable}{@{}p{0.35\linewidth}p{0.61\linewidth}@{}}
\caption{Common task and content notation.}\label{tab:shared-notation}\\
\toprule
Symbol & Role \\
\midrule
\endfirsthead
\multicolumn{2}{@{}l}{\textit{Table~\thetable\ continued}}\\
\toprule
Symbol & Role \\
\midrule
\endhead
\midrule
\multicolumn{2}{r@{}}{\textit{Continued on the next page}}\\
\endfoot
\bottomrule
\endlastfoot
$M,T,Y,b$ & Content space, native query family, answer space, and semantic readout. \\
$Q,q,O$ & Complete state space, a retained state, and installed response function. \\
$C(q),c(q)$ & Candidate native contents and unique native content when defined. \\
$E,a,e,\varepsilon$ & Evidence alphabet, symbol, finite word, and empty word. \\
$\delta_a,A_e$ & Actual evidence transition and its extension to evidence words. \\
$\mathsf{Hist}$ & Prefix-closed set of legal evidence histories. \\
$\mathcal H,h,\operatorname{VS}(e)$ & Hypothesis class, candidate classifier, and version space after evidence word $e$. \\
$H_m,H_x^{\rm old},H_{z_{\rm new}}^{\rm new}$ & Exact content fiber, with old and new components distinguished by superscripts. \\
$\mathsf{At},\zeta,F,K$ & Atomic vocabulary, atom, factual input, and Horn theory. \\
$\operatorname{Supp}_K(\zeta)$ & Inclusion-minimal factual supports for an atom under $K$. \\
$\operatorname{cl}_K$ & Horn closure operator. \\
$\upsilon_{\rm qry},\upsilon_{\rm ret}$ & Native-query error and retention error. The tolerance and risk parameters $\tau,\rho$ of the acquisition specification are not redefined as resource quantities. \\
\end{longtable}

\begin{longtable}{@{}p{0.35\linewidth}p{0.61\linewidth}@{}}
\caption{Resource notation. Cell counts are binary unless a word width or cell alphabet is stated.}\label{tab:resource-notation}\\
\toprule
Symbol & Role \\
\midrule
\endfirsthead
\multicolumn{2}{@{}l}{\textit{Table~\thetable\ continued}}\\
\toprule
Symbol & Role \\
\midrule
\endhead
\midrule
\multicolumn{2}{r@{}}{\textit{Continued on the next page}}\\
\endfoot
\bottomrule
\endlastfoot
$b_{\rm fix}$ & Fixed program and preinstalled-data description length, in bits. \\
$s,s_{\rm ready},s_{\rm peak}$ & Number of retained cells, ready storage, and peak storage including temporary workspace. \\
$w,q_{\rm alph}$ & Bits per word and cell-alphabet cardinality, with $q_{\rm alph}=2^w$ in the word model. \\
$r_{\rm qry}$ & Worst-case native-query reads, charged across the whole declared request. \\
$r_{\rm upd}$ & Update reads. A blind overwrite uses no update read. \\
$c_{\rm upd}$ & All update probes: reads plus actual writes. A blind one-cell overwrite costs one probe. \\
$c_{\rm qry},c_{\rm find},c_{\rm union}$ & Total operation probes in the cited dynamic-data-structure interfaces; these can include writes. \\
$u$ & Maximum number of changed retained cells between ready endpoints; actual writes can be more numerous. \\
$\mathsf{enc}$ & Physical encoding of contents. Canonicality also requires that actual updates respect this encoding. \\
$S_{\rm req}$ & Set of requested coordinates in a compound query. \\
$N,d,n_{\rm at}$ & Number of independent acquired distinctions or target bits, instance features, and Horn atoms. \\
$n_{\rm addr},n_{\rm obj},n_{\rm inst}$ & Number of addresses in a minimum-length index, objects in relational learning, and instances in fault-tolerant acquisition. \\
$r_{\rm old},n_{\rm fam}$ & Reads per old-content bit and number of profiles in a supported family. \\
$k_{\rm old},k_{\rm new}$ & Counts of independent old and new content bits. \\
$r_{\rm syn},r_{\rm rep}$ & Syndrome dimension and repetition length. \\
$N_{\rm coal},d_{\rm act},d_{\rm cap}$ & Guaranteed coalescing-path count, active-cell count, and its storage-capped bound in the neighborhood arguments. \\
$s_{\rm free}$ & Surplus or unprotected retained cells in the stated local-capacity result. \\
$n_{\rm new},\mathcal L,z_{\rm new}$ & Number and family of newly installed labels, and one requested label. \\
$\upsilon_{\rm qry},\upsilon_i,\upsilon_{\rm ret}$ & Pointwise query error, coordinate retrieval error, and a retention-error budget. \\
$\mathsf{Exc}_{\rm loc},\mathsf{Exc}_{\rm run},\mathsf{Def}_{\rm can}$ & Excess mean encoding stretch, excess mean actual change, and mean endpoint defect in bits. \\
$\upsilon_{\rm can}$ & Mean per-bit state discrepancy from a reference encoding. \\
$R_{\rm H},U_{\rm H}$ & Canonical Horn-rule count and maximal one-insertion change in the conditional-answer table. \\
$\mathsf H$ & Parity-check matrix in the syndrome construction. \\
$\mathbf b_i,d_H,\mathrm e$ & Coordinate unit vector, Hamming distance, and Euler's number. \\
$\mathbf X,\mathbf Q$ & Random target content and random complete ready state. \\
$\mathsf{Avail}_{r_{\rm qry}}$ & Information about a specified answer exposed by the best permitted read procedure. \\
$\mathsf A_{r_{\rm qry}},\mathsf{Gap}_{r_{\rm qry}}$ & Sum of separate coordinate availabilities and the gap between jointly stored information and that sum. \\
$\mathsf{Str},\operatorname{Inf}_i,\mathsf I(f)$ & Uniform mean edge stretch, coordinate influence, and total Boolean influence. \\
$\mathsf{DT}(F,\zeta)$ & Optimal deterministic read depth for a factual Horn request in the rule-vector encoding. \\
$\mathsf{Ded}(d),\mathsf{Disj}$ & Dedekind number and the Boolean disjointness answer matrix. \\
$w_{\rm req},d_{\rm blk}$ & Maximum compound-query width and block size. \\
$J_{\rm end}$ & Number of distinct endpoints of the specified last-insertion histories. \\
$s^*,u^*,n_{\rm new,one}^*$ & Optimal storage, maintenance, and one-shot label capacity, with the stated parameters. \\
$\chi_S,\widehat f(S)$ & Boolean Fourier character and Fourier coefficient. \\
$\mathsf{Edges}_N,\mathsf{Dec}$ & Undirected semantic cube edges and a fixed interpretation program. \\
$\mathsf{ReqCount}(w_{\rm req}),\mathsf{VSCount}(d)$ & Number of bounded-width requests, including the empty request, and number of reachable consistent version spaces. \\
$\mathcal D_{r_{\rm qry}}$ & Declared family of read decision trees at the given budget. \\
\end{longtable}
\endgroup

\section{Verification materials and proof coverage}
\label{focus:verification}

The verification materials, available from the author on request, contain fifteen executable finite-check programs and a Lean~4 development. Table~\ref{tab:verification-scope} states the domain of every program. Integer ranges include both endpoints; a stated family is exhaustive only within that range. The paper supplies the general proofs, with external results cited where used. Finite enumeration checks selected instances and counterexamples. The Lean development certifies the implications specified below, rather than the entire article or its cited external theorems.

\begingroup
\small
\setlength{\tabcolsep}{5pt}
\renewcommand{\arraystretch}{1.12}
\begin{longtable}{@{}>{\raggedright\arraybackslash}p{0.30\linewidth}>{\raggedright\arraybackslash}p{0.66\linewidth}@{}}
\caption{Executable checks and their finite coverage. Filenames identify the programs in the verification materials.}\label{tab:verification-scope}\\
\toprule
Program & Checked domain and claims \\
\midrule
\endfirsthead
\multicolumn{2}{@{}l}{\textit{Table~\thetable\ continued: executable verification scope}}\\
\toprule
Program & Checked domain and claims \\
\midrule
\endhead
\midrule
\multicolumn{2}{r@{}}{\textit{Continued on the next page}}\\
\endfoot
\bottomrule
\endlastfoot
\nolinkurl{check_operational_and_locality.py} &
All injective Boolean encodings with (logical bits, physical bits) equal to $(1,3),(2,3),(2,4),(3,3)$: one-bit edge preservation and signed coordinate form. Also the four-state, history-dependent parity counterexample.\\
\nolinkurl{check_horn_blocks.py} &
All $4{,}096$ bases over the twelve non-tautological unit-Horn rules on three atoms, including facts, and all five atom partitions: local closure dependence, Cartesian factorization, and minimal-support components.\\
\nolinkurl{check_bit_probes.py} &
All injective encodings of two logical bits into two or three physical bits and all three nonempty conjunction requests: exact adaptive depth and one-read carriers. All $256$ Boolean functions on three variables: degree versus depth. Final-fact cache writes for $N=1,\ldots,8$.\\
\nolinkurl{check_tradeoffs.py} &
Truthful middle-layer evidence with $2$--$5$ features; balanced caches for $N=1$--$7$ and every read budget $1$--$N$. All encodings normalized at zero for $(N,s)=(1,1),(1,2),(2,2),(2,3),(2,4),(3,3)$: exact adaptive/nonadaptive depths, write costs, and star bounds. Two-read tree profiles on $1$--$4$ active cells.\\
\nolinkurl{check_content_organization.py} &
All nonempty consistent version spaces and their consistent evidence transitions for $0$--$4$ features. Zero-normalized one-cell acquisition embeddings for (semantic coordinates, physical cells, alphabet size) $(1,1,3)$, $(1,2,3)$, $(2,2,3)$, $(2,3,3)$, $(3,3,3)$, $(1,1,2)$, $(1,2,2)$, $(2,3,2)$, $(2,4,2)$. Carrier/split counts, the noncanonical example, and common-neighbor bounds with $1$--$3$ cells and alphabet size $2$--$5$.\\
\nolinkurl{check_word_random.py} &
Word-feature ranks for (alphabet, cells, reads) $(2,3,1)$, $(2,3,2)$, $(3,2,1)$, $(3,2,2)$, $(3,3,2)$, $(4,3,2)$; all read-tree functions in the first four cases. Conjunction ranks for $N=1$--$6$. Every vector, request, upward update, repeat, and commuting pair for the word construction with $N=1$--$7$, word size $1$--$N+2$, and read budget $1$--$N+1$. A specified three-cell private channel, all $2\times3$ binary public hash matrices, and the four-bit/six-cell/two-read communication simulation.\\
\nolinkurl{check_retention.py} &
One-read old-content decoders for $s=1$--$5$ and $1\le k_{\rm old}\le\min(3,s)$, with all neighborhood radii. Saturated recurrent decoders on $0$--$3$ spare bits. Syndrome installations with $2^j-1$ spare bits for $j=1$--$4$, including actual reads and overwrites. Fault neighborhoods on $1$--$7$ bits at radius $0$--$\min(3,s)$; repetition constructions for one or two labels and fault radius $0$--$2$.\\
\nolinkurl{check_catalogue_partition.py} &
De Bruijn windows of orders $1$--$10$; every binary catalogue one bit below the window bound for orders $1$--$3$. Indicator updates and all compound requests for $N=1$--$7$. Every history of length at most four over the eleven-symbol, three-bit interface. All reachable union-find states and pair queries for $2$--$5$ objects.\\
\nolinkurl{check_horn_resources.py} &
Every canonical Horn base on $1$--$3$ atoms: model-reader answers checked against independent forward chaining, insertion formulas, actual model-bit update costs, named-deletion diagnostics, conditional-query signatures, and middle-layer closed-set constructions.\\
\nolinkurl{check_index_geometry.py} &
All cube bijections for $N=1,2,3$: exact mean edge cost, variance estimates, one-write counts, and antipodal maxima. Every semantic edge of the stated FKM offset encoding for $N=4$--$14$. The specified two-read/high-write family for $N=1$--$14$. These are finite checks, not a verification of the general concentration theorem.\\
\nolinkurl{check_availability.py} &
Exact finite adaptive read-tree optimization for the sharing construction with (target bits, hiding threshold) $(1,1)$--$(1,5)$, $(2,1)$--$(2,3)$, and $(3,1),(3,2)$. All $4^4$ maps from two target bits to two cells, at read budgets $0,1,2$. All injections for $(N,s)=(1,1),(1,2),(1,3),(2,2),(2,3)$: influence/stretch and equality cases. Also finite Fano, monotonicity, and information-gap checks.\\
\nolinkurl{check_shared_retention.py} &
All nonconstant Boolean old decoders on $1$--$3$ bits and all surjections from three physical bits to two old bits, with all radii. All binary old/new recurrent decoders on $1$--$2$ bits; all $2$-, $3$-, and $4$-label assignments in three-bit majority fibers. Optimal ordinary-write/flip action trees for $s=2$--$4$; repeatable and promised-fresh learning for $N=1$--$4$. Zero-read support unions on full cubes through three bits and every nonempty invariant ready domain through two bits, with the stated domain and primitive counterexamples.\\
\nolinkurl{check_frontier_extensions.py} &
All vectors, requests, and updates of the bounded-width construction for $N=1$--$5$ and widths $1$--$N$. All injections of two semantic bits into two or three cells. Every nonempty star in at most three cells, at read budgets $1,2,3$. The stated history-dependent learners for $N=1$--$4$ and every last-input group count $1$--$N$, including all first-presentation histories and compound queries.\\
\nolinkurl{check_horn_query_endpoint.py} &
Every Horn base and factual request on $1$--$3$ atoms ($98{,}436$ native queries); exact decision-tree minima at those dimensions. Every assignment in the specified acyclic lower-bound restriction through four atoms, checking the nonzero full-degree coefficient. Arbitrary tight-capacity encodings are covered by the proof, not by this enumeration.\\
\nolinkurl{check_robust_locality.py} &
All balanced Boolean functions for $N=1$--$4$ and all cube bijections for $N=1$--$3$. Forty seeded near-coordinate bijections and forty canonical-defect couplings at each $N=4$--$8$. Specified adjacent-image transpositions for $N=2$--$12$, including $N=5$ with $\mathsf{Exc}_{\rm loc}=1/10$. XOR boundary instances for $N=2$--$12$, all signed-permutation comparisons for their $N=2$--$4$ instances, and the three-content-bit/four-cell history gadget. Checks include matching, expected Hamming error, and the information bound.\\
\end{longtable}
\endgroup

The word-probe ranks and geometric averages use exact rational arithmetic. Entropy calculations in the availability and robust-locality checks use floating-point tolerances declared in the source and scope file. Seeded samples are identified as samples, and the supplied seed makes them reproducible. In particular, the robust theorem is tested beyond four dimensions without claiming enumeration of all higher-dimensional bijections. Each program returns a nonzero exit status on failure; the runner records outcomes and source hashes. The machine-readable scope file records the parameters summarized in Table~\ref{tab:verification-scope}.

The Lean development has three substantive groups. First, a fixed native decoder makes endpoints for different installed labels distinct; the resulting finite injection gives a capacity inequality. This is the counting core of Theorem~\ref{focus:retention-local}; the old-carrier reduction and the binomial Hamming-ball count are proved in the text. Second, one-read star queries force distinct physical addresses, and actual changed-cell lists bound their number by the product of the neighbor count and the update budget. This certifies a one-read resource consequence of the neighborhood method; the adaptive and nonadaptive higher-depth counts in Theorem~\ref{focus:star} are outside this formal development. Third, binary Hamming geometry and a native separation contract imply disjoint correctable fibers and the clean-endpoint changed-bit lower bound in Proposition~\ref{focus:fault-lower}. The separate adaptive robust-read argument is supplied in the paper.

The development compiles with Lean~4.19.0 and contains twenty declarations, including supporting lemmas, with no \texttt{sorry}, \texttt{admit}, or declared user axioms. Its reported foundational dependencies are \texttt{propext}, \texttt{Classical.choice}, and \texttt{Quot.sound}, where used. The accompanying readme gives the commands, prerequisites, and declaration-to-result mapping. The Horn endpoint, shared-fiber recurrence, update-read, bounded-access information, geometric stability, query-width, and history-coalescence theorems have written proofs and the finite checks listed above; they are outside the Lean coverage. Proposition~\ref{d:multivalued} has a written proof and lies outside both the finite checks and the Lean coverage. Cited external theorems are likewise outside that coverage.

\clearpage
\section*{References}\addcontentsline{toc}{section}{References}
\begingroup\small
\setlength{\parskip}{0.10em}
\renewcommand{\bibitem}[1]{\par\noindent\hangindent=1.5em\hangafter=1}
\bibitem{AhlswedeZhang1989}
Ahlswede, R., \& Zhang, Z. (1989). Coding for write-efficient memory.
\emph{Information and Computation}, 83(1), 80--97.
\url{https://doi.org/10.1016/0890-5401(89)90048-5}

\bibitem{AlstrupBenAmramRauhe1999}
Alstrup, S., Ben-Amram, A. M., \& Rauhe, T. (1999). Worst-case and amortised optimality in union-find (extended abstract). In \emph{Proceedings of the Thirty-First Annual ACM Symposium on Theory of Computing} (pp.\ 499--506). ACM. \url{https://doi.org/10.1145/301250.301383}

\bibitem{Ambainis1999}
Ambainis, A., Nayak, A., Ta-Shma, A., \& Vazirani, U. (1999).
Dense quantum coding and a lower bound for 1-way quantum automata.
In \emph{Proceedings of the Thirty-First Annual ACM Symposium on Theory
of Computing} (pp.\ 376--383). ACM.
\url{https://doi.org/10.1145/301250.301347}

\bibitem{Azuma1967}
Azuma, K. (1967). Weighted sums of certain dependent random variables.
\emph{Tohoku Mathematical Journal}, 19(3), 357--367.
\url{https://doi.org/10.2748/tmj/1178243286}

\bibitem{Bertet2010}
Bertet, K., \& Monjardet, B. (2010). The multiple facets of the canonical
direct unit implicational basis.
\emph{Theoretical Computer Science}, 411(22--24), 2155--2166.
\url{https://doi.org/10.1016/j.tcs.2009.12.021}

\bibitem{Block1981}
Block, N. (1981). Psychologism and behaviorism. \emph{The Philosophical Review}, 90(1), 5--43. \url{https://doi.org/10.2307/2184371}

\bibitem{Blum1986}
Blum, N. (1986). On the single-operation worst-case time complexity of the disjoint set union problem. \emph{SIAM Journal on Computing}, 15(4), 1021--1024. \url{https://doi.org/10.1137/0215072}

\bibitem{BrodyLarsen2015}
Brody, J., \& Larsen, K. G. (2015). Adapt or die: Polynomial lower bounds for non-adaptive dynamic data structures. \emph{Theory of Computing}, 11(19), 471--489. \url{https://doi.org/10.4086/toc.2015.v011a019}

\bibitem{BrouwerCohenNeumaier1989}
Brouwer, A. E., Cohen, A. M., \& Neumaier, A. (1989).
\emph{Distance-regular graphs}. Springer.
\url{https://doi.org/10.1007/978-3-642-74341-2}

\bibitem{BuhrmandeWolf2002}
Buhrman, H., \& de Wolf, R. (2002). Complexity measures and decision tree
complexity: A survey. \emph{Theoretical Computer Science}, 288(1), 21--43.
\url{https://doi.org/10.1016/S0304-3975(01)00144-X}

\bibitem{Buhrman2002}
Buhrman, H., Miltersen, P. B., Radhakrishnan, J., \& Venkatesh, S. (2002).
Are bitvectors optimal? \emph{SIAM Journal on Computing}, 31(6), 1723--1744.
\url{https://doi.org/10.1137/S0097539702405292}

\bibitem{Carey2009}
Carey, S. (2009). \emph{The origin of concepts}.
Oxford University Press.
\url{https://doi.org/10.1093/acprof:oso/9780195367638.001.0001}

\bibitem{Chandran2014}
Chandran, N., Kanukurthi, B., \& Ostrovsky, R. (2014). Locally updatable and locally decodable codes. In Y. Lindell (Ed.), \emph{Theory of Cryptography} (LNCS 8349, pp.\ 489--514). Springer. \url{https://doi.org/10.1007/978-3-642-54242-8_21}

\bibitem{ChenPapadimitriouPeng2022}
Chen, X., Papadimitriou, C., \& Peng, B. (2022). Memory bounds for
continual learning. In \emph{2022 IEEE 63rd Annual Symposium on
Foundations of Computer Science} (pp.\ 519--530). IEEE.
\url{https://doi.org/10.1109/FOCS54457.2022.00056}

\bibitem{CorriganGibbsKogan2019}
Corrigan-Gibbs, H., \& Kogan, D. (2019). The function-inversion problem:
Barriers and opportunities. In D. Hofheinz \& A. Rosen (Eds.),
\emph{Theory of Cryptography} (Lecture Notes in Computer Science,
Vol.\ 11891, pp.\ 393--421). Springer.
\url{https://doi.org/10.1007/978-3-030-36030-6_16}

\bibitem{CrutchfieldEllisonMahoney2009}
Crutchfield, J. P., Ellison, C. J., \& Mahoney, J. R. (2009).
Time's barbed arrow: Irreversibility, crypticity, and stored information.
\emph{Physical Review Letters}, 103(9), Article~094101.
\url{https://doi.org/10.1103/PhysRevLett.103.094101}

\bibitem{deBruijn1946}
de Bruijn, N. G. (1946). A combinatorial problem. \emph{Proceedings of the Section of Sciences of the Koninklijke Nederlandse Akademie van Wetenschappen te Amsterdam}, 49(7), 758--764. \url{https://research.tue.nl/en/publications/a-combinatorial-problem/}

\bibitem{Djokovic1973}
Djokovi\'{c}, D. \v{Z}. (1973). Distance-preserving subgraphs of hypercubes.
\emph{Journal of Combinatorial Theory, Series B}, 14(3), 263--267.
\url{https://doi.org/10.1016/0095-8956(73)90010-5}

\bibitem{Dohare2024}
Dohare, S., Hernandez-Garcia, J. F., Lan, Q., Rahman, P., Mahmood, A. R.,
\& Sutton, R. S. (2024). Loss of plasticity in deep continual learning.
\emph{Nature}, 632, 768--774.
\url{https://doi.org/10.1038/s41586-024-07711-7}

\bibitem{DvirGolovnevWeinstein2019}
Dvir, Z., Golovnev, A., \& Weinstein, O. (2019). Static data structure
lower bounds imply rigidity. In \emph{Proceedings of the 51st Annual ACM
SIGACT Symposium on Theory of Computing} (pp.\ 967--978). ACM.
\url{https://doi.org/10.1145/3313276.3316348}

\bibitem{Fodor1975}
Fodor, J. A. (1975). \emph{The language of thought}.
Harvard University Press.

\bibitem{FodorMcLaughlin1990}
Fodor, J. A., \& McLaughlin, B. P. (1990). Connectionism and the problem
of systematicity: Why Smolensky's solution doesn't work.
\emph{Cognition}, 35(2), 183--204.
\url{https://doi.org/10.1016/0010-0277(90)90014-B}

\bibitem{FodorPylyshyn1988}
Fodor, J. A., \& Pylyshyn, Z. W. (1988). Connectionism and cognitive
architecture: A critical analysis. \emph{Cognition}, 28(1--2), 3--71.
\url{https://doi.org/10.1016/0010-0277(88)90031-5}

\bibitem{Foster2007}
Foster, J. N., Greenwald, M. B., Moore, J. T., Pierce, B. C.,
\& Schmitt, A. (2007). Combinators for bidirectional tree transformations:
A linguistic approach to the view-update problem.
\emph{ACM Transactions on Programming Languages and Systems}, 29(3), Article~17.
\url{https://doi.org/10.1145/1232420.1232424}

\bibitem{Foster2016}
Foster, S., Zeyda, F., \& Woodcock, J. (2016). Unifying heterogeneous
state-spaces with lenses. In \emph{Theoretical Aspects of Computing---ICTAC
2016} (Lecture Notes in Computer Science, Vol.\ 9965, pp.\ 295--314).
Springer. \url{https://doi.org/10.1007/978-3-319-46750-4_17}

\bibitem{Fredman1989}
Fredman, M. L., \& Saks, M. E. (1989). The cell probe complexity of dynamic
data structures. In \emph{Proceedings of the Twenty-First Annual ACM
Symposium on Theory of Computing} (pp.\ 345--354). ACM.
\url{https://doi.org/10.1145/73007.73040}

\bibitem{FredricksenMaiorana1978}
Fredricksen, H., \& Maiorana, J. (1978). Necklaces of beads in $k$ colors
and $k$-ary de Bruijn sequences. \emph{Discrete Mathematics}, 23(3),
207--210. \url{https://doi.org/10.1016/0012-365X(78)90002-X}

\bibitem{FreinLevequeSebo2008}
Frein, Y., L\'ev\^eque, B., \& Seb\H{o}, A. (2008). Generating all sets with bounded unions. \emph{Combinatorics, Probability and Computing}, 17(5), 641--660. \url{https://doi.org/10.1017/S096354830800922X}

\bibitem{French1999}
French, R. M. (1999). Catastrophic forgetting in connectionist networks.
\emph{Trends in Cognitive Sciences}, 3(4), 128--135.
\url{https://doi.org/10.1016/S1364-6613(99)01294-2}

\bibitem{FriedgutKalaiNaor2002}
Friedgut, E., Kalai, G., \& Naor, A. (2002). Boolean functions whose Fourier transform is concentrated on the first two levels.
\emph{Advances in Applied Mathematics}, 29(3), 427--437.
\url{https://doi.org/10.1016/S0196-8858(02)00024-6}

\bibitem{FriedgutKalaiNaor2021}
Friedgut, E., Kalai, G., \& Naor, A. (2021). \emph{FKN, first proof, rewritten}.
arXiv:2105.03246. \url{https://arxiv.org/abs/2105.03246}

\bibitem{Geiger2021}
Geiger, A., Lu, H., Icard, T., \& Potts, C. (2021).
Causal abstractions of neural networks.
In \emph{Advances in Neural Information Processing Systems}
(Vol.\ 34, pp.\ 9574--9586). Curran Associates.
\url{https://proceedings.neurips.cc/paper_files/paper/2021/hash/4f5c422f4d49a5a807eda27434231040-Abstract.html}

\bibitem{GladziejewskiMilkowski2017}
G{\l}adziejewski, P., \& Mi{\l}kowski, M. (2017). Structural
representations: Causally relevant and different from detectors.
\emph{Biology \& Philosophy}, 32, 337--355.
\url{https://doi.org/10.1007/s10539-017-9562-6}

\bibitem{Gonen2020}
Gonen, A., Lovett, S., \& Moshkovitz, M. (2020).
Towards a combinatorial characterization of bounded-memory learning.
In \emph{Advances in Neural Information Processing Systems}
(Vol.\ 33, pp.\ 9028--9038). Curran Associates.
\url{https://proceedings.neurips.cc/paper/2020/hash/665d5cbb82b5785d9f344c46417c6c36-Abstract.html}

\bibitem{Gupta1993}
Gupta, A., Mumick, I. S., \& Subrahmanian, V. S. (1993). Maintaining views
incrementally. \emph{ACM SIGMOD Record}, 22(2), 157--166.
\url{https://doi.org/10.1145/170036.170066}

\bibitem{Harary2000}
Harary, F. (2000). The automorphism group of a hypercube.
\emph{Journal of Universal Computer Science}, 6(1), 136--138.
\url{https://doi.org/10.3217/jucs-006-01-0136}

\bibitem{Haussler1988}
Haussler, D. (1988). Quantifying inductive bias: AI learning algorithms
and Valiant's learning framework. \emph{Artificial Intelligence}, 36(2),
177--221. \url{https://doi.org/10.1016/0004-3702(88)90002-1}

\bibitem{HirshMishraPitt1997}
Hirsh, H., Mishra, N., \& Pitt, L. (1997). Version spaces without boundary
sets. In \emph{Proceedings of the Fourteenth National Conference on
Artificial Intelligence} (pp.\ 491--496). AAAI Press.
\url{https://dl.acm.org/doi/10.5555/1867406.1867482}

\bibitem{Hirsh2004}
Hirsh, H., Mishra, N., \& Pitt, L. (2004).
Version spaces and the consistency problem.
\emph{Artificial Intelligence}, 156(2), 115--138.
\url{https://doi.org/10.1016/j.artint.2003.04.003}

\bibitem{Hoeffding1963}
Hoeffding, W. (1963). Probability inequalities for sums of bounded random variables.
\emph{Journal of the American Statistical Association}, 58(301), 13--30.
\url{https://doi.org/10.1080/01621459.1963.10500830}

\bibitem{Hupkes2020}
Hupkes, D., Dankers, V., Mul, M., \& Bruni, E. (2020). Compositionality
decomposed: How do neural networks generalise? \emph{Journal of Artificial
Intelligence Research}, 67, 757--795.
\url{https://doi.org/10.1613/jair.1.11674}

\bibitem{JiangBohossianBruck2007}
Jiang, A., Bohossian, V., \& Bruck, J. (2007). Floating codes for joint
information storage in write asymmetric memories. In \emph{2007 IEEE
International Symposium on Information Theory} (pp.\ 1166--1170). IEEE.
\url{https://doi.org/10.1109/ISIT.2007.4557381}

\bibitem{focus-kalyanasundaram-schnitger}
Kalyanasundaram, B., \& Schnitger, G. (1992).
The probabilistic communication complexity of set intersection.
\emph{SIAM Journal on Discrete Mathematics, 5}(4), 545--557.
\url{https://doi.org/10.1137/0405044}

\bibitem{Kleitman1969}
Kleitman, D. (1969).
On Dedekind's problem: The number of monotone Boolean functions.
\emph{Proceedings of the American Mathematical Society, 21}(3), 677--682.
\url{https://doi.org/10.1090/S0002-9939-1969-0241334-6}

\bibitem{KnoblauchHusainDiethe2020}
Knoblauch, J., Husain, H., \& Diethe, T. (2020). Optimal continual learning
has perfect memory and is NP-hard. In H. Daum{\'e} III \& A. Singh (Eds.),
\emph{Proceedings of the 37th International Conference on Machine
Learning} (Proceedings of Machine Learning Research, Vol.\ 119,
pp.\ 5327--5337). PMLR.
\url{https://proceedings.mlr.press/v119/knoblauch20a.html}

\bibitem{KociumakaRadoszewskiRytter2016}
Kociumaka, T., Radoszewski, J., \& Rytter, W. (2016). Efficient ranking
of Lyndon words and decoding lexicographically minimal de Bruijn sequence.
\emph{SIAM Journal on Discrete Mathematics}, 30(4), 2027--2046.
\url{https://doi.org/10.1137/15M1043248}

\bibitem{KST1999}
K{\"o}nig, H., Sch{\"u}tt, C., \& Tomczak-Jaegermann, N. (1999).
Projection constants of symmetric spaces and variants of Khintchine's inequality.
\emph{Journal f{\"u}r die reine und angewandte Mathematik}, 511, 1--42.
\url{https://doi.org/10.1515/crll.1999.511.1}

\bibitem{KuznetsovTsybakov1974}
Kuznetsov, A. V., \& Tsybakov, B. S. (1974). Coding in a memory with
defective cells. \emph{Problems of Information Transmission}, 10(2),
132--138. \url{https://www.mathnet.ru/eng/ppi1029}

\bibitem{LakeBaroni2018}
Lake, B. M., \& Baroni, M. (2018). Generalization without systematicity:
On the compositional skills of sequence-to-sequence recurrent networks.
In J. Dy \& A. Krause (Eds.), \emph{Proceedings of the 35th International
Conference on Machine Learning} (Proceedings of Machine Learning Research,
Vol.\ 80, pp.\ 2873--2882). PMLR.
\url{https://proceedings.mlr.press/v80/lake18a.html}

\bibitem{MahoneyEllisonCrutchfield2009}
Mahoney, J. R., Ellison, C. J., \& Crutchfield, J. P. (2009).
Information accessibility and cryptic processes.
\emph{Journal of Physics A: Mathematical and Theoretical}, 42(36),
Article~362002. \url{https://doi.org/10.1088/1751-8113/42/36/362002}

\bibitem{Mazumdar2014}
Mazumdar, A., Chandar, V., \& Wornell, G. W. (2014). Update-efficiency and local repairability limits for capacity approaching codes. \emph{IEEE Journal on Selected Areas in Communications}, 32(5), 976--988. \url{https://doi.org/10.1109/JSAC.2014.140517}

\bibitem{McClellandMcNaughtonOReilly1995}
McClelland, J. L., McNaughton, B. L., \& O'Reilly, R. C. (1995).
Why there are complementary learning systems in the hippocampus and
neocortex: Insights from the successes and failures of connectionist
models of learning and memory. \emph{Psychological Review}, 102(3),
419--457. \url{https://doi.org/10.1037/0033-295X.102.3.419}

\bibitem{Miltersen1993}
Miltersen, P. B. (1993). The bit probe complexity measure revisited.
In P. Enjalbert, A. Finkel, \& K. W. Wagner (Eds.),
\emph{STACS 93} (Lecture Notes in Computer Science, Vol.\ 665,
pp.\ 662--671). Springer.
\url{https://doi.org/10.1007/3-540-56503-5_65}

\bibitem{MiltersenNisanSafraWigderson1998}
Miltersen, P. B., Nisan, N., Safra, S., \& Wigderson, A. (1998). On data structures and asymmetric communication complexity. \emph{Journal of Computer and System Sciences}, 57(1), 37--49. \url{https://doi.org/10.1006/jcss.1998.1577}

\bibitem{Mitchell1982}
Mitchell, T. M. (1982).
Generalization as search.
\emph{Artificial Intelligence}, 18(2), 203--226.
\url{https://doi.org/10.1016/0004-3702(82)90040-6}

\bibitem{Morgan2014}
Morgan, A. (2014). Representations gone mental.
\emph{Synthese}, 191(2), 213--244.
\url{https://doi.org/10.1007/s11229-013-0328-7}

\bibitem{NatarajanRamamoorthyRao2018}
Natarajan Ramamoorthy, S., \& Rao, A. (2018). Lower bounds on non-adaptive data structures maintaining sets of numbers, from sunflowers. In R. A. Servedio (Ed.), \emph{33rd Computational Complexity Conference} (LIPIcs 102, pp.\ 27:1--27:16). Schloss Dagstuhl--Leibniz-Zentrum f\"ur Informatik. \url{https://doi.org/10.4230/LIPIcs.CCC.2018.27}

\bibitem{PatrascuThorup2011}
P\u{a}tra\c{s}cu, M., \& Thorup, M. (2011). Don't rush into a union: Take time to find your roots. In \emph{Proceedings of the Forty-Third Annual ACM Symposium on Theory of Computing} (pp.\ 559--568). ACM. \url{https://doi.org/10.1145/1993636.1993711}

\bibitem{Piantadosi2021}
Piantadosi, S. T. (2021). The computational origin of representation.
\emph{Minds and Machines}, 31(1), 1--58.
\url{https://doi.org/10.1007/s11023-020-09540-9}

\bibitem{Piccinini2022}
Piccinini, G. (2022). Situated neural representations: Solving the problems
of content. \emph{Frontiers in Neurorobotics}, 16, Article~846979.
\url{https://doi.org/10.3389/fnbot.2022.846979}

\bibitem{Plotkin2002}
Plotkin, G., \& Power, J. (2002). Notions of computation determine monads.
In \emph{Foundations of Software Science and Computation Structures}
(Lecture Notes in Computer Science, Vol.\ 2303, pp.\ 342--356). Springer.
\url{https://doi.org/10.1007/3-540-45931-6_24}

\bibitem{Ramsey2007}
Ramsey, W. M. (2007). \emph{Representation reconsidered}.
Cambridge University Press. \url{https://doi.org/10.1017/CBO9780511597954}

\bibitem{Raz2016}
Raz, R. (2016). Fast learning requires good memory:
A time-space lower bound for parity learning.
In \emph{2016 IEEE 57th Annual Symposium on Foundations of Computer Science}
(pp.\ 266--275). IEEE.
\url{https://doi.org/10.1109/FOCS.2016.36}

\bibitem{focus-razborov-disjointness}
Razborov, A. A. (1992).
On the distributional complexity of disjointness.
\emph{Theoretical Computer Science, 106}(2), 385--390.
\url{https://doi.org/10.1016/0304-3975(92)90260-M}

\bibitem{Razborov2003}
Razborov, A. A. (2003). Quantum communication complexity of symmetric predicates. \emph{Izvestiya: Mathematics}, 67(1), 145--159. \url{https://doi.org/10.1070/IM2003v067n01ABEH000422}

\bibitem{RivestShamir1982}
Rivest, R. L., \& Shamir, A. (1982). How to reuse a ``write-once'' memory.
\emph{Information and Control}, 55(1--3), 1--19.
\url{https://doi.org/10.1016/S0019-9958(82)90344-8}

\bibitem{Sakcak2024}
Sak\c{c}ak, B., Timperi, K. G., Weinstein, V., \& LaValle, S. M. (2024).
A mathematical characterization of minimally sufficient robot brains.
\emph{The International Journal of Robotics Research}, 43(9), 1342--1362.
\url{https://doi.org/10.1177/02783649231198898}

\bibitem{Shamir1979}
Shamir, A. (1979). How to share a secret.
\emph{Communications of the ACM}, 22(11), 612--613.
\url{https://doi.org/10.1145/359168.359176}

\bibitem{Shea2023}
Shea, N. (2023). Organized representations forming a computationally useful
processing structure. \emph{Synthese}, 202, Article~175.
\url{https://doi.org/10.1007/s11229-023-04373-2}

\bibitem{Sherstov2011}
Sherstov, A. A. (2011). The pattern matrix method. \emph{SIAM Journal on Computing}, 40(6), 1969--2000. \url{https://doi.org/10.1137/080733644}

\bibitem{Smolensky1988}
Smolensky, P. (1988). The constituent structure of connectionist mental
states: A reply to Fodor and Pylyshyn. \emph{The Southern Journal of
Philosophy}, 26(S1), 137--161.

\bibitem{Smolensky1990}
Smolensky, P. (1990). Tensor product variable binding and the
representation of symbolic structures in connectionist systems.
\emph{Artificial Intelligence}, 46(1--2), 159--216.
\url{https://doi.org/10.1016/0004-3702(90)90007-M}

\bibitem{vanGelder1990}
van Gelder, T. (1990). Compositionality: A connectionist variation on a
classical theme. \emph{Cognitive Science}, 14(3), 355--384.
\url{https://doi.org/10.1207/s15516709cog1403_2}

\bibitem{Westfeld2001}
Westfeld, A. (2001). F5---A steganographic algorithm: High capacity despite better steganalysis. In I. S. Moskowitz (Ed.), \emph{Information Hiding} (LNCS 2137, pp.\ 289--302). Springer. \url{https://doi.org/10.1007/3-540-45496-9_21}

\bibitem{Winkler1984}
Winkler, P. M. (1984). Isometric embedding in products of complete graphs.
\emph{Discrete Applied Mathematics}, 7(2), 221--225.
\url{https://doi.org/10.1016/0166-218X(84)90069-6}

\bibitem{WuCapacity2026}
Wu, Y. (2026a). \emph{Acquiring, keeping, and reacquiring representations}. Unpublished manuscript, BridgeM, Inc.

\bibitem{WuLearning2026}
Wu, Y. (2026b). \emph{From learning guarantees to usable content}. Unpublished manuscript, BridgeM, Inc.

\bibitem{WuTasks2026}
Wu, Y. (2026c). \emph{Tasks that force representational acquisition}. Unpublished manuscript, BridgeM, Inc.

\bibitem{XuEtAl2020}
Xu, Y., Zhao, S., Song, J., Stewart, R., \& Ermon, S. (2020).
A theory of usable information under computational constraints.
In \emph{International Conference on Learning Representations}.
\url{https://openreview.net/forum?id=r1eBeyHFDH}

\bibitem{ZhangLi2005}
Zhang, W., \& Li, S. (2005). Steganographic codes---A new problem of coding theory. \emph{arXiv}, cs/0505072. \url{https://arxiv.org/abs/cs/0505072}
\endgroup

\end{document}